\documentclass{article}
\usepackage{jfrExample}
\usepackage{algorithm}
\usepackage{algorithmic}
\usepackage{graphicx}
\usepackage{subfig}
\usepackage{apalike}
\usepackage{setspace}
\usepackage{hyperref}
\usepackage[normalem]{ulem} 
\usepackage{wrapfig}
\usepackage{amsmath}
\usepackage{amssymb,amsfonts,bm,bbm}
\usepackage{multicol}

\usepackage{xcolor}

\title{Multi-Agent Autonomy: Advancements and \\Challenges in Subterranean Exploration}

\author{
Michael T. Ohradzansky\\
Aerospace Engineering Sciences\\
University of Colorado Boulder\\
\texttt{Michael.Ohradzansky@colorado.edu}
\And
Eugene R. Rush\\
Mechanical Engineering\\
University of Colorado Boulder\\
\texttt{Eugene.Rush@colorado.edu}
\And
Danny G. Riley\\
Computer Science\\
University of Colorado Boulder\\
\texttt{Dan.Riley@colorado.edu}
\And
Andrew B. Mills\\
Aerospace Engineering Sciences\\
University of Colorado Boulder\\
\texttt{Andrew.B.Mills@colorado.edu}
\And
Shakeeb Ahmad\\
Aerospace Engineering Sciences\\
University of Colorado Boulder\\
\texttt{Shakeeb.Ahmad@colorado.edu}\\
\And
Steve McGuire\\
Electrical \& Computer Engineering\\
University of California Santa Cruz\\
\texttt{Steve.Mcguire@ucsc.edu}
\And
Harel Biggie\\
Computer Science\\
University of Colorado Boulder\\
\texttt{Harel.Biggie@colorado.edu}
\And
Kyle Harlow\\
Computer Science\\
University of Colorado Boulder\\
\texttt{Kyle.Harlow@colorado.edu}
\And
Michael J. Miles\\
Mechanical Engineering\\
University of Colorado Boulder\\
\texttt{Mike.Miles@colorado.edu}
\And
Eric W. Frew\\
Aerospace Engineering Sciences\\
University of Colorado Boulder\\
\texttt{Eric.Frew@colorado.edu}
\And
Christoffer Heckman\\
Computer Science\\
University of Colorado Boulder\\
\texttt{Christoffer.Heckman@colorado.edu}
\And
J. Sean Humbert\\
Mechanical Engineering\\
University of Colorado Boulder\\
\texttt{Sean.Humbert@colorado.edu}
}

\begin{document}
\maketitle

\begin{abstract}

Artificial intelligence has undergone immense growth and maturation in recent years, though autonomous systems have traditionally struggled when fielded in diverse and previously unknown environments. DARPA is seeking to change that with the Subterranean Challenge, by providing roboticists the opportunity to support civilian and military first responders in complex and high-risk underground scenarios. The subterranean domain presents a handful of challenges, such as limited communication, diverse topology and terrain, and degraded sensing. Team MARBLE proposes a solution for autonomous exploration of unknown subterranean environments in which coordinated agents search for artifacts of interest. The team presents two navigation algorithms in the form of a metric-topological graph-based planner and a continuous frontier-based planner. To facilitate multi-agent coordination, agents share and merge new map information and candidate goal-points. Agents deploy communication beacons at different points in the environment, extending the range at which maps and other information can be shared. Onboard autonomy reduces the load on human supervisors, allowing agents to detect and localize artifacts and explore autonomously outside established communication networks. Given the scale, complexity, and tempo of this challenge, a range of lessons were learned, most importantly, that frequent and comprehensive field testing in representative environments is key to rapidly refining system performance.

\end{abstract}


\section{Introduction}
\label{sec:intro}
Subterranean environments pose a significant challenge for autonomous robotic exploration. Maintaining consistent and accurate state estimates across long distances is critical to mapping and planning, and is difficult without the aid of GPS. Varied lighting and other environmental conditions such as heavy dust, smoke, or fog can degrade sensor measurements, and a range of sensing modalities may be needed to maintain accurate state estimates and maps. Diverse and sometimes hazardous terrain present mobility challenges to ground platforms, and can require more complex path planning and guidance algorithms or even specialized robots to navigate. Additionally, underground environments generally make wireless communication difficult by limiting the range and bandwidth that information can be transmitted. Depending on mission requirements, additional sensors may be needed such as cameras for detecting visual artifacts or Bluetooth modules for detecting cellular devices. Developing robots that are capable of overcoming all of the aforementioned challenges requires rigorous testing in representative environments.

The DARPA Subterranean Challenge \cite{DARPA2021} was created to spur the advancement of novel approaches to autonomous mapping and exploration of underground environments, and is motivated by mission-critical scenarios such as disaster response and time-sensitive combat operations. Teams are challenged with deploying robotic systems capable of navigating diverse subterranean environments in search of artifacts. The challenge is split into three circuit events (Tunnel, Urban, and Cave) and a final event that combines the challenges of all the subterranean domains. To score points during a timed run, teams must detect and localize various types of artifacts placed throughout the course within the time limit and report it to the DARPA scoring server. Searching for artifacts efficiently requires intelligent exploration strategies accompanied with able navigation and guidance. Terrain in the challenge includes flat concrete, soft dirt with rocks and boulders of varying size and shape, aggressive slopes and stairs, and even areas of thick mud, all of which provide unique planning and mobility challenges for different robot platforms. Varied lighting conditions and severely limited communication makes artifact detection, localization, and reporting a significant challenge. Competitors in the challenges are allowed one Human Supervisor who can interact directly with the robots.

Team MARBLE has competed in the Tunnel, Urban, and virtual Cave circuit events to date. At these events, the team deployed a heterogeneous fleet of ground and aerial robots with autonomy stacks consisting of perception, localization, mapping, exploration, artifact detection, communication, and multi-agent coordination algorithms. Our solution focuses on multi-agent autonomy, where agents are able to explore unknown environments efficiently by sharing information like maps and goal points. Generating consistent pose estimates is critical to coordination across platforms, and our team uses the open-source LiDAR-Inertial SLAM package Google Cartographer for localization \cite{hess2016}. In our approach, each agent localizes independently. Other teams have demonstrated distributed SLAM methods where incorporating localization information from multiple robots can improve pose estimates \cite{Ebadi2020}, \cite{Lajoie2020}. For the purposes of planning and situational awareness, a volumetric occupancy grid of the environment is created in real time using the open-source OctoMap package \cite{hornung13octomap}. Sharing entire maps over limited communications networks is undesirable, and so additional features like map difference segmenting and map merging were added to the OctoMap package. To facilitate autonomous exploration, our team has developed graph and frontier-based path planning algorithms. For the circuit events, our team's focus has been on autonomy while other teams have relied heavily on the Human Supervisor to coordinate the movements and actions of their systems \cite{Roucek2020} \cite{Otsu2020}. To traverse the subterranean environments, our team has pursued development of four-wheeled and tracked ground robots complimented with multi-rotor platforms. Other teams have deployed four-legged quadrupedal robots \cite{Miller2020}, \cite{bouman2020autonomous} and blimps \cite{Huang2019}. 

In the following work, we present MARBLE, which is a multi-agent solution for autonomous subterranean exploration and artifact detection. First, an overview of our platforms and autonomy solution is presented in Section \ref{sec:SysOver}. In Section \ref{sec:mapping}, a deeper dive into the mapping solution is provided where the implementation and benefits of map-merging is discussed. Our team has developed several planning strategies and solutions for the different subterranean environments, which are presented in Section \ref{sec:metrictopo} and Section \ref{sec:contfront}. A vision-based method for detecting and avoiding obstacles onboard aerial vehicles is discussed in Section \ref{sec:AVnav}. Section \ref{sec:comms} contains a brief discussion on a UDP-based mesh communication network for message passing between robots. A core capability developed by the team is a multi-agent coordination algorithm, which is presented in greater detail in Section \ref{sec:multiagent}. Finally, lessons the team has learned as a result of developing and deploying autonomous robot fleets for the DARPA SubT Challenge are presented in Section \ref{sec:LessionsLearned}.

\section{System Overview}
\label{sec:SysOver}
\begin{figure}[b!]
		\centering
		\subfloat[]{{\includegraphics[width=.33\textwidth]{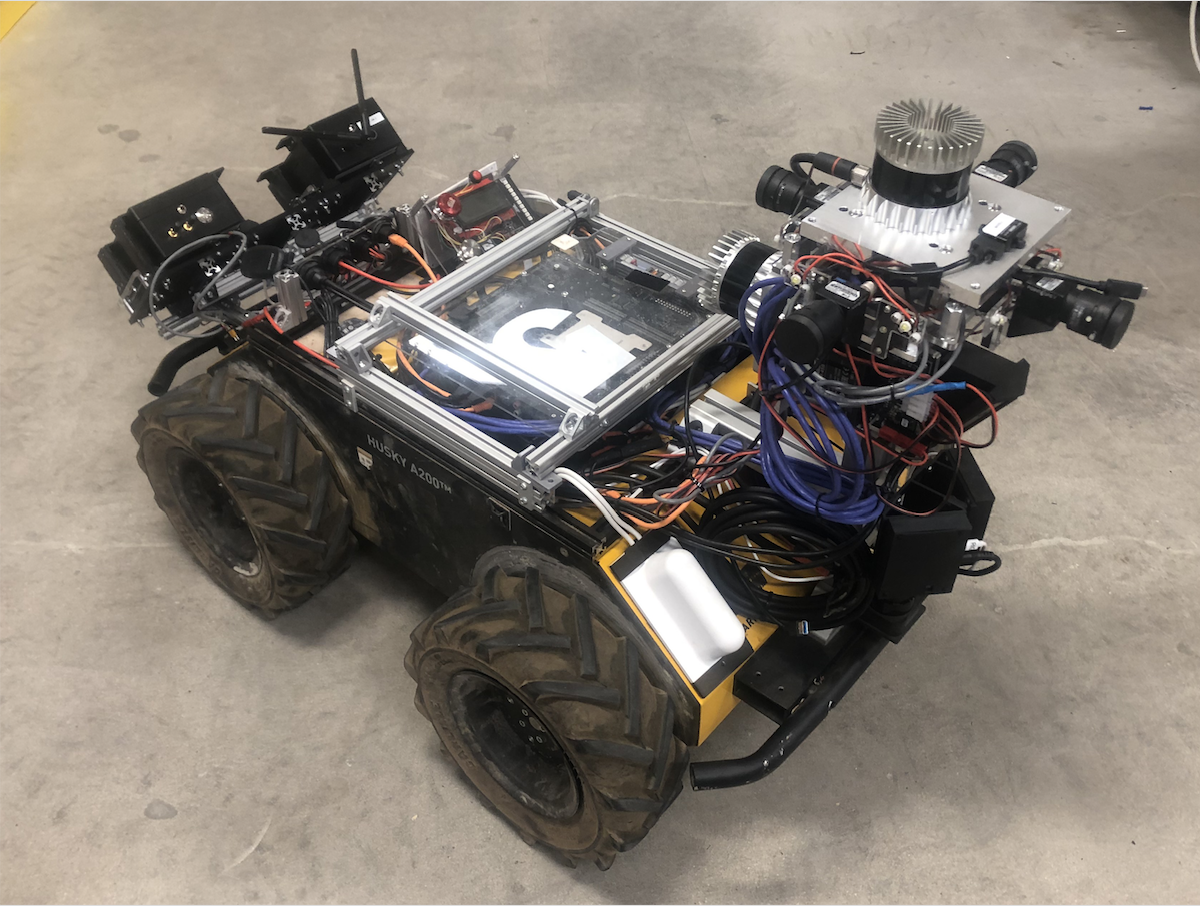}\label{fig:Husky} }}
		\subfloat[]{{\includegraphics[width=.33\textwidth]{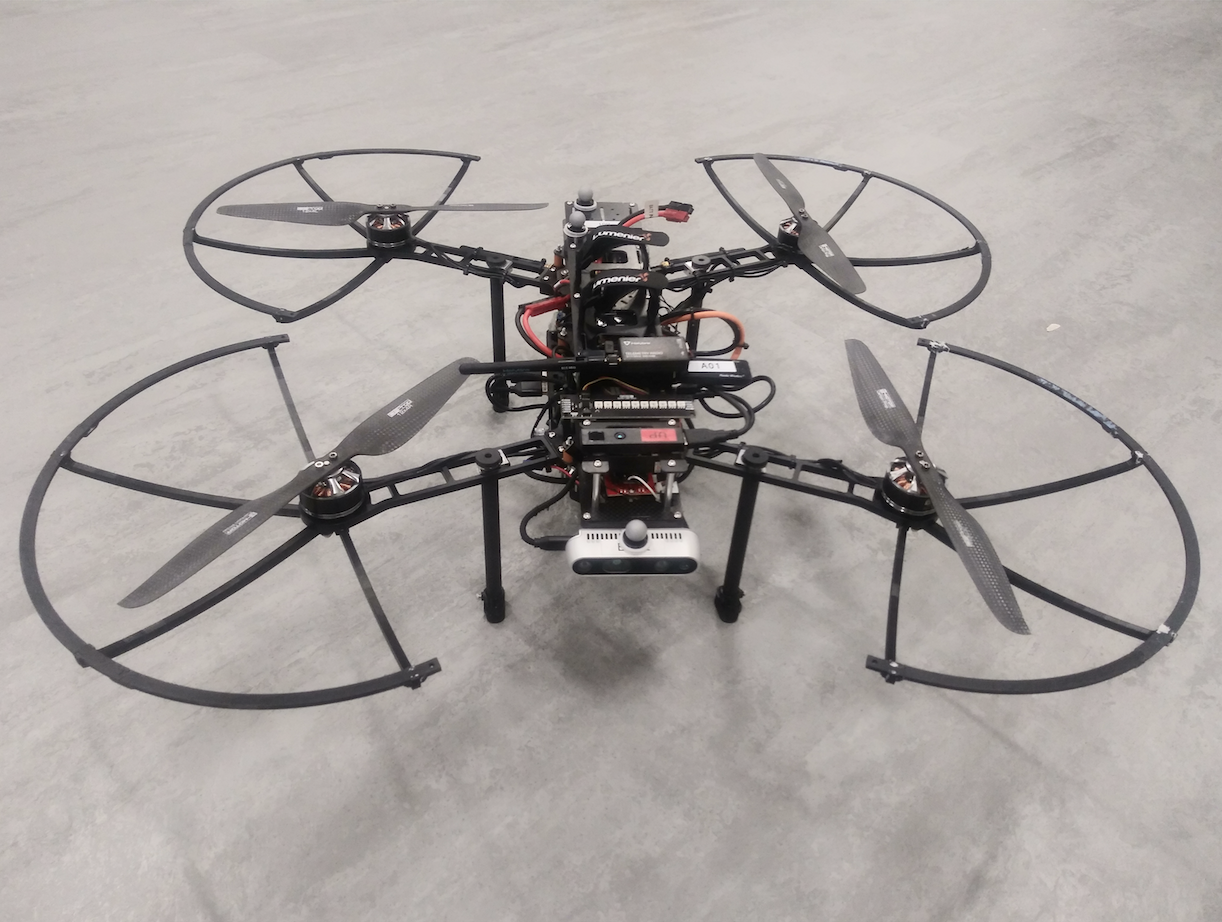}\label{figAV} }}
		\subfloat[]{{\includegraphics[width=.33\textwidth]{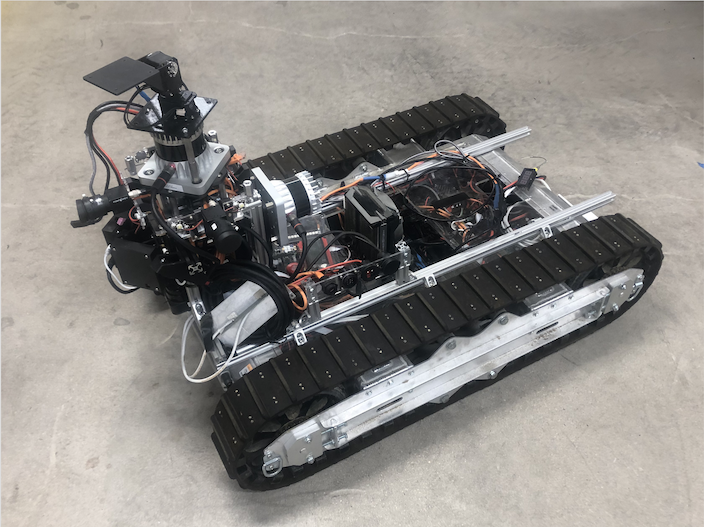}\label{fig:HD2} }}
		\caption{The robot fleet is composed of three classes of robotic agents: (a) Clearpath Husky, (b) Lumenier QAV500, and (c) Superdroid HD2.}
		\label{fig:platforms}
\end{figure}

\begin{figure}[h!]
    \centering
    \includegraphics[width=.8\linewidth]{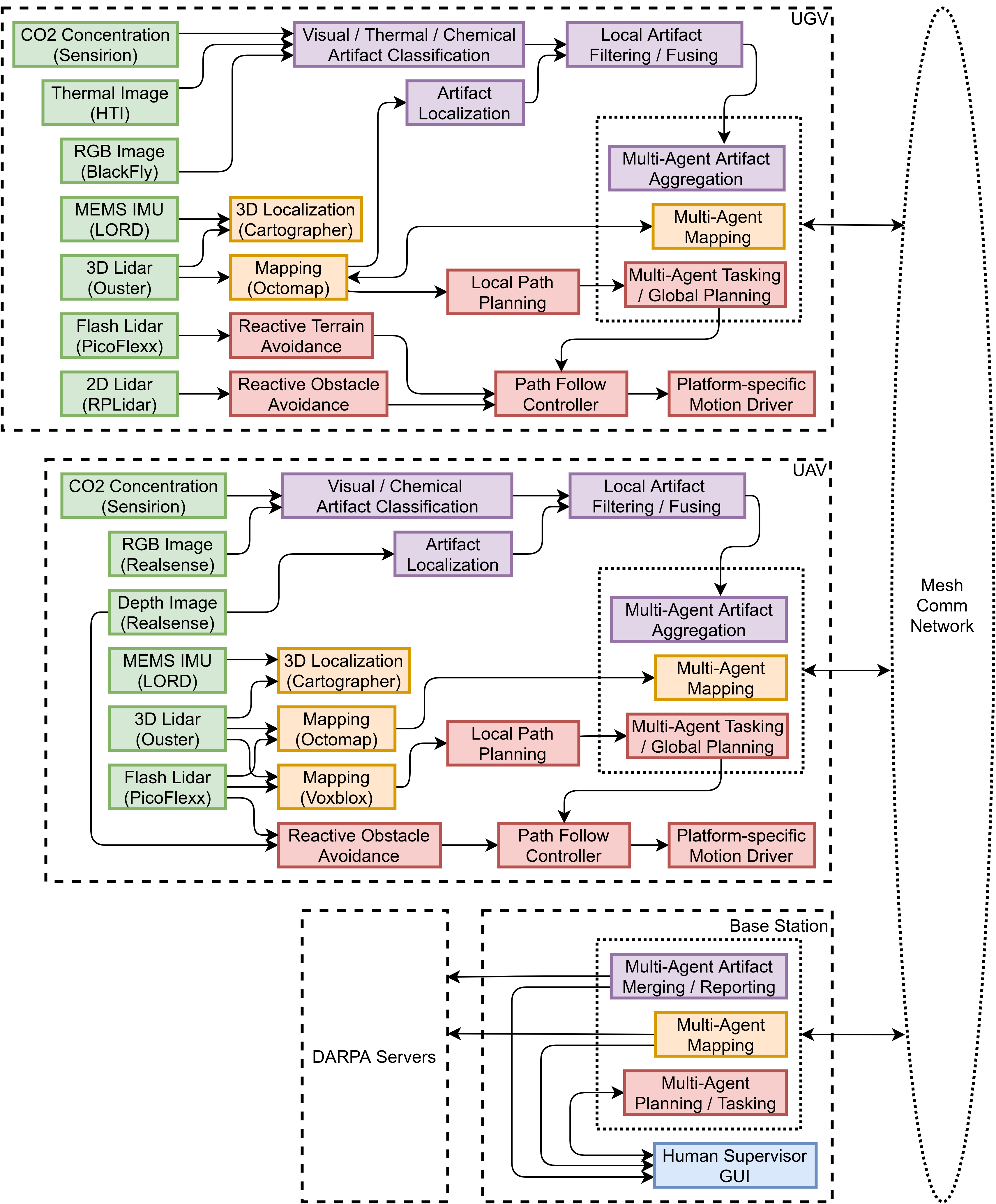}
    \caption{General overview of the software pipeline on the ground platforms, aerial platforms, and base station. Data sources are shown in green, Artifact Detection processes in purple, Localization and Mapping processes in orange, Navigation processes in red, and Operational processes in blue.}
    \label{fig:general_solution}
\end{figure}

In the following subsections, an overview of the hardware and software solutions is presented. The following terms are used interchangeably to refer to the different systems: agent, robot, platform, and vehicle. MARBLE robotic platforms are shown in Figure \ref{fig:platforms}, and a general overview of the multi-agent autonomy solution is shown in Figure \ref{fig:general_solution}.

\subsection{Platforms and Hardware}
\textbf{Ground Vehicles} - The Clearpath Husky operates as a general all-purpose platform with sufficient payload, endurance, and customizability to support the team's higher-level needs. The Husky has proven to be a capable motion base, however several aspects of the rugged underground environments, such as the introduction of stairs and increasingly rough terrain, have led to greater fleet diversification and specialization. The Superdroid HD2, a large tracked ground vehicle, was incorporated into the fleet to address the Husky's limitations. The ground vehicle platforms are equipped with a range of sensors needed to support critical processes like localization, artifact detection, and obstacle avoidance. The main sensor onboard is the Ouster OS1-64 LiDAR (Light Detection and Ranging), which generates 3D pointclouds that are used for localization, mapping, and obstacle avoidance. A LORD Microstrain 3DM-GX5-15 VRU provides inertial measurements for state estimation. RealSense D435 cameras are used to generate color and depth images for artifact detection, and are also used for object avoidance. An RPLIDAR S1 is placed at the front of the vehicle for navigation and obstacle avoidance. Special-purpose sensors, such as a carbon dioxide sensor and a Bluetooth module, have enabled agents to detect gas and cellphone artifacts during past Circuit Events. All onboard processing is handled by a 32-core AMD Ryzen CPU equipped with 128GB of RAM and 4TB of SSD storage. Artifact detection is performed on a NVIDIA GTX 1650 using 4 Gig-E cameras. To communicate with the mesh network, the ground vehicle's have their own router and high-power antenna. Each vehicle is equipped with approximately 940Wh of usable battery storage in the form of four 7S lithium-ion packs. With all motors at stall torque and the compute element fully loaded, the estimated runtime is one hour; in normal operations, the platforms can run for approximately three hours. The fully outfitted ground vehicles are shown in Figures \ref{fig:Husky} and \ref{fig:HD2}.

\textbf{Aerial Vehicles} - The UAV assembly, which was deployed at the Urban circuit event, is based on the Lumenier QAV500 airframe, and the frame shape is elongated to provide more space for onboard sensing and compute. To achieve a suitable  thrust-weight ratio, form factor, and endurance, each UAV is outfitted with Tiger MN3510 700kV motors with 12$''\times $4$''$ propellers, and is powered by a 8000mAh 4S LiPo battery. This design is rated to carry a maximum of 1600 grams per rotor. Pixracer R15, a PX4 based flight controller, is used for attitude stabilization and velocity control. To maximize UAV mission endurance, lightweight sensing and compute hardware has been selected. Similar to the ground vehicles, the aerial vehicle has an Ouster and Lord Microstrain IMU for localization. However, the aerial vehicles are configured with upward and downward-facing Camboard pico flexx cameras instead of the heavier left-facing and right-facing RealSense D435 cameras. The Intel NUC 7i7DNBE operates as the onboard computer, and is accompanied by a NVIDIA Jetson Nano for visual artifact detection purposes. Similar to the ground vehicles, the UAV is also equipped with a router for communicating with the mesh network. The UAV design is shown in Figure \ref{fig:platforms}b.

\textbf{Beacons} - The ground vehicles are outfitted with communications beacons that relay information between robots and the Base Station, as well a mechanism to deploy them. Beacons are deployed by the multi-agent robot node described in Section \ref{sec:multiagent} based on several different conditions including distance, loss of communication, junction, turn, and Human Supervisor-commanded. In addition to acting as a simple communications relay, a multi-agent node running on the beacons allow them to act as a stationary agent. As agents enter communications range of a beacon, information is collected, stored, and shared with other agents on the network. This allows the network to be more resilient to intermittent communications and facilitates the sharing of information critical to coordination. Examples of different communication beacon designs are shown in Figure \ref{fig:beacon_evolution} and examples of the latest iteration can be seen on the back of the Husky in Figure \ref{fig:Husky}.

\begin{figure}
    \centering
    \includegraphics[scale=0.2]{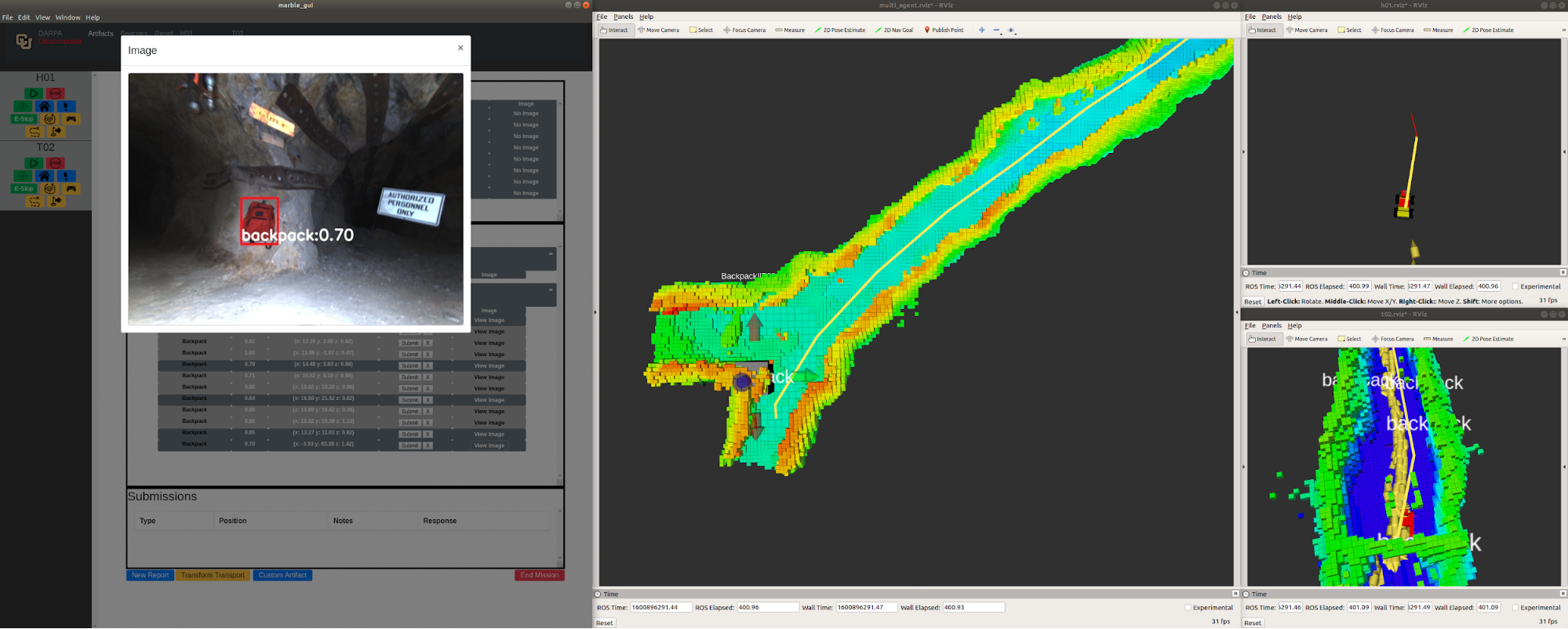}
    \caption{View of the MARBLE Base Station GUI during a competition run.}
    \label{fig:cave_base_station}
\end{figure}

\textbf{Human - Robot Interface} - The MARBLE Base Station and Graphical User Interface (GUI) provides rapid situational awareness to the Human Supervisor and acts as a direct interface to the DARPA scoring server. Through the GUI, the Human Supervisor is able to view received maps and artifact reports from agents on the network. Similar to the beacons, the Base Station runs a special version of the ROS multi-agent node to facilitate information transfer in both directions. The GUI allows high-level interactions with each robot and manages artifact reporting. A screen capture of the GUI and RViz can be seen in Figure \ref{fig:cave_base_station}. High-level commands, including emergency stop, manual goal points, return home, and beacon deployment, can be sent to agents across the network. Additional features include the ability to reset the map each agent uses for planning and a view of each robot's position, intended goal, and current planned path within the map.

\subsection{Architecture}
To succeed in the DARPA Subterranean Challenge, the agents must have a number of different capabilities. In order to score points, they must be able to detect and distinguish a variety of artifacts, localize them in the environment relative to a fixed world frame, and report this information back to the base station for final submission to DARPA.  The RGB image is processed by YOLO \cite{redmon2016you}, which produces a bounding box around the object with an estimated confidence score. A local projection for the artifact is obtained by fusing the center of the bounding box with the corresponding aligned depth measurement. The projection is then transformed into the fixed world frame using the robot's current state estimate. Over the different phases we have learned a great deal about developing a flexible object detection system which we discuss in Section \ref{sec:AD}. 

Artifact fusion is performed on each robot to aggregate individual detections from that robot. The Base Station merges incoming artifact reports from all robots into a single report. Artifacts with similar positions and types are merged prior to manual submission. The Human Supervisor is able to view artifact report images, as well as the robot's map and odometry, which enables manual corrections to the artifact location or classification before submitting the final report. Artifact reports are transmitted over a deployable WiFi mesh-network, as well as map segments, global robot positions, and other useful information for multi-agent coordination.  The Base Station does not act as a central agent, however, it can relay information between robots as any other agent can.

Critical autonomy processes like mapping, navigation, and artifact localization rely on accurate and reliable state estimation. The fleet has relied on Google's Cartographer \cite{hess2016}, which uses LiDAR scan matching for local odometry, and scan-to-submap optimization to correct for drift via global loop closures. Despite other LiDAR-inertial SLAM algorithms having greater accuracy such as LOAM \cite{zhang_low-drift_2017}, Cartographer was selected for its real-time loop closure drift correction, ease of use, and thorough documentation. 

Optimizing goal point selections and the paths to reach them is critical to exploration efficiency. Custom solutions have been developed by our team for selecting new goal points and navigating to them based on the expected topology of each circuit environment. A graph-based navigation and exploration strategy was employed for the Tunnel circuit event. The environments for Urban and Cave circuit events are more spatially complex than the tunnel environments, and so a more comprehensive frontier-based exploration algorithm was developed. These solutions are discussed in greater detail in Sections \ref{sec:metrictopo} and \ref{sec:contfront}. To map the environment, the ground robots use the open source package OctoMap \cite{hornung13octomap} to generate a probabilistic occupancy grid representation of the world. To support the needs of our multi-agent solution, several additions like map-merging were made to the OctoMap package and are discussed in Section \ref{sec:mapping}.

Some of the processes, such as localization, global planning, and artifact detection are common to both aerial and ground platforms with minor adjustments. However, the lower level subsystems are designed separately. The aerial vehicles use Voxblox \cite{oleynikova2017voxblox}, to generate a Euclidean signed distance field representation of the environment, which is useful for generating paths with sufficient clearance from obstacles. The UAVs also use upward and downward facing depth sensors in addition to the horizontal Ouster LiDAR for mapping. This setup increases the volumetric information gain in the vertical direction, which is essential in traversing shafts and stairs. The UAVs also have a different local planning solution developed exclusively for 3D navigation, and is based around planning within the depth image space. Detecting obstacles directly in the sensor frame provides immunity to mapping and localization uncertainties to a large extent, serving as an additional layer of safety while following the path. This is discussed in greater detail in Section \ref{sec:AVnav}.

\section{Map Merging}
\label{sec:mapping}

Three-dimensional maps were required for our planning solution in Section \ref{sec:contfront}, human supervisor situational awareness, and for use by DARPA.  The MARBLE mapping solution uses OctoMap to generate 3D occupancy grid representations of the environment, which is a direct input to global planning and provides situational awareness for the Human Supervisor. The octree structure of OctoMap's occupancy grids makes storing and transmitting maps more efficient than other representations such as point clouds; this efficiency is highly desirable when trying to transmit maps over wireless communication networks. While octrees are efficient data structures, transmitting entire large-volume OctoMaps through bandwidth-limited communication networks can lead to network congestion. For this reason, incremental map transmission can assist in efficient information transfer \cite{Sheng2004}. A more complete view of the environment can be generated using map data from multiple agents and merging maps from multiple robots can also reduce redundant coverage \cite{Ko2003}, \cite{Simmons2000}, \cite{Zlot2002}. In addition to obtaining map data from a direct neighbor, map data from other agents that the immediate neighbor has come into contact with can also be added to the agent's map \cite{Howard2006}. 

\begin{figure}[t!]
    \centering
    \includegraphics[scale=0.45]{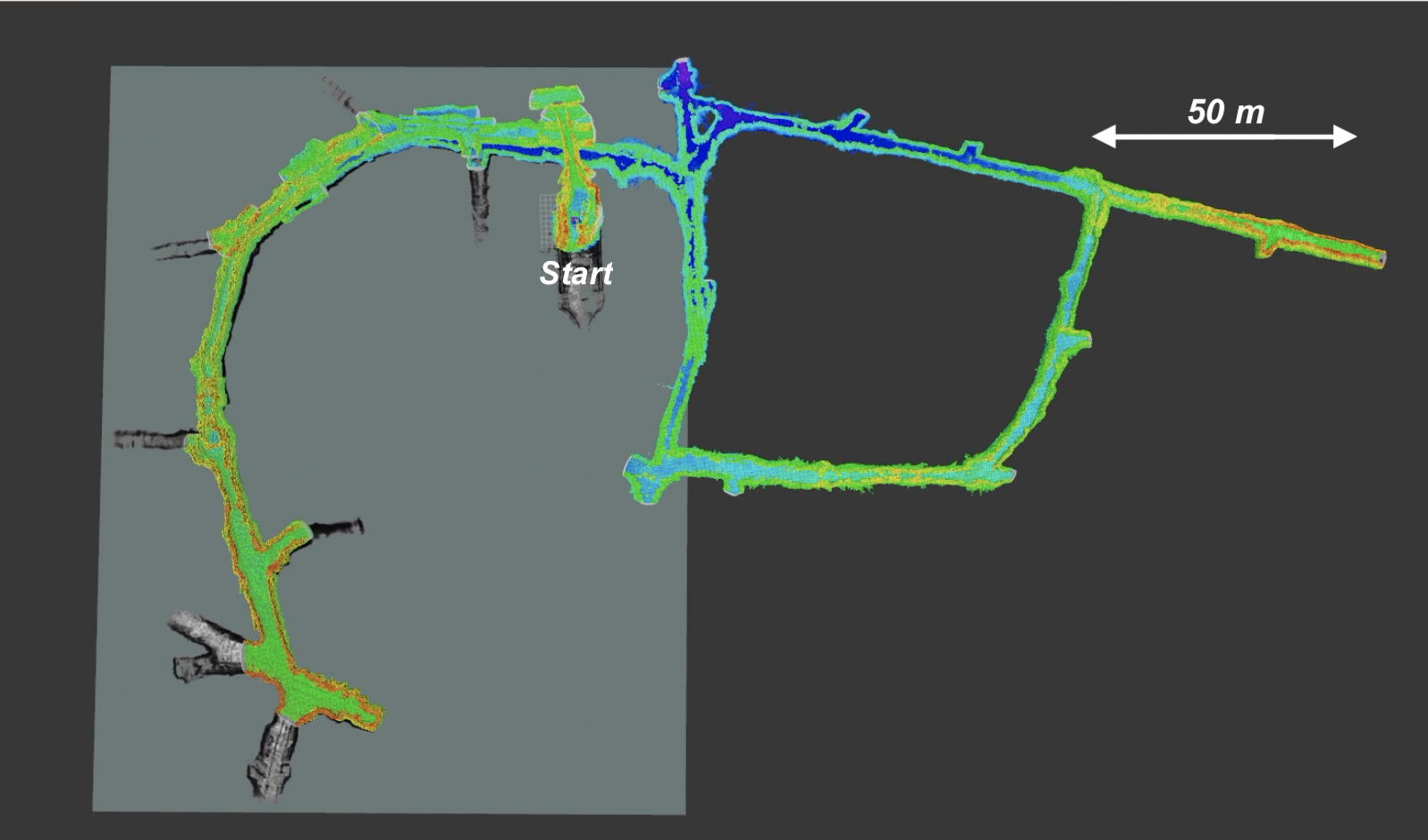}
    \caption{Example of a multi-agent merged map, from a field deployment at Edgar Experimental Mine in Idaho Springs, CO. This map was generated from a collection of diff maps from two robots.}
    \label{fig:multiagent_merged_map}
\end{figure}

Merging maps across robots is critical to our multi-agent coordination strategy, detailed in Section \ref{sec:multiagent}, but merging is not a capability in the standard OctoMap package. Following lessons learned discussed in Section \ref{sec:mapping_lessons}, we developed our own mapping package \cite{RileyFrew}, which is a fork of OctoMap Server for real-time mapping with map merging built-in. Instead of merging in maps from other robots after an agent's self map has been generated, which would require an additional iteration over an agent's ``self" map, new map sections are merged in at the same time as new sensor data. By adding the merging functionality directly to the map creation process inside of OctoMap, valuable computing resources can be conserved.  Figure \ref{fig:multiagent_merged_map} illustrates a merged map during a test deployment at Edgar Experimental Mine in Idaho Springs, CO.  The lighter gray image represents the Cartographer map generated by one of the agents who explored the left side of the mine.  Another robot traveled to the right, and transmitted its map over the communications network, creating the complete map shown. 

\begin{figure}[t!]
    \centering
    \includegraphics[scale=0.48]{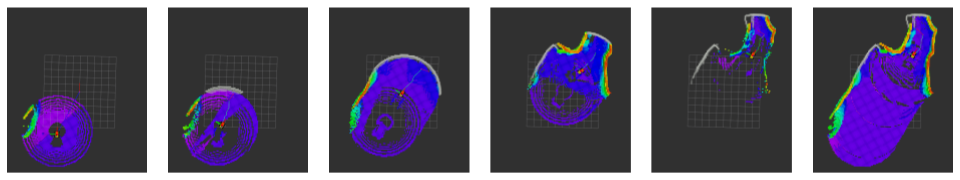}
    \caption{Sequential diff maps from left to right, with the final map on the far right constructed in real time for comparison. The diff maps can be merged to fully reconstruct the original map.}
    \label{fig:diff_maps_grid}
\end{figure}

Another desirable feature that is not provided with the stock OctoMap package is the ability to generate differences between different map sections, or ``diff maps". OctoMap generates the full Octree map structure, which can become prohibitively large to transmit for map sharing purposes. While incremental point cloud transmission is a common technique, we are not aware of another technique of breaking up OctoMaps for incremental transmission. In our modified OctoMap package, diff maps can be created at a predetermined rate, and contains all the mapping data for that time interval. The sum of an agent's diff maps make up its ``self" map. Having access to smaller sections of an agent's map allow the multi-agent node to transmit just the new diff maps over the communications network, instead of the entire map. This also benefits the map merging process, as smaller sections of the map are received for merging instead of the entire neighbor map, as seen in Figure \ref{fig:diff_maps_grid}. Another example of the benefits of transmitting map diffs is demonstrated in Figure \ref{fig:SATSOP_multi_agent}, where one agent has started to explore and is sharing map differences back to the base station. This allows the next agent to plan a path directly to a candidate goal point over sections of the map it has received from another agent.

The majority of the mapping processes occur on both the robots and the Base Station.  Both types of agents receive and transmit all diff maps for every robot, as communication allows.  Robots generate diff maps and merge map segments received from other robots and the Base Station.  The Base Station does not generate new map segments, but is able to re-transmit segments to robots that need them, and also merges the maps for use by the Human Supervisor.  The multi-agent framework discussed in Section \ref{sec:multiagent} uses a request/response method to retrieve any missing diffs.  The diffs are allowed to arrive out of order, and in the case where nodes have differing values, the latest sequence number, which is embedded in the diff, is used to ensure the data is current.

Implementing differences maps resulted in a massive bandwidth savings over repeatedly transmitting the entire map at the same intervals.  For a typical one hour mission, a robot may produce a two megabyte OctoMap, which when transmitted at 10 second intervals totals approximately 360 megabytes over the entire hour.  The same map when transmitted as a differences map typically requires more data due to the overlap that occurs when nodes change values across transmissions, and is not a fixed rate; however observation showed differences maps average approximately 10 kilobytes in size per transmission.  Therefore a one hour mission totals approximately 3.6 megabytes, or about 1/100 of the repeatedly transmitted map.
The diff map implementation also enables the removal of ``bad" map sections that have been shared to other agents. This is useful in cases when another agent shares sections of a map that are misaligned or corrupted, which can occur for a number of reasons including localization drift and sensor failures. In the case of narrow hallways, merging in bad map sections from another agent can result in an inability to plan paths effectively. Through the Base Station GUI and the multi-agent node, the Human Supervisor is able to reset a particular agent’s map on every other agent, effectively clearing bad map sections that have been shared to other agents. Using diff maps, the Human Supervisor can be relied upon to resolve mapping inconsistencies and potentially resolve otherwise mission-ending conflicts.

\section{Metric-Topological Planning and Reactive Control} 

\label{sec:metrictopo}
Subterranean environments are a challenging communications environment. It was determined that our systems would need to be able to travel outside the communication range of the base station, explore the environment, identify artifacts, and return to communications with the base station to relay artifact information back. For these reasons, a heavy focus was placed on developing robust exploration and navigation strategies that are entirely autonomous and can operate outside of communications range of the Base Station. Some exploration approaches represent the environment through topological graphs \cite{Dudek1991}. It has been shown that graph-based exploration is useful in tunnels \cite{Thrun2004}, indoor rooms \cite{Bormann2016}, and maze-like indoor environments \cite{Oleynikova2018}. After a graph is built, it can be used to efficiently navigate from one node to another \cite{Brass2011}. The centering algorithm used to traverse edges of the graph is bio-inspired. Insects rely on patterns of optic flow to generate a centering flight response \cite{Srinivasan1998RobotNI}, which was tested in \cite{SantosVictor1997} and \cite{Griffiths2006MaximizingMA}, whereas our systems rely on LiDAR depth measurements. Insects also rely on optic flow for small-obstacle detection and avoidance \cite{Escobar2019}, and which has been demonstrated as a viable obstacle detection method onboard multi-rotor platforms \cite{Ohradzansky2018}.

\begin{figure}[t!]
    \centering
    \includegraphics[width=\textwidth]{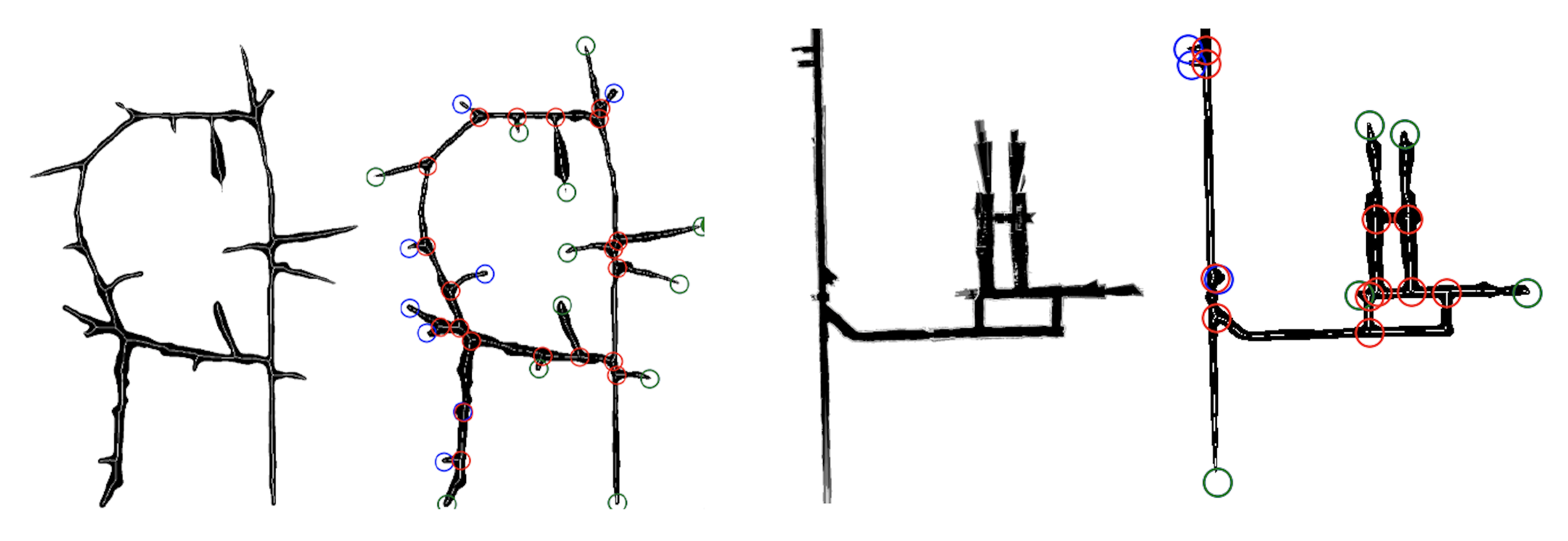}
    \caption{Examples of the graph conversion from a blurred occupancy grid image. The two images on the left are from the Edgar Experimental Mine in Idaho Springs, CO, and the two images on the right are from NIOSH mine during the Tunnel circuit event. In both sets, the left image is the raw binary occupancy grid generated by the robot’s perception system and the right image is the resultant graph estimate. Edges are shown in white and vertices are labeled with circles. Red circles are multi-edged vertices, blue circles are single-edged vertices, and green circles are the ends of unexplored edges.} \label{fig_plan_graph}
\end{figure}

For the Tunnel circuit event, our team deployed a fleet of ground vehicles in a largely 2D environment. The planning strategy assumed a 2D environment that could be approximated accurately with a graph (vertices and edges), and that the graph could be traversed using the reactive centering controller on edges. Tunnel environments fit reasonably well within this assumption with a few exceptions in larger open areas. The approach relied on a binary image thinning algorithm which we performed over a Gaussian-blurred occupancy grid image \cite{zhang1984fast}. Then, an image convolution is performed on the skeletonized map to detect vertices and edges. The results of the approach can be seen in Figure \ref{fig_plan_graph} which shows sample graphs from Edgar Experimental and NIOSH mines. Vertices are labeled differently if they contain 1 or 3 or more edges or if they are an unexplored edge. Estimating a graph using simple image processing proved quite effective at the Tunnel circuit event.

To traverse the graph, the vehicles relied on a reactive centering controller operating on depth scans from the front mounted RPLIDAR. Estimates of environment relative states, such as the lateral and angular displacement of the vehicle from the centerline of a hallway, are extracted from the depth scans using bio-inspired wide-field sensor integration methods. The estimates are used as feedback to generate steering commands that drive the vehicle to the center of corridor like structures. An additional steering controller is integrated with the centering controller for reactive small obstacle avoidance. At vertices, a simple proportional steering controller was used to travel to the center of each vertex, turn to face the desired new edge, and resume centering to the next vertex.

\begin{figure}[t!]
    \centering
    \includegraphics[width=\textwidth]{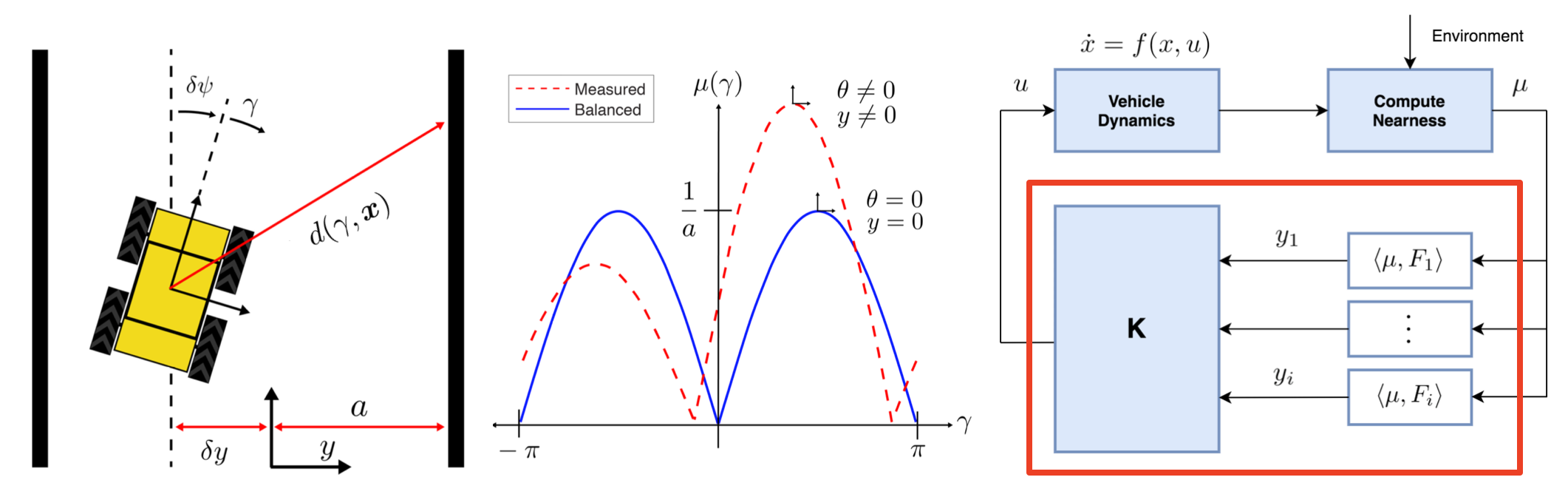}
    \caption{Overview of the centering controller algorithm. On the left is a top-down view of a ground vehicle in a hallway. The plots in the middle show examples of different nearness signals. A data flow for the controller is shown on the right side of the figure.} \label{fig:centeringcontroller}
\end{figure}

An overview of the reactive centering controller is presented in Figure \ref{fig:centeringcontroller}. On the left side of the figure, a top down view of a ground robot in a hallway environment is shown. The robot has both a lateral displacement $\delta y$ and an angular displacement $\delta \psi$ relative to the center-line of the hallway. These are the states we are interested in regulating for the purpose of centering. The robot is equipped with a 2D depth scan sensor that measures depth $d$ as a function of the viewing angle $\gamma$ and robot state $\boldsymbol{x} = \begin{bmatrix} \delta y& \delta \psi \end{bmatrix}$. Measured depth is hypothetically unbounded, and so the depth scans are converted to nearness $\mu = \frac{1}{d}$, which is a bounded signal. Examples of nearness scans are shown in the plot in the middle of Figure \ref{fig:centeringcontroller}, where the red signal represents the measured nearness of a robot with both lateral and angular displacements and the blue signal represents the measured nearness of a robot that is perfectly centered in a hallway ($\delta y = \delta \psi = 0$). Nearness scans are useful in the case of centering because they encode the hallway relative states $\delta y$ and $\delta \psi$. Equation \ref{eq:planarnear} represents analytic nearness in an infinite hallway as a function of the hallway relative vehicle states $\delta y$ and $\delta \psi$ and the hallway half-width $a$.

\begin{equation}
\mu(\gamma, \boldsymbol{x}) =     \begin{cases}
      -\frac{\sin(\gamma+\delta \psi)}{a + \delta y} & \hspace{-.075cm} \ -\pi \leq \gamma + \delta \psi < 0 \vspace{.15cm}\\ 
      \frac{\sin(\gamma+\delta \psi)}{a + \delta y} & \hspace{.25cm} \ 0 \leq \gamma + \delta \psi \leq \pi \\
    \end{cases}
    \label{eq:planarnear}
\end{equation}

By projecting the measured nearness signal onto different predetermined shapes, estimates of the hallway relative vehicle states $\delta y$ and $\delta \psi$ can be generated and used to produce the desired steering response. This process is shown in the right side of Figure \ref{fig:centeringcontroller} where the measured nearness signal is projected onto different shapes $F_{i}$. The scalar outputs of the projections, which represent the environment relative states, are pooled through the gain matrix $K$ to generate the different control signals. The generation of the steering command using the spacial inner product is detailed in Equation \ref{eq:inner_product}, and the weighting shape that achieves a centering response $F_{u_{\dot{\psi}}}$ is given in Equation \ref{eq:steering_shape}. In Equations \ref{eq:inner_product} and \ref{eq:steering_shape}, scalar gains $k_1$ and $k_2$ are used to adjust the responsiveness of the steering controller to lateral and angular displacement respectively. 

\begin{align}
\centering
u_{\dot{\psi}} = \langle \mu, F_{u_{\dot{\psi}}} \rangle &=  \int^{\pi}_{\text{-}\pi}\mu(\gamma,\boldsymbol{x})\cdot F_{u_{\dot{\psi}}}(\gamma) \hspace{.1cm} d \gamma =  k_1 b_1 + k_2 b_2 \label{eq:inner_product}\\
F_{u_{\dot{\psi}}} &= k_1 \sin(\gamma) + k_2 \sin(2\gamma) \label{eq:steering_shape}
\end{align}

\begin{figure}[b!]
    \centering
    \includegraphics[width=\textwidth]{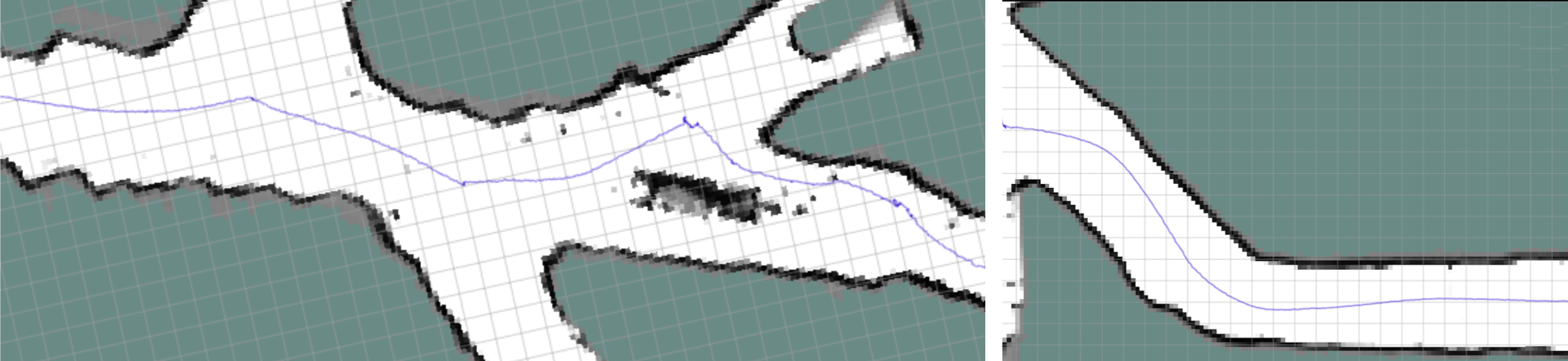}
    \caption{Example trajectories from testing in Edgar Experimental Mine (left) and a small section of Tunnel circuit event at NIOSH Mine (right). The centering controller guides the vehicle to nodes on the graph.} \label{fig:centerpaths}
\end{figure}

This centering algorithm is well suited for environments that resemble hallways, but it is also able to achieve a centering response in non-uniform environments as seen in Figure \ref{fig:centerpaths}. Due to integrating hundreds or thousands of sensor measurements, the generated control commands are robust to noise in the measured nearness signal. The algorithm is also robust to small-obstacles in the measured environment; small-obstacles, which produce perturbations in the measured nearness signal, do not cause large deviations from the center of the environment. As a result, a separate small-obstacle detection and avoidance algorithm is used in combination with the centering controller to ensure the system avoids obstacles of all sizes while navigating the graph. The relative distance and angular position of small obstacles in the environment are extracted from the depth scan through inhibition of the measured depth scan with a blurred, low-frequency reconstruction of the scan. The small-obstacle detection and avoidance process is presented in greater detail in \cite{Ohradzansky2018}, and the complete navigation solution for the Tunnel circuit event is detailed in 
\cite{OhradzanskyMills}.

For determining the next best edge to explore, the team developed different exploration strategies. Initially, a Silent Explore strategy was developed in which the different systems were completely independent and did not share information. This exploration strategy was designed to keep the vehicle moving quickly through the environment while it searches for artifacts. Once an artifact is detected, it switches its current mission mode from ``Explore" to ``Report", and a path is planned back to the base station to relay the artifact information back. The second strategy was Greedy Explore, which requires a more robust communications network. This allowed the robots to explore deeper into the environment without returning all the way to the base station to report artifacts and maps. This was implemented primarily through the communications beacons detailed in Section \ref{sec:comms}, to increase the range of network, but they were not required to partially implement this strategy, as the robots themselves could also act as a relay. In this strategy, like the previous, when an artifact was detected, the robot entered Report mode and began a path toward the base station. However, once a robot received acknowledgement from the base station that an artifact had been reported, the robot returned to Explore mode and continued to explore the environment until the next artifact was found.  The Greedy Explore method was used through all phases of the competition, although communications beacons were not deployed during the initial tunnel event.

\begin{table}[h!]
\begin{center}
 \begin{tabular}{c c | c c c}
 \hline
 \textbf{Run} & \textbf{Course} & \textbf{H01} & \textbf{H02} & \textbf{H03}\\ 
 \hline
 1 & SR & - & - & - \\ 
 2 & EX & 150 & 840 & 450 \\
 3 & EX & - & 1595 & 953 \\
 4 & SR & - & 420 & 445 \\
 \hline
\end{tabular}
\caption{\label{tab:tunnelresults} Linear distance traveled in meters, of MARBLE's three-Husky fleet, during the NIOSH Experimental Mine (EX) and Safety Research Mine (SR) deployments at Tunnel circuit event.}
\end{center}
\end{table}

\begin{figure}[t!]
    \centering
    \includegraphics[scale=0.25]{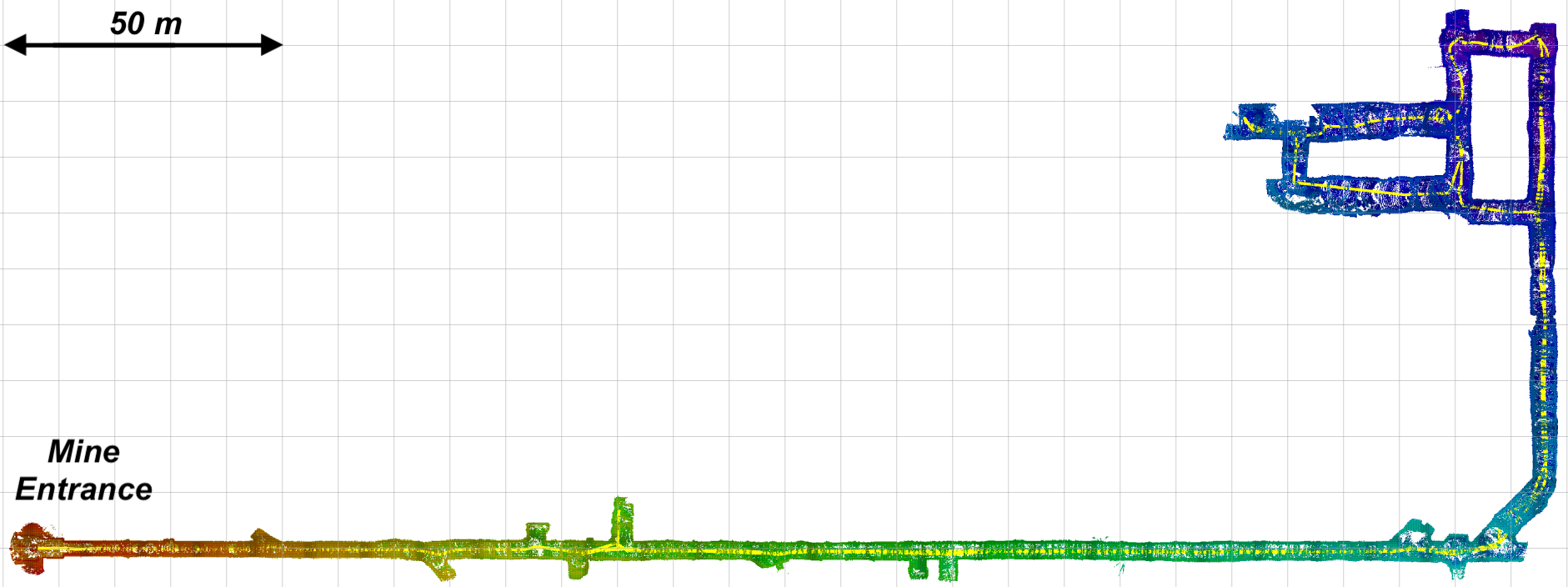}
    \caption{UGV agent trajectory (yellow) and OctoMap (rainbow) during autonomous exploration mission at NIOSH Experimental Mine in Bruceton, PA.} \label{fig:NIOSH_map_traj}
\end{figure}

Experimental results from the Safety Research (SR) and Experimental (EX) section of NIOSH Mine are presented in Table \ref{tab:tunnelresults}. We had difficulty getting our systems localized against DARPA's gate during the first run, and we were unable to deploy any vehicles in the time limit. During runs where we were able to get our systems out of the gate, we averaged just under $700\text{m}$ of linear distance traveled per vehicle. Figure \ref{fig:NIOSH_map_traj} shows a run in the Experimental section where the system covered 1.6 linear kilometers in under an hour. Figure \ref{fig_plan_graph} shows the graph structure generated from a single agent during another Tunnel competition run, as well as a deployment at Edgar Experimental Mine in Idaho Springs, CO. We had not developed a multi-agent coordination strategy fully at the time, and so our agents often cover the same areas. A similar graph-based exploration approach was used by Team CERBERUS, and the results from larger scale experiments inside different subterranean environments, including Tunnel circuit, are presented in \cite{Dang2020}. In their ground vehicle tests, their systems are able to cover hundreds of meters in one hour deployments. Team EXPLORER deployed a graph-based exploration strategy at Tunnel circuit, the results of which are summarized in \cite{Miller2020}. While Team CoSTAR used a frontier-based exploration strategy at Tunnel circuit, they were able to cover an impressive amount of the course in the time limit \cite{Otsu2020}. For our graph-based approach to perform robustly in more complex 3D environments, a more generic graph-building process could be adopted as a part of the team's future work.

\section{Continuous Frontier View Planning}
\label{sec:contfront}
The graph-based planning approach presented in Section \ref{sec:metrictopo} assumes that the environment is readily approximated by a graph. For more general 3D environments, previous works have looked at representing the environment using occupancy grids \cite{hornung13octomap} and signed distance fields \cite{oleynikova2017voxblox}.  Given a continuous map representation, a robot can explore an unknown environment by consistently navigating towards the frontier, or the threshold between the observed and unobserved \cite{Yamauchi1999}.  Although this can be effective in 2D, the frontier is often sparse in 3D environments because of gaps in depth sensor fields of view and noise in the observed map.  As a result, modern exploration approaches focus on exploring the environment by selecting trajectories or goal poses that maximize information added to the map \cite{palazzolo2017information} \cite{charrow2015information}.  This is often accomplished by considering an outward expanding network of motion primitives \cite{dharmadhikari2020motion} or trajectories generated through an RRT* \cite{bircher2016receding} \cite{schmid2020efficient}.  These approaches generate efficient exploring behaviors when the edge of the map is in close proximity to the robot, but they can struggle to break out of local planning minima when reaching dead-ends in larger environments.

The MARBLE 3D exploration solution is an integrated planning approach that considers the map information of frontier-facing goal poses with a rapid cost-to-go calculation using fast marching methods that works on both air and ground vehicles.  By limiting the sampling domain to frontier-relative goal poses, the robot's attention is constantly focused on the edge of the environment resulting in navigation away from dead-ends and local information minima. Although frontiers can be sparse in 3D, we utilize a two filtering methods to focus on frontiers that are connected to other frontiers. The frontier view planning strategy takes as input a map, $\mathcal{M}$, of the environment and the current vehicle pose, $\textbf{x}$, and outputs a utility table. The map, $\mathcal{M}$, is a 3D voxel representation of the environment where each voxel contains an occupancy probability and Euclidean signed distance to the closest obstacle. This utility table is a list of goal poses, their utilities, and paths from the current robot pose to each goal pose.  Global planning is split into two phases: (1) frontier view sampling, where frontiers are filtered and goal poses are sampled relative to the frontier and (2) utility optimization, where the utility of each sampled goal pose is calculated by dividing the information gain of each goal pose by a multi-goal cost-to-go calculation.

\subsection{Planning Preliminaries}
To understand the approach taken in this section, it's important to establish some preliminary definitions.  The map $\mathcal{M}$ is a set of voxelized grid cells $\bm{c}$, defined as 
\begin{equation}
    \mathcal{M} = \{ \bm{c} : \bm{c} \in \mathbb{Z}^3,
    \bm{c} \succeq \bm{0},
    \bm{c} \preceq \bm{c}_{\text{max}.}
    \}
\end{equation}
where $\mathbb{Z}$ is the set of all integers and $\bm{c}_{\text{max}}$ is an upper bound on the map dimensions.  The map, $\mathcal{M}$, can be separated into two subsets, the seen space, $\mathcal{S}$, and the unseen space, $\mathcal{U}$.  Within the seen space, voxels are either free, $\bm{c} \in \mathcal{S}_f$, or occupied, $\bm{c} \in \mathcal{S}_o$.  More formally, these subsets are defined below

\begin{equation}
\begin{aligned}
    \mathcal{M} &= \mathcal{S} \cup \mathcal{U} \\
    \mathcal{S} &= \mathcal{S}_f \cup \mathcal{S}_o \\
    \mathcal{S}_o &= \{\bm{c} : \bm{c} \in \mathcal{M}, O(\bm{c}) \geq p_{occ} \} \\
    \mathcal{S}_f &= \{\bm{c} : \bm{c} \in \mathcal{M}, O(\bm{c}) \leq p_{free} \} \\
    \mathcal{U} &= \{\bm{c} : \bm{c} \in \mathcal{M}, p_{\text{free}} < O(\bm{c}) < p_\text{occ} \}
\end{aligned}
\end{equation}

where $O(\bm{c}) \in [0,1]$ is the probability that the voxel cell $\bm{c}$ is occupied, and $p_\text{free}$ and $p_\text{occ}$ are the thresholds for considering a voxel free or occupied respectively.  In addition to occupancy probability, the distance to the closest occupied voxel $D(\bm{c})$ is defined at each map voxel.  Occupancy probability at each map voxel is stored in the merged octomap data structure maintained by our mapping solution and the distance transform of the occupancy grid is stored and updated incrementally by running an efficient Euclidean distance transform function at regular map update intervals \cite{zhao2007parallel}.

The set of traversable voxels, $\mathcal{T}$, are the set of voxels containing valid robot poses.  For a quadrotor this is defined as
\begin{equation}
    \mathcal{T}_\text{quad} = \{ \bm{c} : \bm{c} \in \mathcal{S}_f, D(\bm{c}) \geq r_\text{safe} \}
\end{equation}
where $r_\text{safe}$ is some safety radius.  The safety radius should be a bit larger than the radius of a sphere containing the quadrotor.  For a ground vehicle, free voxels with a bottom neighbor voxel, $N(\bm{c})_\text{bottom}$, that is a member of the ground voxel set, $\mathcal{G}$, are considered traversable.  A ground voxel is an occupied voxel whose local surface normal vector's z-component, $\text{Normal}(\bm{c})_z$, is greater than or equal to some threshold, $n_{z,\text{ground}}$.  A formal definition is shown below. 
\begin{equation}
\begin{aligned}
    \mathcal{T}_\text{ground} &= \{ \bm{c} : \bm{c} \in \mathcal{S}_f, N(\bm{c})_\text{bottom} \in \mathcal{G} \} \\
    \mathcal{G} &= \{ \bm{c} : \bm{c} \in \mathcal{S}_o, \text{Normal}(\bm{c})_z \geq n_{z,\text{ground}} \}
\end{aligned}
\end{equation}
This restricts the ground vehicle's planned motion to surfaces below a certain inclination.  When computing the Euclidean distance transform for ground vehicle planning, we ignore voxels in the ground voxel set so that $D(\bm{c})$ is the distance from the nearest occupied voxel that is not in the ground voxel set, $\{\mathcal{S}_o \setminus \mathcal{G}\}$.

\subsection{Goal Pose Sampling}
The first phase of the global planning algorithm is goal pose sampling.  Goal poses are sampled relative to the frontier, $\mathcal{F},$ which are free map voxels adjacent to an unseen voxel.  This is defined below as
\begin{equation}
\label{eq_frontier_definition2}
    \mathcal{F} = \{\bm{c} : \bm{c} \in \mathcal{S}_f \text{ and } (0<\sum_{\bm{n} \in N(\bm{c})} \mathbbm{1}_{\mathcal{U}}(\bm{n}))\}
\end{equation}
where $\mathcal{S}_f$ is the set of map voxels that are seen and free, $\mathcal{U}$ is the set of unseen map voxels, $N(\bm{c})$ are the neighbors of voxel $\bm{c}$, and  $\mathbbm{1}_{\mathcal{U}}$ is the indicator function for membership in the unseen voxel set.  Frontier voxels are labeled by the FindFrontierVoxels$(\mathcal{M})$ method on line 2 of Algorithm \ref{algorithm:FrontierPlanning}.  In 2D this is often sufficient for finding potential robot goal points, however in 3D the raw frontier voxels are often spurious and can lead to observing little unseen map volume. For this reason, our approach filters the frontier voxels before sampling goal poses in a two step process. Due to the way our 3D LiDAR sensors are mounted on our platforms, it can be difficult to observe frontier voxels along the ceiling or negative space near the ground plane. Frontiers whose local plane fits have a large normal z-component can often be difficult to view with a sensor field of view aligned with the x-y plane and lead to sampling goal poses with small information gains, and so they are filtered out. Similarly, frontier points that are not part of a larger contiguous cluster often produce goal poses with little map information, and so voxels are filtered to eliminate spurious frontier voxels that are not part of a larger cluster.  Frontiers are filtered by the FilterFrontierNormals and FilterFrontierContiguous methods on lines 3 and 4 of  Algorithm \ref{algorithm:FrontierPlanning}.  $r_{normal}$ is the frontier plane fitting radius, $[n_{z,\text{min}}, n_{z,\text{max}}]$ is the range of acceptable z-components for the local frontier plane normal vector, and $N_{c,\text{min}}$ is the minimum number of contiguous frontier voxels to form a cluster.

\begin{algorithm}[t!]
\begin{algorithmic}[1]
  \STATE path$ \gets \{\}$ \\
  \STATE $\mathcal{F} \gets $FindFrontierVoxels$(\mathcal{M})$
  \STATE $\mathcal{F} \gets$FilterFrontierNormals$(\mathcal{F}, r_{normal}, n_{z,\text{min}}, n_{z,\text{max}})$ \\
  \STATE $\mathcal{F} \gets$FilterFrontierContiguous$(\mathcal{F},N_{c,\text{min}})$\\
  \STATE $\mathcal{P} \gets $SampleGoalPoses$(\mathcal{F},r_{group})$
  \STATE $T \gets $ComputeCostToGo$(\mathcal{M},c(\textbf{x}))$ \\
  \STATE $\textbf{p}_{goal} \gets \max_{\textbf{p} \in \mathcal{P}} U(\textbf{p}, \textbf{x})$ \\
  \STATE path$ \gets $FollowGradient$(c(\textbf{p}_{goal}), c(\textbf{x}))$ \\
\end{algorithmic}
\caption{Frontier View Planner$(\mathcal{M},\textbf{x}, r_{normal}, n_{z,max}, n_{z,max}, N_{c,min}, r_{contig}, r_{group})$}
\label{algorithm:FrontierPlanning}
\end{algorithm}

\begin{figure}[b!]
    \centering
    \includegraphics[width=0.75\textwidth]{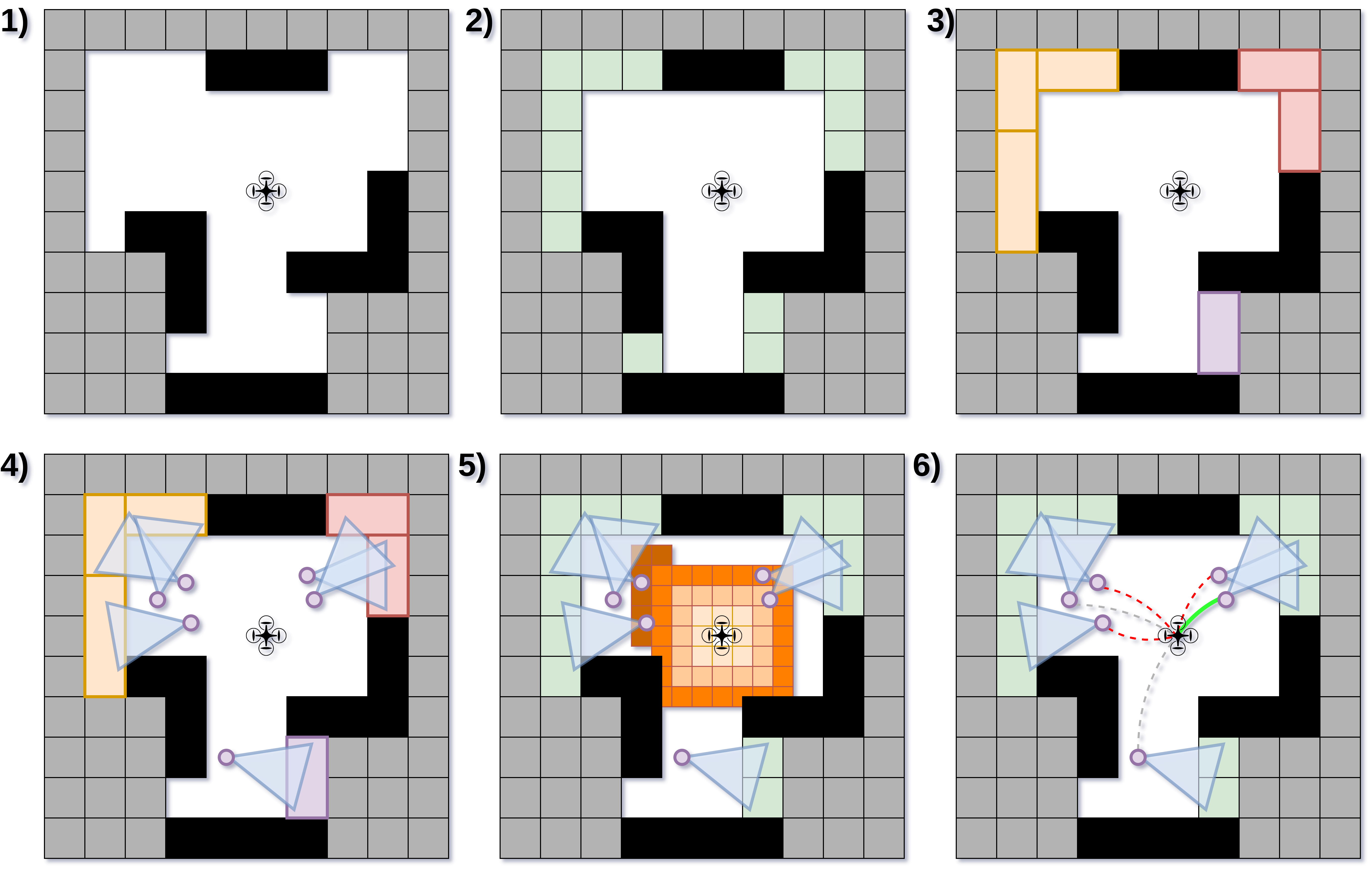}
    \caption{A 2D drawing of the frontier view pose sampling and planning approach.  1) The robot builds an occupancy grid of the environment.  White cells are free, black cells are occupied and gray are unseen.  2) Frontiers are labeled in green as free cells adjacent to unseen.  3) Frontiers are clustered based on contiguity (orange, red, and purple) and filtered if the clusters don't contain a minimum number of cells (2 cells in this example).  Clusters are further divided into groups, denoted by the colored lines, using a greedy sample-based grouping algorithm.  4) Groups are used to sample informative goal poses (purple circles) relative to the each group's centroid.  Information gains are calculated based on cells in the sensor field of view at each goal pose.  5) A Fast Marching Method is used to calculate the cost-to-go to each goal pose and the gradient of the wavefront is used to generate a path to each pose that the front reaches.  6) A list of the best n goal poses is published where n is the number of robots within comms.  The green path is selected in this n=1 example.} \label{fig_plan_6step}
\end{figure}

After filtering the frontier voxels in the map, the planner samples candidate goal poses relative to those frontier voxels.  This procedure is detailed in Algorithm \ref{algorithm:ViewPoseSampling}.  The goal sampling algorithm starts with a greedy grouping procedure where frontier voxels are randomly sampled from the ungrouped set, $\{\mathcal{F}$ \textbackslash Groups$\}$, and grouped with their frontier neighbors within a radius of $r_\text{group}$ that have not yet been added to the grouped set.  Once every frontier voxel is grouped, candidate goal poses are sampled relative to each group centroid, $\bar{\bm{c}}_g$, until an admissible and non-occluded goal pose, $\bm{p}_\text{sample}$, is found for each group.  Admissible poses have a position within a traversable voxel.  Poses are sampled uniformly in polar coordinates relative to the corresponding group centroid. The sampling radius, azimuth, and elevation ranges are bounded such that the group centroid is within the sensor field of view of the robot at the sampled pose.  A 2D example is shown in steps 3) and 4) of Figure \ref{fig_plan_6step}.  The colored rectangles in step 3 are grouped frontiers and the purple circles in step 4) are the sampled poses relative to those groups. This grouping and pose-sampling procedure was derived and adapted from an approach used by underwater surface inspection robots \cite{englot2013three}.

\begin{algorithm}[tbh]
\begin{multicols}{2}
\begin{algorithmic}[1]
  \STATE $\mathcal{P} \gets \{\}$, Groups $\gets \{\}$ \\
  \WHILE{$\{\mathcal{F}$ \textbackslash Groups$\} \neq \{\}$}
    \STATE $\bm{c}_g \gets$ Sample$(\{\mathcal{F}$ \textbackslash Groups$\})$ \\
    \STATE $G_{\text{new}} \gets \{\}$ \\
    \FOR{$\bm{c} \in$ GetFrontierNeighbors$(\bm{c}_g, r_\text{group})$}
      \IF{$\bm{c} \notin$ Groups}
        \STATE $G_{\text{new}} \gets [G_{\text{new}}; \bm{c}]$ \\
      \ENDIF
    \ENDFOR
    \STATE Groups $\gets [\text{Groups}; G_{\text{new}}]$ \\
  \ENDWHILE
  \FOR{$G \in$ Groups}
    \WHILE{True}
      \STATE $\bm{p}_{\text{sample}} \gets$ SamplePose$(\bar{\bm{c}}_g, \text{FoV})$
      \IF {CheckAdmissible$(\bm{p}_{\text{sample}})$}
        \IF{!CheckOcclusion$(\bm{p}_{\text{sample}},\bar{\bm{c}}_g)$}
          \STATE $\mathcal{P} \gets [\mathcal{P}; \bm{p}_{\text{sample}}]$ \\
          \STATE \textbf{Break} \\
        \ENDIF
      \ENDIF{}
    \ENDWHILE
  \ENDFOR
\end{algorithmic}
\end{multicols}
\caption{SampleGoalPoses$(\mathcal{F}, r_\text{group})$}
\label{algorithm:ViewPoseSampling}
\end{algorithm}

\subsection{Utility Optimization and Cost Evaluation}
After candidate goal pose destinations are sampled, the planner evaluates the poses using a simple mapping rate utility function defined below
\begin{equation}
    U(\bm{p}, \bm{x}) = \frac{\text{FrontierInFoV}(\bm{p})}{\text{Cost}(\bm{p}, \bm{x})}
\end{equation}
where $\bm{p}$ is a candidate goal pose, $\bm{x}$ is the current robot pose, FrontierInFoV$(\bm{p})$ is the number of frontier voxels visible from the candidate goal pose, and Cost$(\bm{p}, \bm{x})$ is the approximate travel time to get from the current goal pose to the candidate goal pose.  This is required to select the best utility among the candidate goal poses (line 7 in Algorithm \ref{algorithm:FrontierPlanning}).  Other approaches have looked at information-theoretic utility functions that consider the entropy reduction of a particular path or pose.  These approaches consider the value in obtaining more accurate occupancy estimates of the seen voxels in the map.  However, for the purpose of exploring the environment as quickly as possible we assume the only valuable voxels to see within the map are those adjacent to unseen portions of the environment.

\begin{figure}[t!]
		\centering
		\subfloat[]{{\includegraphics[width=.48\textwidth]{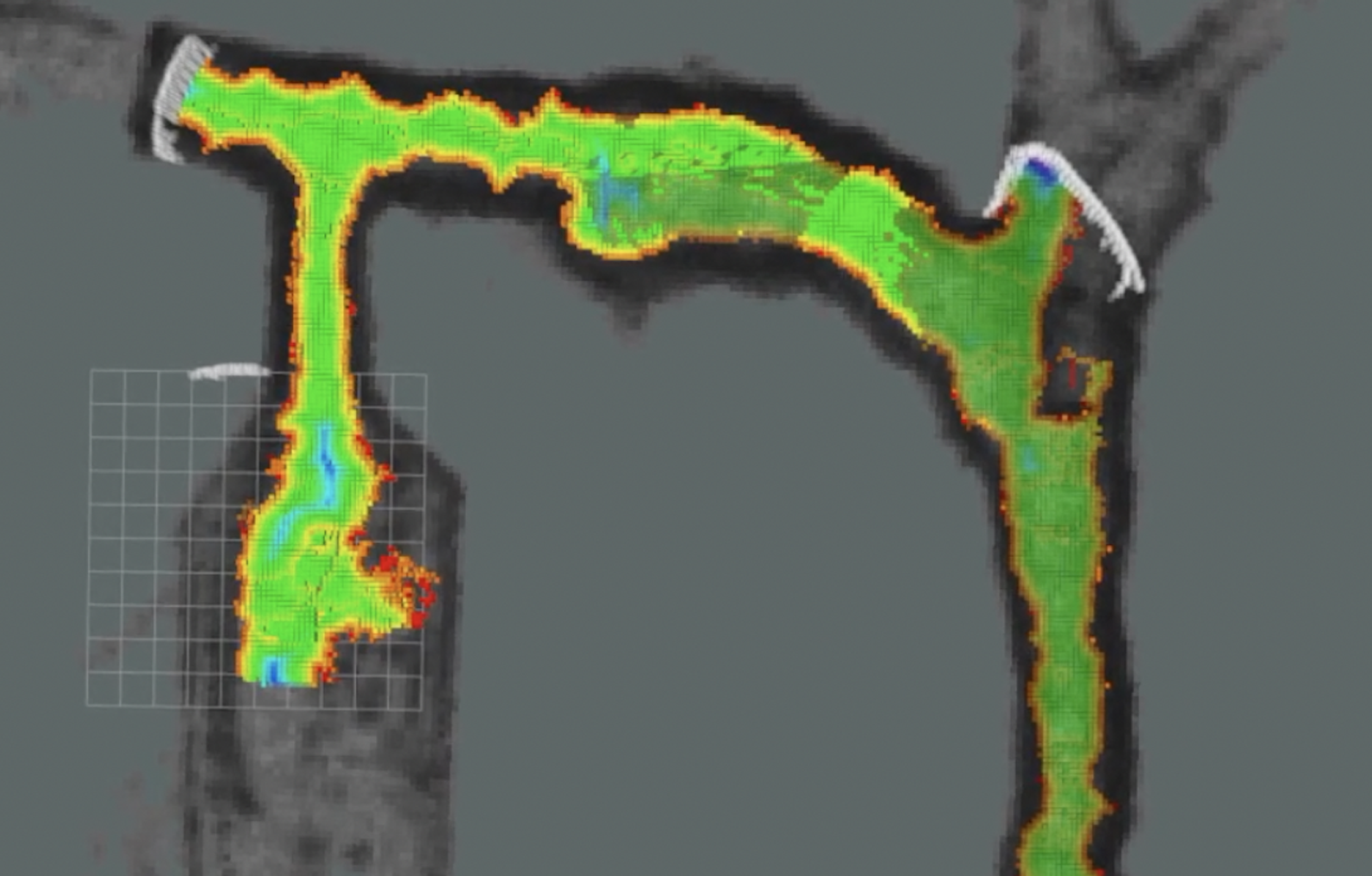}\label{fig:edgaredt} }}
		\hspace{5pt}
		\subfloat[]{{\includegraphics[width=.48\textwidth]{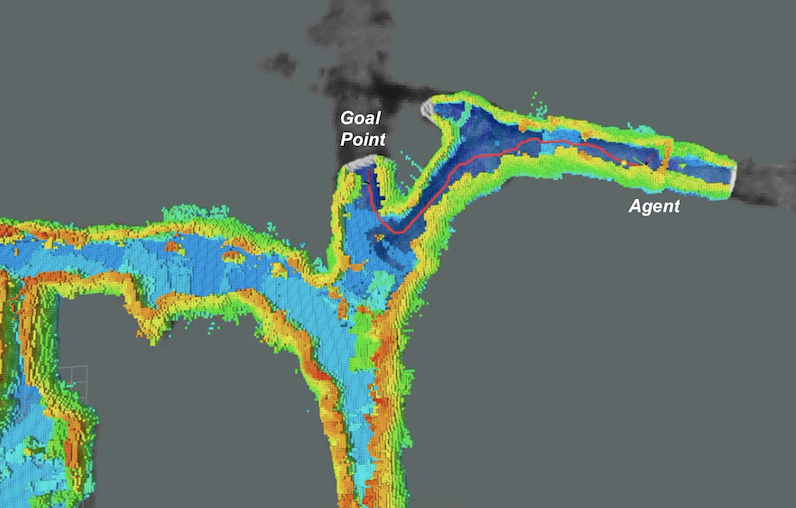}\label{fig:edgarplan} }}
		\caption{RViz views of a single agent exploring the Edgar Experimental Mine in Idaho Springs, CO. a) Example of the Euclidean distance transform ($D(\bm{c})$) map generated during the planning phase. The color of the voxel encodes distance to the nearest occupied cell with red being next to an occupied cell and green / blue voxels as far away from occupied cells. b) Example of a path planned over the speed map to a cluster of frontier. Colored voxels represent occupied cells in the occupancy grid, opaque white voxels represent frontier, and the red line the is planned trajectory to the next goal point. }
		\label{fig:edgarexplore}
\end{figure}

In order to evaluate the utility function at each candidate goal pose, we first need the cost to get there.  Line 6 in Algorithm \ref{algorithm:FrontierPlanning}, ComputeCostToGo$(\mathcal{M}, c(\bm{x})$ evaluates the cost for each candidate goal pose according to the equation below.
\begin{equation}
    \text{Cost}(\bm{p}, \bm{x}) = T(c(\bm{p})) + \frac{|\psi_{\bm{x}} - \psi_{\text{path},0} | + |\psi_{\bm{p}} - \psi_{\text{path},f} |}{\dot{\psi}_{\text{max}}}
\end{equation}
The first term, $T(c(\bm{p}))$, is an approximation of the travel time from the robot's current position to the voxel position of the candidate goal pose, $c(\bm{p})$, using a multi-stencil fast marching method.  The second term is the time required for the robot to turn in place from its current heading, $\psi_{\bm{x}}$, to align with the initial path heading $\psi_{\text{path},0}$, added to the time required to turn from the final path heading, $\psi_{\text{path},f}$, to the final goal pose heading.  Fast marching is a level set method that calculates the arrival time, $T$, of a wave front starting at the robot's current voxel location $c(\bm{x}$), to each voxel in the set of traversable voxels, $\bm{c} \in \mathcal{T}$.  This is calculated using a fast approximation of the solution to the Eikonal Equation 
\begin{equation}
    |\nabla T(\bm{c})|S(\bm{c}) = 1
\end{equation}
where $S(\bm{c})$ is the wave propagation speed at voxel $\bm{c}$ \cite{sethian1999level} \cite{kroon2020MSFM}.  A constant speed map yields a fast approximation of the cost-to-go to each voxel in the map equivalent to Dijkstra's Algorithm.  The safety of the resultant path produced by the planner can be augmented by computing a speed map at each voxel, $\bm{c}$, according to the following equation 
\begin{equation}
    S(\bm{c}) = \frac{1}{2}(\text{tanh}(D(\bm{c}) - e) + 1)
\end{equation}
where $D(\bm{c})$ is the distance to the closest occupied cell. This function maps the distance to the closest occupied voxel to a speed in the range $[0,1]$, resulting in paths that are obstacle-aware.  This means that the planned paths will tend towards the medial axis of the environment, which is analogous to the middle of a hallway, whenever possible \cite{lee1982medial}.  Algorithm \ref{algorithm:FrontierPlanning} finishes by computing paths to the best candidate goal pose(s) by following the gradient of the wave front arrival time, $T(\bm{c})$, from each pose back to the robot's current voxel.  This is quickly done by evaluating the 3D Sobel operator on the 26-voxel neighborhood of each voxel from the candidate goal pose back to the robot.

\begin{figure}[b!]
		\centering
		\subfloat[]{{\includegraphics[width=.48\textwidth]{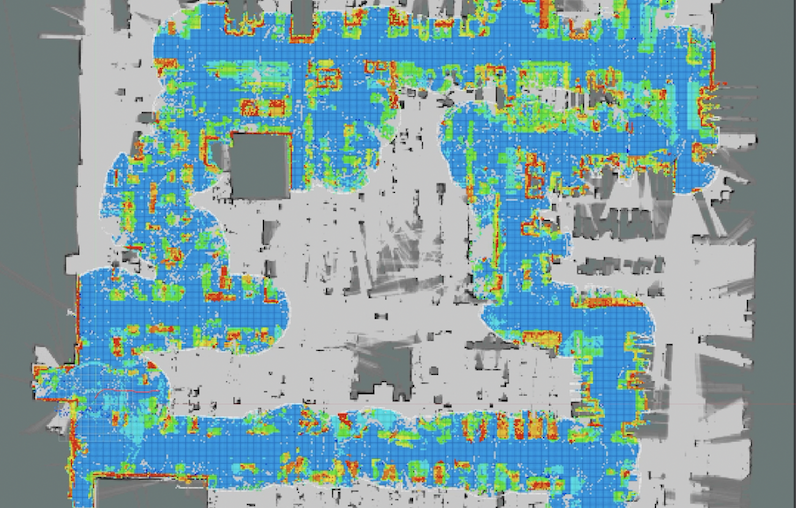}\label{fig:geotechtopdown} }}
		\hspace{5pt}
		\subfloat[]{{\includegraphics[width=.48\textwidth]{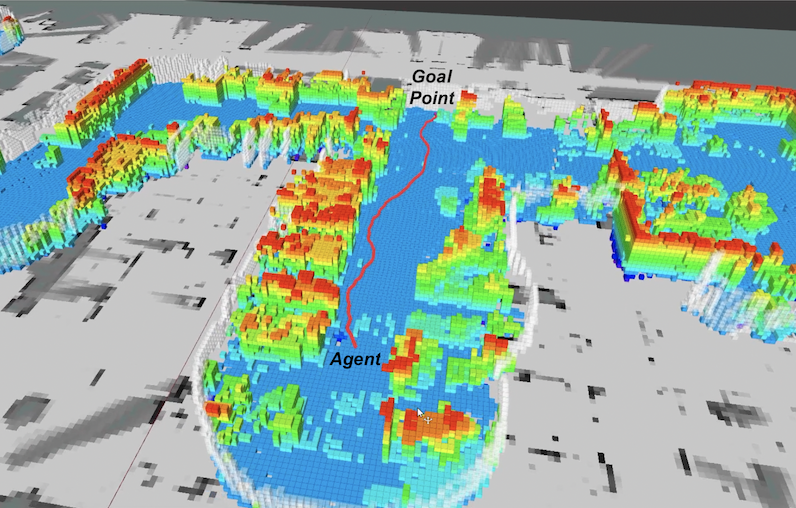}\label{fig:geotechexplore} }}
		\caption{RViz views of a single agent exploring the $100\text{m} \times 80\text{m} \times 6\text{m}$ GeoTech warehouse in Denver, CO., where (a) is a top-down view of the final map after a 25-minute deployment, and (b) show the agent actively planning a path. Colored voxels represent occupied cells in the occupancy grid, opaque white voxels represent frontier, and the red line the is planned trajectory to the next goal point. Planned paths tend towards the medial axis of the obstacle environment, but here we see the path hug the left side of the corridor due to a false obstacle in the underlying distance transform.}
\end{figure}

Although this guarantees that the planner will find the candidate goal pose that minimizes the proposed utility function, calculating the fast marching cost-to-go to every goal pose can become prohibitively expensive if some of the candidates are far away from the robot's current position.  For this reason, there is a stopping criterion based upon the best possible utility of any candidate goal pose that has not been evaluated, $\max_{\bm{p} \in \mathcal{P}_{ne}}{U(\bm{p})}$, is bounded according to the equation below.

\begin{equation}
    \max_{\bm{p} \in \mathcal{P}_{ne}}{U(\bm{p})} \leq \frac{\max_{\bm{q} \in \mathcal{P}_{ne}} \text{FrontierInFoV}(\bm{q})}{T(\bm{c}_{\text{FMM}}) + 2\pi/\dot{\psi}_\text{max}}
\end{equation}

This bound exists because the current fast marching voxel wavefront arrival time, $T(\bm{c}_\text{FMM})$, is an increasing function with algorithm run time.  If we consider the worst possible heading penalty term, $2\pi\/\psi_\text{max}$, the best utility of any remaining candidate goal pose must be below this bound.  The fast marching algorithm stops if the current best goal pose utility among those evaluated is better than any remaining goal pose utility.  This extends naturally to the case of multiple robots. Fast marching evaluates until the $M$ best candidate goal poses that are some minimum distance apart are evaluated, where $M$ is the number of neighboring robots. Then the global planner returns the $M$ best poses, utilities, and paths for the coordination scheme. Paths are followed using a simple steering controller driven by the vehicle's heading error with respect to a look-ahead point a fixed distance down the path.

\begin{figure}[b!]
		\centering
		\subfloat[]{{\includegraphics[width=.48\textwidth]{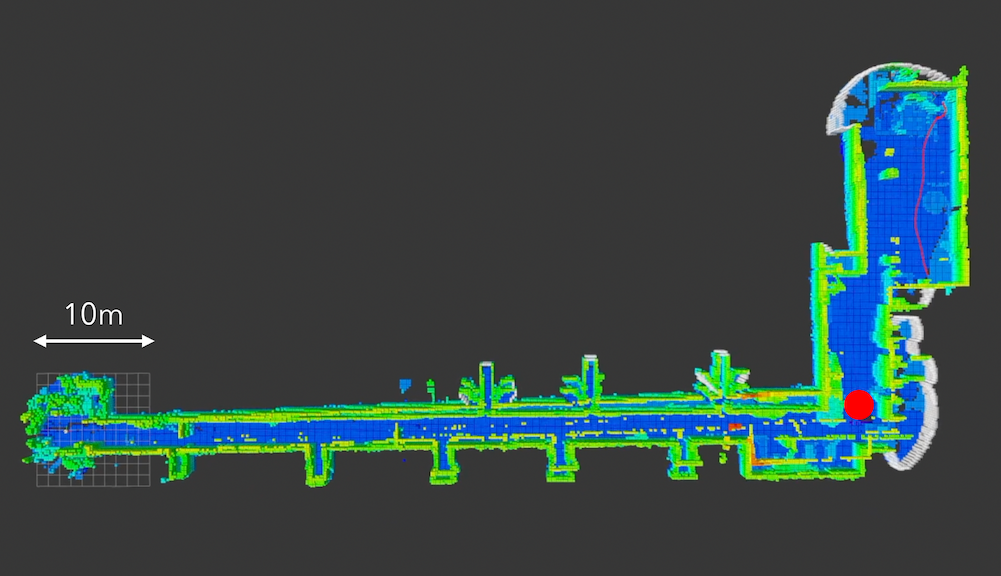}\label{fig:satsopexploretopdown} }}
		\hspace{5pt}
		\subfloat[]{{\includegraphics[width=.48\textwidth]{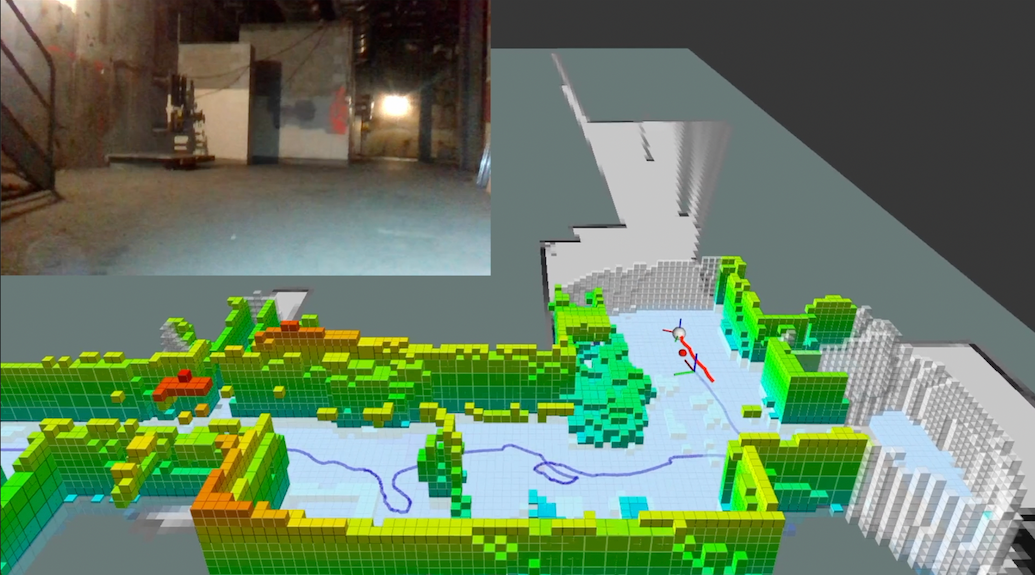}\label{fig:satsopthirdp} }}
		\caption{RViz views of a single agent exploring the Beta Course at SATSOP nuclear facility in Elma, WA, during the Urban circuit event where (a) is a top down view of the map with the red dot indicating the location of (b) a third person perspective.}
		\label{fig:satsopexplore}
\end{figure}

\subsection{Experimental Performance}

\begin{figure}[t!]
		\subfloat[]{\includegraphics[height=.4\textwidth]{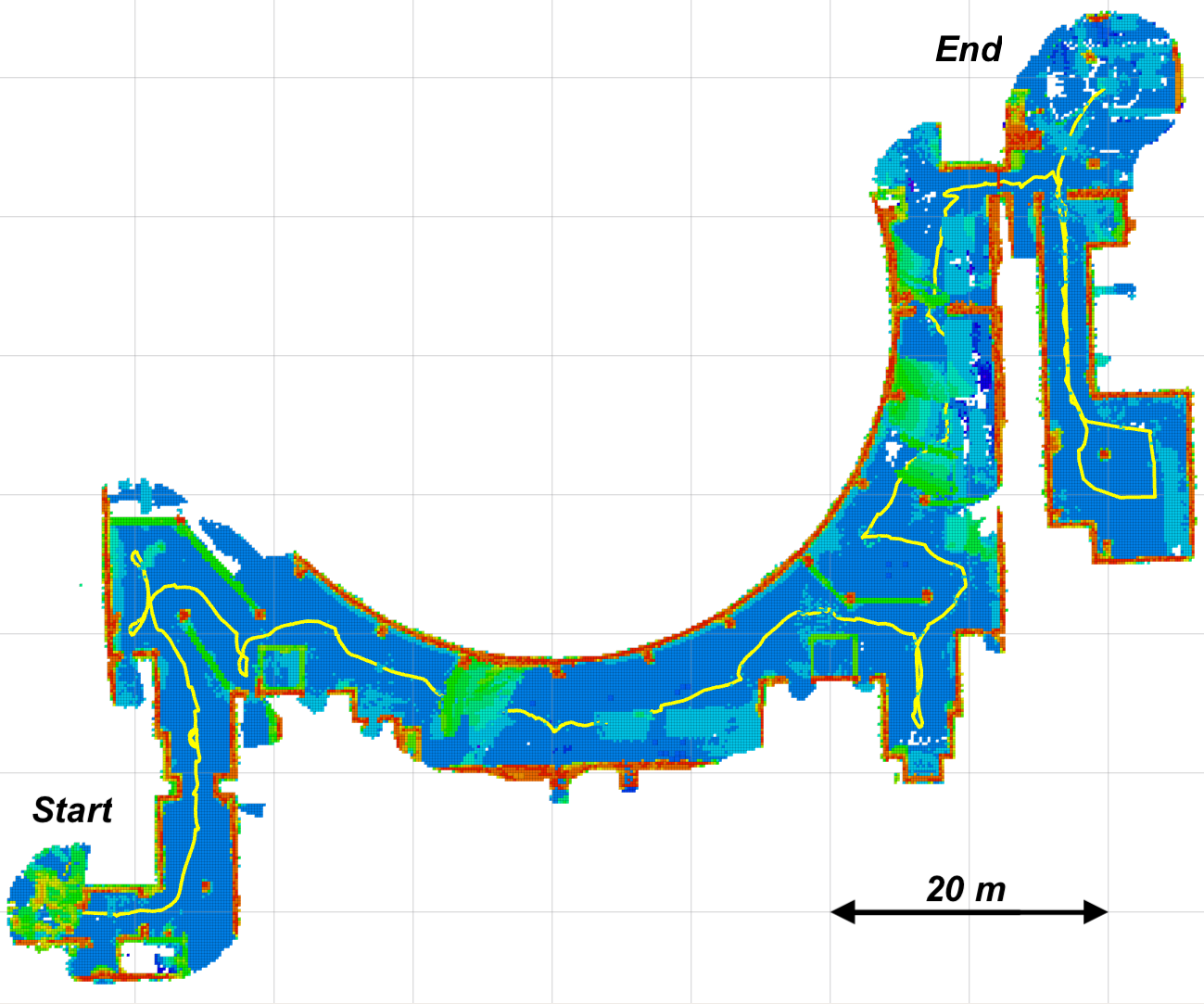}}
		\centering
		\hspace{10pt}
		\subfloat[]{\includegraphics[height=.4\textwidth]{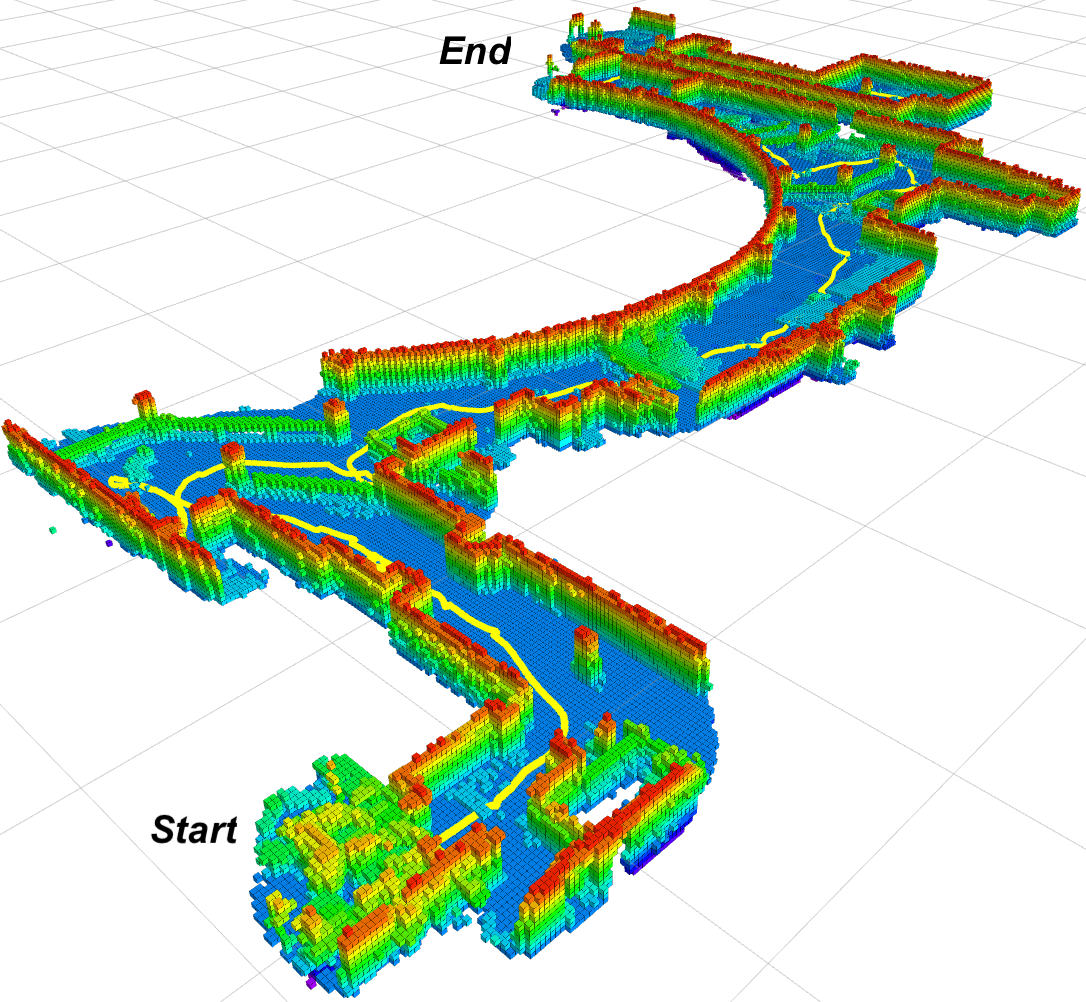}}
		\caption{UGV trajectory (yellow) and OctoMap (rainbow) during autonomous exploration mission, shown both at an (a) top-down view, and (b) angled view. This field deployment took place in Alpha Course at SATSOP nuclear facility facility in Elma, WA, during the Urban circuit event.}
		\label{fig:SATSOP_map_traj}
\end{figure}

Examples of planned paths on the UGV in the Edgar Experimental mine and GeoTech warehouse environments are shown in Figures \ref{fig:edgarplan} and \ref{fig:geotechexplore}.  As shown in these two map images, the robot successfully plans over the mapped obstacle environment to a candidate goal point $10$s of meters away. Figure \ref{fig:geotechtopdown} shows the full route taken by an agent when exploring the $100\text{m} \times 80\text{m} \times 6\text{m}$ GeoTech warehouse, which is an extremely cluttered environment. Examples of deployments at SATSOP nuclear facility for the Urban circuit are shown in Figures \ref{fig:satsopexplore} and \ref{fig:SATSOP_map_traj}. Figure \ref{fig:multiagent_merged_map} shows a map of sections of the Edgar Experimental Mine that was generated by two separate robots who were planning over a shared, merged map using this exploration method. This global planning method was also tested on our aerial vehicle in the GeoTech warehouse. The results of this test are shown in Figure \ref{fig:uav_geotech_warehouse} where the aerial vehicle was able to explore about a fifth of the entire warehouse during a single 10 minute flight. As can be seen in these figures, both air and ground vehicles successfully map a reasonably large portion of these cluttered environments with a minimal amount of overlap of previously explored locations. 

A similar approach to exploration is used in \cite{charrow2015information}, where 2D exploration is demonstrated in an office like environment with a ground vehicle and 3D exploration is demonstrated in an indoor stairwell with an aerial vehicle. In \cite{schmid2020efficient}, a comparison of the performance of different exploration algorithms in a $40\text{m} \times 40\text{m} \times 3\text{m}$ maze environment is presented. Team CoSTAR presented their results from Tunnel circuit in \cite{Otsu2020} where they use a frontier-based exploration algorithm when navigating outside of communication range and cover an impressive amount of ground during their runs. One area of opportunity for our frontier-based planner lies in determining what terrain is actually traversable for each of our platforms and incorporating this information into path generation. Due to the size of our map voxels ($.15\text{m}$) and the relatively low ground clearance of our UGVs, it can be difficult to discern precisely which sections of terrain will cause the vehicle to get physically stuck. Team CoSTAR has presented their results from Urban circuit in \cite{bouman2020autonomous} where they use a quadrupedal robot and a traversability focused local planner to avoid hazardous terrain and navigate stairs.

\section{Aerial Vehicle Vision-Based Local Control}
\label{sec:AVnav}
Typical motion planners in the literature rely on an environment map to plan a guidance path followed by trajectory optimization to obtain a feasible smooth motion plan for the robot \cite{gao2018online,debord2018trajectory,liu2018search}. In an extreme environment where reliability is of major concern, this approach alone poses several safety challenges. During field testing, we found that uncertainty around the locations of obstacles can lead to collisions, especially in confined and extremely cluttered spaces. Sources of uncertainty include localization and mapping errors during agile maneuvers, as well as small and thin obstacles that may go undetected by the map. Moreover, if the environment becomes too complex, the continuous optimization runs into a risk of not being able to converge to a feasible solution within the limits of imposed constraints. Tuning constraints on such a problem is often non-trivial and highly dependent on the environment.  The concept of planning, based on direct high resolution depth information, has been explored in past in the context of bio-inspired obstacle avoidance \cite{Ohradzansky2018,Escobar2019,OhradzanskyMills} and lookahead planning within a depth image \cite{matthies2014stereo,dubey2017droan,ahmad2019real}. These methods, however, are either only suitable for 2D navigation or are based on lookahead planning. On the contrary, our vision-based obstacle avoidance method is well-suited for 3D and senses and avoids obstacles actively, making it more resilient to a changing environment, localization uncertainty and imperfect path following due to model uncertainty. \par

\begin{figure}[b!]
    \centering
    \includegraphics[width=0.6\linewidth]{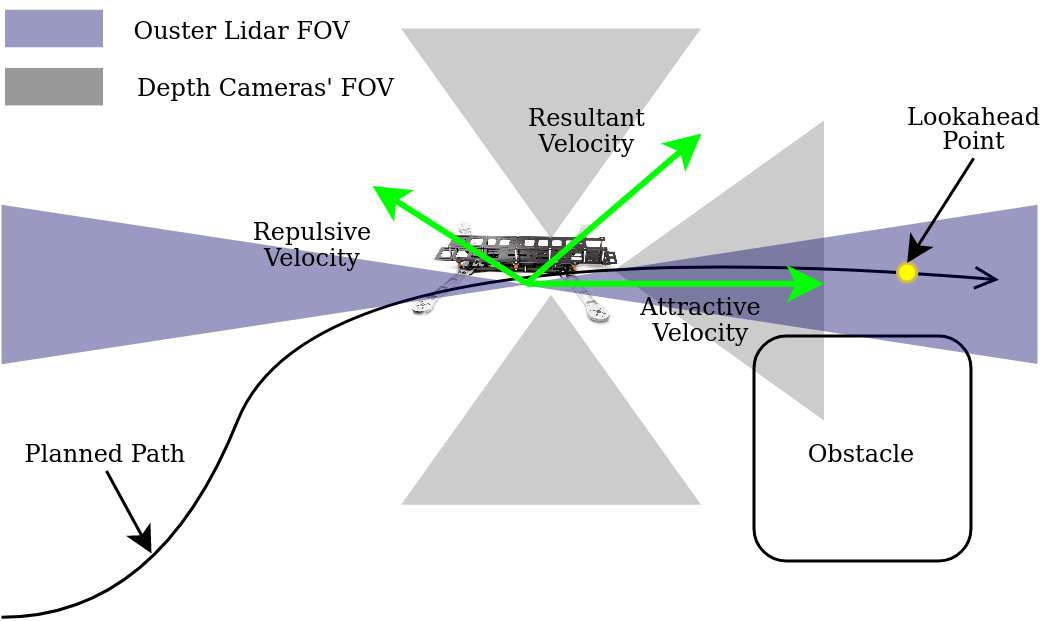}
    \caption{A depiction of the UAV motion planning strategy. The map-based frontier view planner generates a global path at each replan time instant. A lookahead point is chosen along the path that is a constant distance away from the robot at all times. The reactive controller relies on direct depth sensor information to avoid obstacles that are not captured by the map due to localization and mapping uncertainties. The lookahead point moves along the path as the UAV follows and serves as the goal for the reactive controller.}
    \label{fig:uav_planner}
\end{figure}

We proposed and implemented a potential-based solution that directly utilizes high resolution depth information for actively sensing and avoiding local obstacles while following a global path. The controller ensures that the vehicle kinematics are respected during path-following and the heterogeneous perception scheme (map-based and direct vision-based) improves the perception reliability. Each MARBLE UAV utilizes a two-tier planning strategy: direct vision-based local control and map-based global path planning. The frontier view planner from Section \ref{sec:contfront} is leveraged to generate a global path for a UAV. A lookahead point is calculated along the path such that the point keeps a fixed distance from the robot at all times until the end of path. The lookahead point serves as the goal point for the local controller. The two-tier planning strategy is depicted in Figure \ref{fig:uav_planner}. \par

We assume that the UAV follows the unicycle nonholonomic kinematics where the system inputs are the forward, vertical and steering velocity commands in the robot's body frame. The goal of the MARBLE UAV local planner is to complement the global planner by avoiding small obstacles that do not appear in the map, and obstacles that become close in proximity due to mapping and localization uncertainty. It is also important to note that the path generated by the frontier view planner is geometric. The local controller is responsible to generate the velocity control commands to respect the vehicle kinematics during path following. 

The depth images come from all of the onboard depth cameras for the vision-based reactive control. The depth cameras are assumed to follow the pinhole model. For a small 3D object projecting to $N$ depth image pixels, the repulsive vector is calculated as

\begin{align}
    \label{eq:uav_repulsive_vec}
    u_{\text{rep}} = \sum \limits_{i=0}^N  k_{\text{rep}} \bigg(\frac{1}{p_z^{\max}} - \frac{1}{p_z^i}\bigg) \hat{p}_i,
\end{align}

where $k_{\text{rep}}$ is the user-defined gain to tune the aggressiveness of the repulsive potential, $p_z^{\max}$ is the maximum range of the depth camera and $p_z^i$ is the depth of the $i$th pixel. In this two-tier planning setup, the goal point, or lookahead point, for the reactive controller is continually updated to be a fixed distance away from the robot, and resides along the global path. The resultant vector is calculated as

\begin{align}
    \label{eq:uav_resultant_vec}
    u_{\text{res}} = (1-|u_{\text{rep}}|/|u_{\text{rep}}^{\max}|) \hat{x}_{\text{lookahead}} + u_{\text{\text{rep}}}/|u_{\text{\text{rep}}}^{\max}|,
\end{align}

where $\hat{x}_{\text{lookahead}}$ is the direction of the lookahead point in the vehicle's body frame and $|u_{\text{rep}}^{\max}|$ is the maximum magnitude of the repulsive vector. The term $1-|u_{rep}|/|u_{rep}^{\max}|$ regulates the speed of the robot depending upon the proximity of the obstacles. As the vehicle enters an increasing level of clutter, the effect of the repulsive vector becomes more dominant. The resulting velocity commands for the UAV are calculated as

\begin{align}
    \label{eq:uav_vel_cmds}
    v_x &= v_x^{\max} u_{\text{res}}^x, \nonumber \\
    v_z &= v_z^{\max} u_{\text{res}}^z, \nonumber \\
    v_{\theta} &= v_{\theta}^{\max} \frac{1}{\pi} \text{atan2} (u_{\text{res}}^y,u_{\text{res}}^x),
\end{align}

where $v_x$, $v_z$ and $v_{\theta}$ are the forward, vertical and steering velocity commands respectively. The maximum velocities for each axis are defined by $v_x^{\max}$, $v_z^{\max}$ and $v_{\theta}^{\max}$. \par

\begin{figure}[b!]
		\centering
		\subfloat[]{{\includegraphics[height=0.32\textwidth]{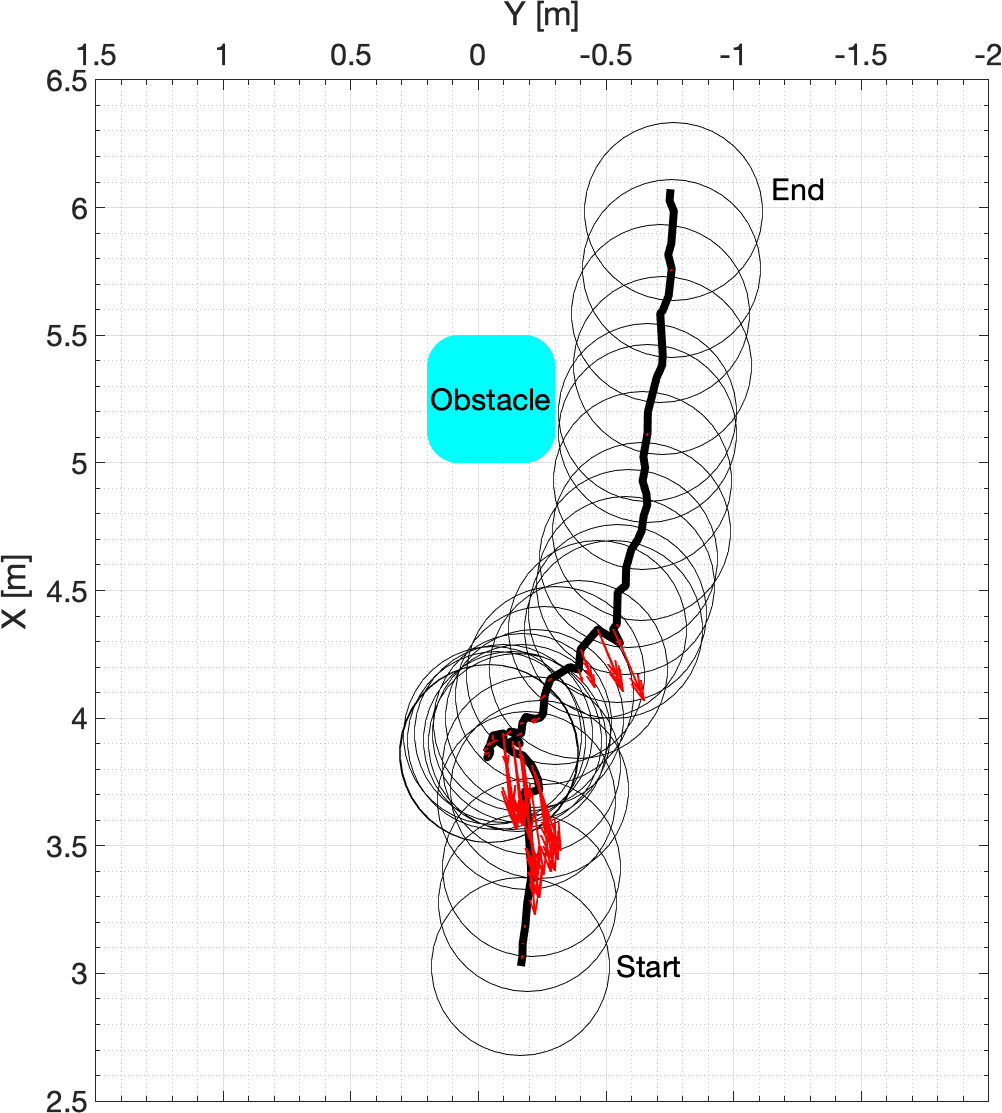}\label{fig:aspen_traj} }}
		\hspace{3pt}
		\subfloat[]{{\includegraphics[height=0.32\textwidth]{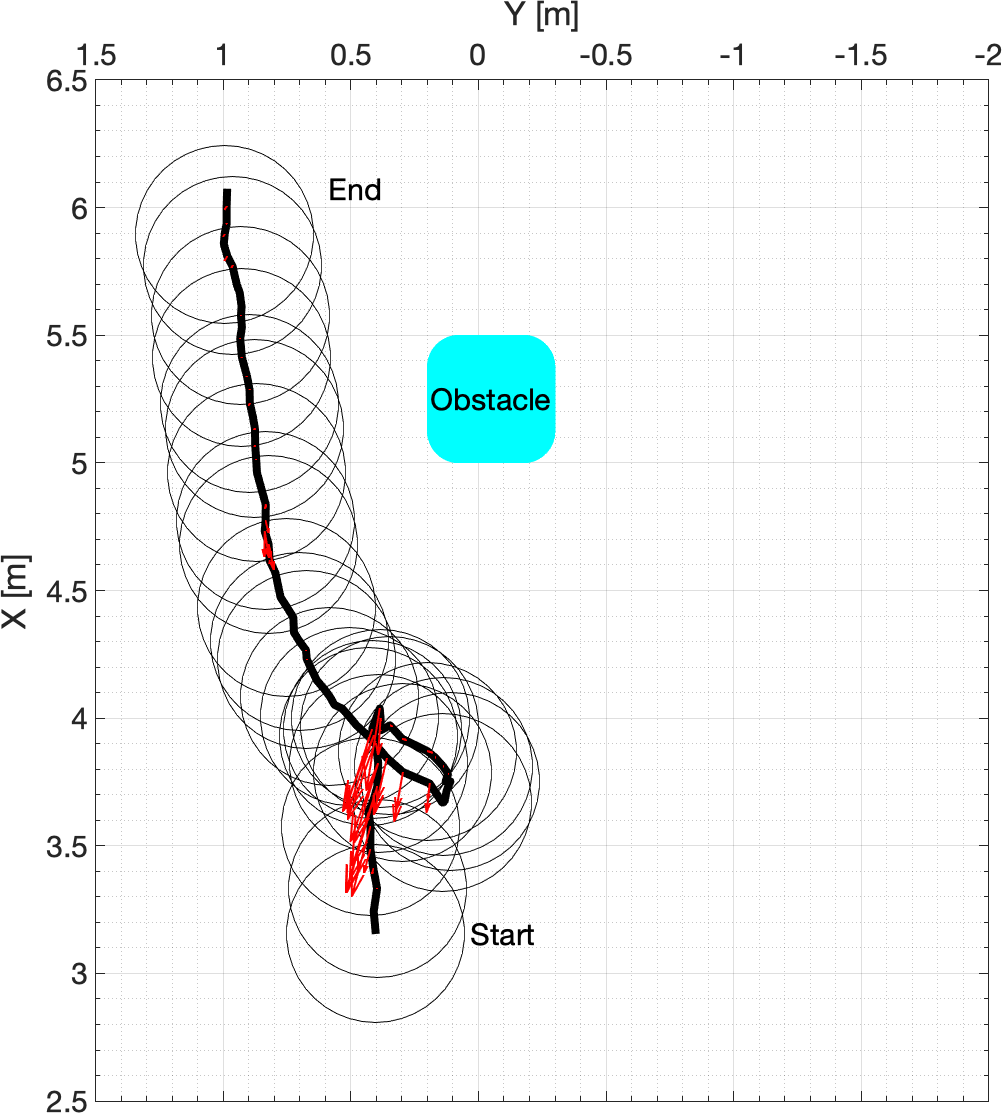}\label{fig:aspen_traj1} }}
		\hspace{3pt}
		\subfloat[]{{\includegraphics[height=0.32\textwidth]{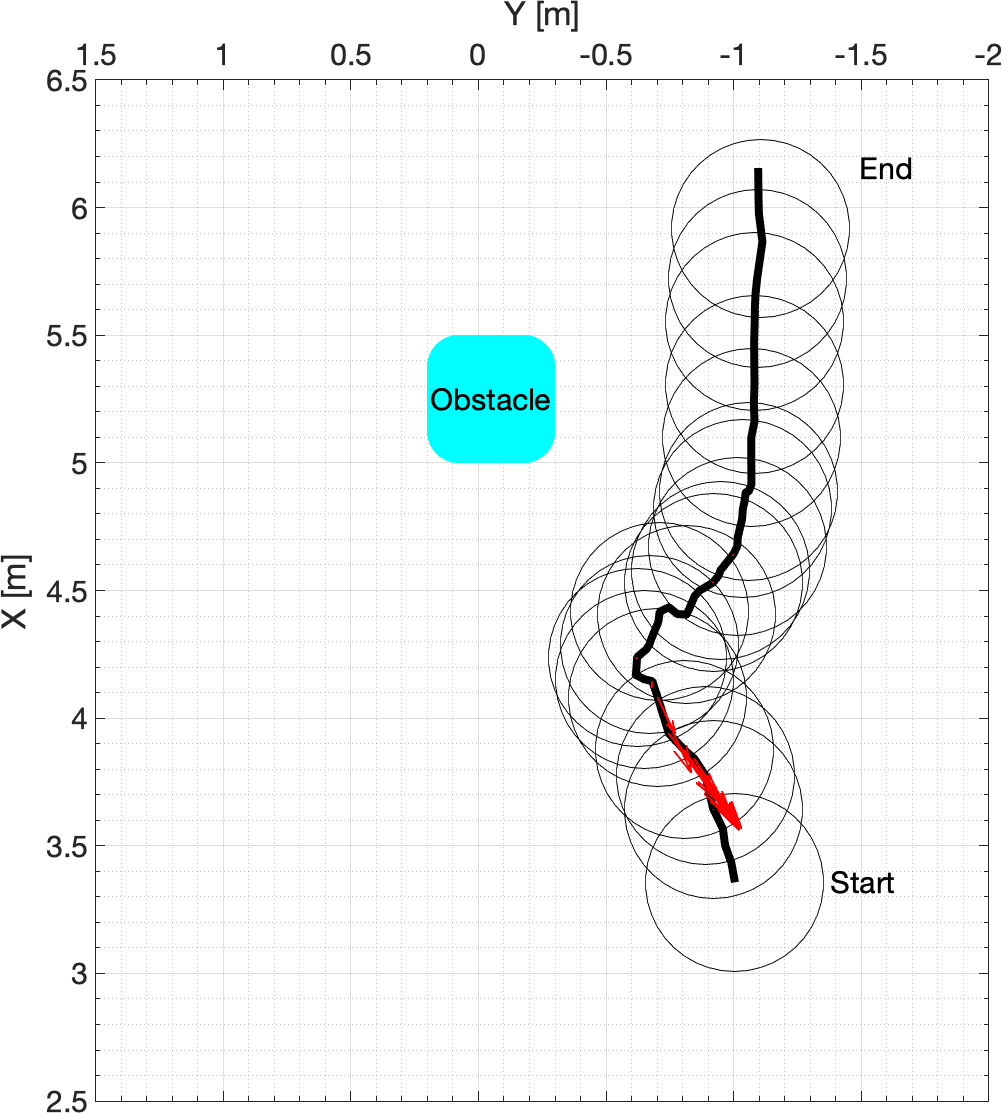}\label{fig:aspen_traj2} }}
		\caption{Results from a single obstacle local planner flight tests. The black circles represent the UAV physical bounds. The black lines show the path followed by the UAV during each of the maneuvers with red arrows representing the repulsive vectors. The UAV starts from the respective start locations for each of the runs and maneuvers towards the goal points that are (a) 2 m behind the obstacle, (b) 5 m behind the obstacle, and (c) 4 m behind the obstacle. The approximate obstacle location is overlaid on the plots.}
		\label{fig:uav_local_control}
\end{figure}

In order to evaluate the effectiveness of the reactive controller in an isolated manner, we first carried out several flight tests in the presence of a single obstacle. These tests were carried out in a simpler lab space with MARBLE UAV completely relying on the onboard sensor stack. A small obstacle was placed in the middle of the lab and the start and goal points were set on the opposite sides of the obstacle. Results from three of the tests are shown in Figure \ref{fig:uav_local_control}.

\begin{figure}[t!]
		\centering
		\subfloat[]{{\includegraphics[height=.5\textwidth]{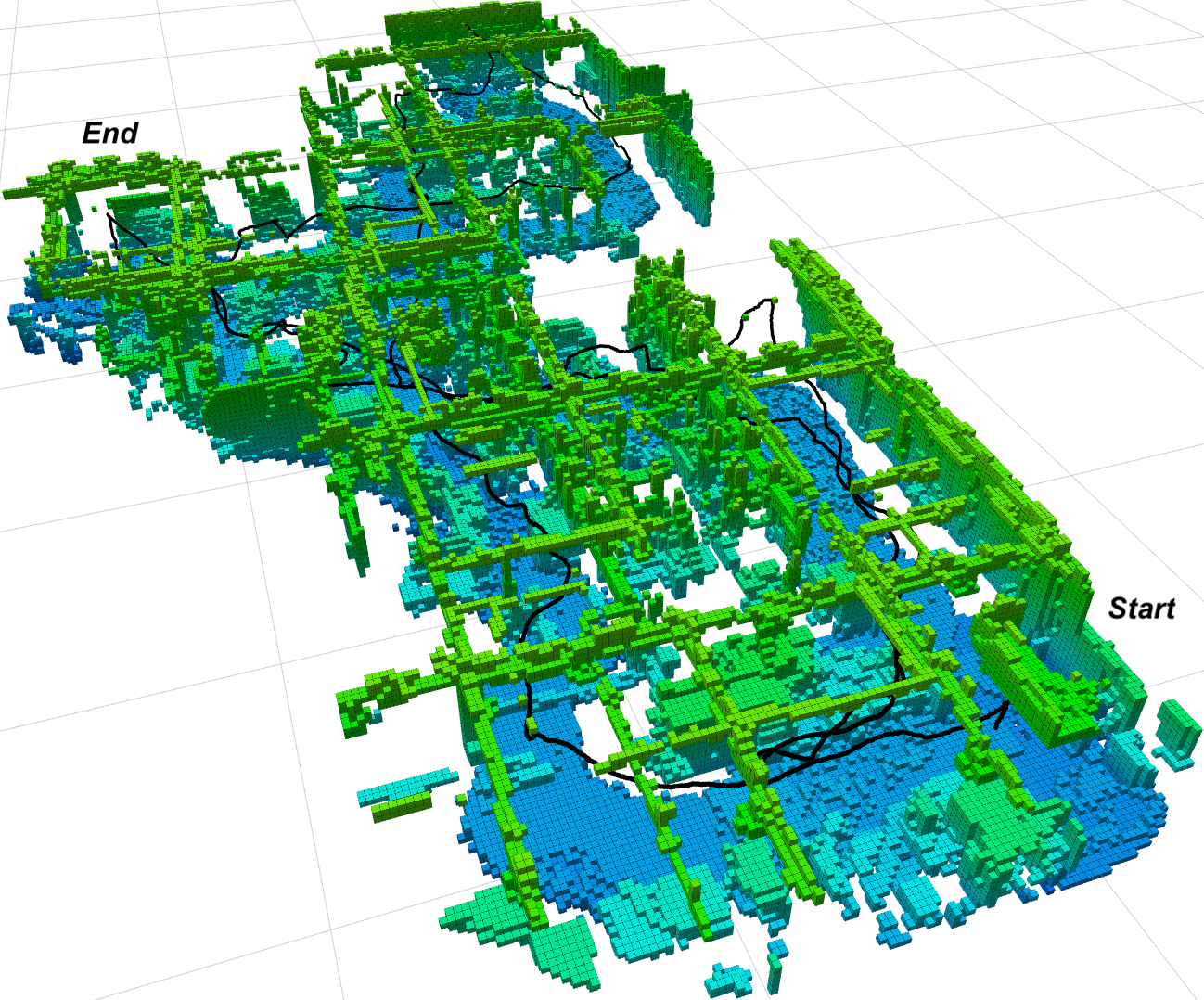}\label{fig:geotechflight} }}
		\hspace{5pt}
		\subfloat[]{{\includegraphics[height=.5\textwidth]{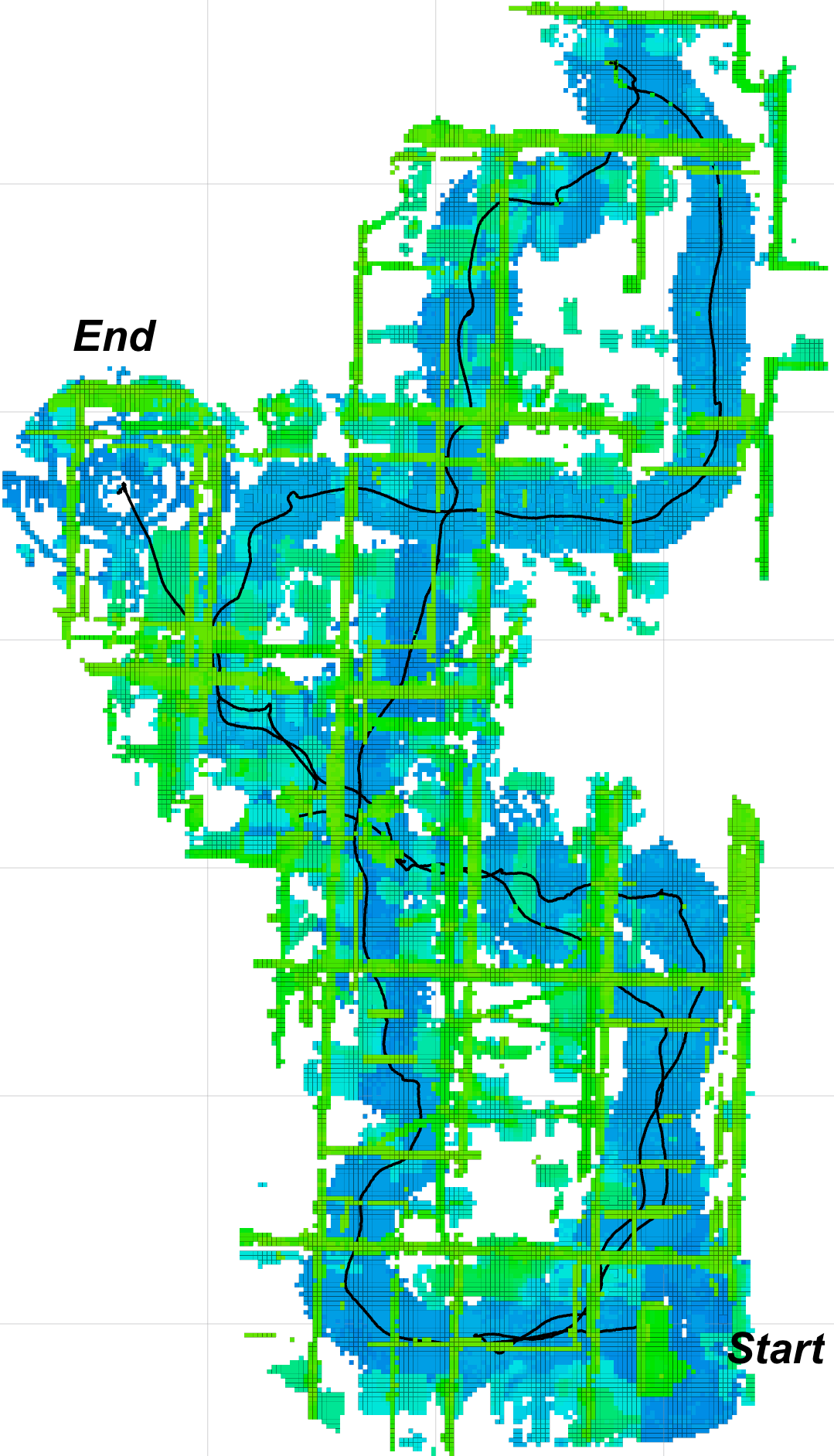}\label{fig:geotechflight2} }}
		\hspace{5pt}
		\caption{The OctoMap and flight path of a ten-minute autonomous exploration mission for the GeoTech warehouse in Denver, CO are presented from both a (a) rotated and (b) top-down perspective. The UAV travelled approximately 300 m within 30 m $\times$ 60 m $\times$ 5 m volume. The ceiling from the original map is removed for better visibility.}
		\label{fig:uav_geotech_warehouse}
\end{figure}

Existing long-range UAV planning methods in the literature such as \cite{bircher2016receding,dang2019explore,Dang2020,dharmadhikari2020motion} have been demonstrated in underground tunnels over distances around 150 m or small confined nearly-empty rooms. These environments exhibit general predictable structures, for instance, underground mines typically consist of long corridors connected through junctions with most of the obstacles appearing close to the walls. We carried out flight tests using both global and local planners integrated to explore the GeoTech warehouse facility in Denver, Colorado. Unlike corridor-like environments which are primarily composed of 2D structures, the complex 3D nature of the warehouse environments makes them challenging for autonomous exploration. This particular environment is highly cluttered and heterogeneous since it houses manufacturing activities. Looking up, steel-truss ceilings support the roof, featuring exposed sprinkler, ductwork, and conduit networks, as well as light fixtures and ceiling-to-floor electrical cabling. In case of a UAV navigating in such an environment, even a single close contact of the robot with an obstacle, for instance an electric cable, can prove detrimental to the exploration mission safety. Our two-tier planning setup demonstrated safe and reliable exploration through the environment \cite{ahmad20213d}. One of the runs is highlighted in Figure \ref{fig:uav_geotech_warehouse} in which the UAV navigated around 300 m with a full battery pack which amounted for 11 minutes of flight time. 

While this solution significantly improved the safety of the UAVs, it had difficulty navigating narrow corridors, discerning thin obstacle from erroneous noisy returns in the depth image, and assessing the quantity and size of local obstacles. In order to overcome these limitations in the Urban circuit event, the UAV local planning method was overhauled with a probabilistic technique that explicitly tracks individual obstacles within the depth image. Each input depth image is first discretized in a 2.5D fashion. This implies pixel discretization along the $x$ and $y$ axes and depth discretization along the $z$ axis of the depth image. Each voxel inside the 2.5D voxel grid is considered a potentially occupied region. We based our probabilistic observation and state transition models on the number of depth image points inside each voxel and the ability of an occupied voxel, hence an obstacle, to move inside the depth image respectively. The latter is modelled as a Gaussian governing the expected speed of an obstacle inside the voxel grid. We then used the Sequential Importance Sampling (SIR) technique \cite{smith2013sequential,doucet2000sequential} which follows a two-step process, prediction and update, to find the posterior of the occupied voxels given a history of depth images. By formulating the problem as a Partially Observable Markov Decision Process (POMDP), we utilized the occupancy probability distribution (SIR posterior) over the voxel grid to generate robot velocity actions. In order to achieve real-time execution of the method onboard the MAV, we used the QMDP approximation to solve the POMDP problem \cite{littman1995learning,smallwood1973optimal,littman1994witness}. The reward function is chosen to be the sum of potentials along the path if a given control input is followed for a fixed horizon. The action space is discretized and the action that best reduces the potential over the fixed horizon is the preferred solution to the POMDP. This technique provides us a robust and more flexible framework to generate robot actions directly from a depth image stream, rectifying major issues with the local planner faced at the Urban circuit. Deeper insights into this motion planner are provided in  \cite{ahmad2021apf,ahmad2021depth}. Using this approach, we could actively localize obstacles as thin as 8 mm in diameter relative to the robot.

\section{UDP-Mesh Based Communications}\label{sec:comms}
Implementing effective communication systems in subterranean environments is challenging for a number of reasons, including sudden lost-comm situations and potentially fleeting windows of connectivity. Within the structure of the competition, artifact reports are high-value data, whereas mapping and coordination data are relatively low-valued. This disparity in data importance suggests an operational requirement for quality-of-service prioritization and message delivery guarantees. Meanwhile, using ROS (version 1 unless otherwise specified) for system message passing presents practical challenges to implementing these operational requirements due to inherent architectural limitations. Alternative message transport mechanisms exist; in particular, the transport-agnostic design of ROS2\footnote{\url{https://design.ros2.org/}} has the potential to provide a quality-of-service guarantee at the transport layer in addition to other desirable services such as discovery and name resolution. However, ROS2 was insufficiently developed to meet our needs without major porting efforts; instead, a replacement system was designed to leverage the advantages of ROS2 in a ROS1 environment.

Per design, ROS connections between nodes are not aware of one another. Initial system development utilized the \textit{multimaster\_fkie} package\footnote{\url{http://wiki.ros.org/multimaster_fkie}} to bridge topics between ROS masters in a master-per-robot configuration.  Under the \textit{multimaster\_fkie} architecture, standard TCPROS transport is still used for inter-node communication; nodes subscribing across robot boundaries are not aware of the potential for network saturation. For low-bandwidth topics such as telemetry, where message size and update rate (1Hz) require less than 1kB/s, this behavior is acceptable. However, as the total bandwidth required increases, there is no method for prioritizing topics among one another. In practice, this lack of prioritization manifested as message latency spikes, in which large map data transmissions (several megabytes in size) prevented reception of artifact reports and telemetry (under 1 kB each, exclusive of artifact images) due to channel saturation. While the map data was in transit, other higher-priority data was delayed. Traditional methods of implementing quality-of-service are not possible in the ROS environment; for example, methods utilizing specific ports are not implementable due to ROS' usage of independent links between every publisher and subscriber. Limitations of existing ROS infrastructure necessitate the use of some other transport to provide prioritization services.

The core innovation in our networking stack is the implementation of a custom ROS transport solution based on UDP, called \textit{udp\_mesh}. \textit{Udp\_mesh} provides discovery services and reliable delivery using UDP datagrams compatible with any underlying layer 2 mesh networking solution without requiring any changes to ROS nodes or inspection tools. As an example of a customization for our particular network environment, \textit{udp\_mesh} was implemented without utilizing multicast traffic to recognize the limitations of multicasting over wireless meshes. \textit{Udp\_mesh} implements standard services such as discovery, address resolution, ROS message encapsulation, point-to-point transport, point-to-multipoint transport, and quality-of-service prioritization. Fundamentally, \textit{udp\_mesh} creates a single node on each ROS master through which all nodes attached to that master transmit and receive arbitrary ROS messages through the mesh. \textit{Udp\_mesh} handles fragmentation, transport, and reassembly of ROS messages ranging from single-byte messages up to approximately 2TB per single message. Since all messages are now routed through a single node, traffic shaping can be readily implemented. While independently conceived and implemented, \textit{udp\_mesh} improves upon the \textit{Pound} \cite{harms2017development}     \footnote{\url{https://github.com/dantard/unizar-pound-ros-pkg}} package by adding additional quality-of-life services such as name resolution and removing a requirement that topics and priorities need to be declared at compile time. Other competitors faced similar limitations with bandwidth and the ROS networking stack. In \cite{9172496} both low frequency radios for high priority topics, and high frequency radios for low priority topics were used in conjunction with open source \textit{nimbro\_network} \footnote{\url{https://github.com/AIS-Bonn/nimbro_network}} package to solve the ROS transport and prioritization problems. \textit{Udp\_mesh} allows us to control the prioritization of our data at a software level.

\section{Multi-Agent Coordination}
\label{sec:multiagent}

\begin{figure}[b!]
    \centering
    \includegraphics[scale=0.45]{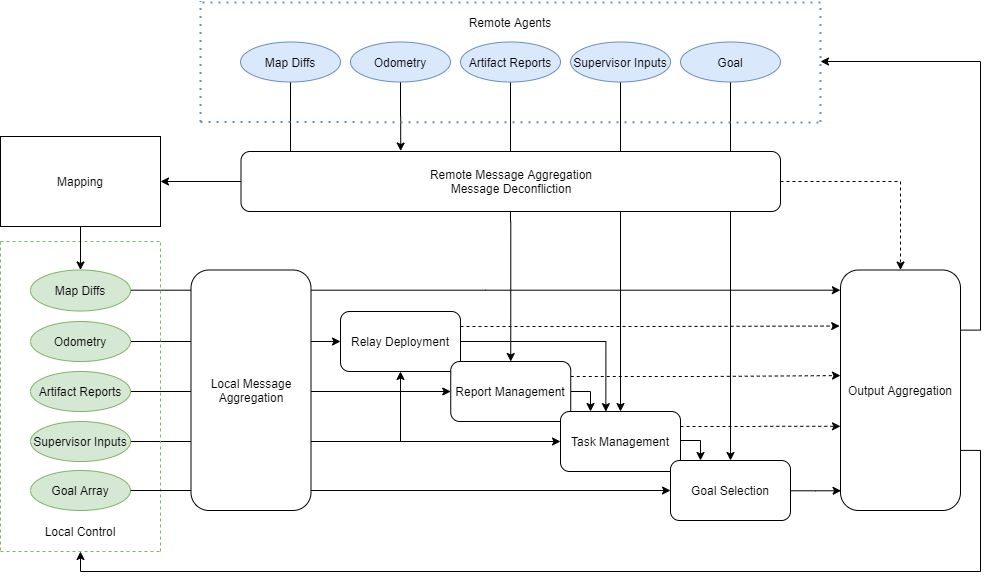}
    \caption{MARBLE Multi-Agent Framework.}
    \label{fig:multi_agent_framework}
\end{figure}

Rapid situational awareness of subterranean environments can be difficult to achieve depending on the size of the environment in question. Having multiple agents can increase the amount of area covered, but they must share information in order to reduce overlap in exploration. Our overall strategy relies on multi-agent coordination through data sharing and goal deconfliction, in order to take advantage of the use of multiple robots with different capabilities and reduce redundant coverage.  The primary challenge was designing a multi-agent system that could handle robots frequently leaving communications, due to the difficult communications coverage in the environments.  Multi-agent coordination can be effectively divided into implicit methods, where information such as map data is shared and reasoned over \cite{Yamauchi1999}, and explicit methods, where shared data is used to direct individual agents \cite{Burgard2002}. Centralized coordination requires a robust communications network, so limited communications environments such as those in the Subterranean Challenge can benefit from a decentralized approach. Although auctions, or market-based coordination \cite{gerkey2002}, are typically thought of using a centralized auctioneer, decentralized systems may also use a market-based approach, such as through an internal single-item auction \cite{Smith2018} or Broadcast of Local Eligibility \cite{Werger2000}.

The multi-agent framework in Figure \ref{fig:multi_agent_framework} evolved throughout the competition from simply handling message transmission between agents to providing the full mission management solution for each agent, including both implicit (through map sharing) and explicit coordination (through goal deconfliction)\cite{RileyFrew}. The framework provides efficient and reliable sharing of information between robots as well as with the Human Supervisor.

Each robot's multi-agent node compiles local information such as odometry, artifacts, current goal, and number of map diffs available. It packages this data, along with the most current data for any neighbor it is aware of, into a small message that is broadcast to every other agent (robots, beacons and base station) every second. Remote agents compare their own stored data for this agent and if necessary request missing map diffs and artifact images. This ensures any robot has the potential to eventually transmit all data for another robot back to the base station and Human Supervisor. 

The multi-agent agent node is responsible for the task management of the robot. Potential tasks include explore, report artifacts, deploy beacon, and process commands from the Human Supervisor. Furthermore, the explore task includes goal selection among several potential goals provided by the global planning node described in Section \ref{sec:contfront}. The selected goal's path is relayed to low-level control, as well as whether to transition to trajectory following mode to return home. The multi-agent node also monitors when the robot is not able to reach a goal, and marks the area to prevent global planning from continuing to attempt to plan there.

\begin{algorithm}[t!]
  \caption{\quad $G \leftarrow$ GoalSelect$(Ga, NeighborGoals)$}
  \begin{algorithmic}[1]
    \IF {$NoNeighbors$ \OR $OneGoal$}
      \STATE $G \leftarrow$ $Ga[0]$
    \ELSE
      \STATE $conflict \leftarrow$ \TRUE
      \STATE $i \leftarrow$ 0
      \WHILE {$conflict$ \AND $i <$ Length$(Ga)$}
        \FOR {$n$ in $NeighborGoals$}
          \IF {GetDistance$(Ga[i](position), n(position)) < DeconflictRadius$ \AND $Ga[i](cost) > n(cost)$}
            \STATE $conflict \leftarrow$ \TRUE
            \STATE \textbf{break}
          \ENDIF
          \STATE $conflict \leftarrow$ \FALSE
        \ENDFOR
        \STATE $i \leftarrow$ $i + 1$
      \ENDWHILE
      \IF {$conflict$}
        \STATE $G \leftarrow$ $Ga[0]$
      \ELSE
        \STATE $G \leftarrow$ $Ga[i-1]$
      \ENDIF
    \ENDIF

  \RETURN $G$
  \end{algorithmic} 
  \label{algorithm:GoalSelection}
\end{algorithm}

\textbf{Explicit Coordination} - The goal selection algorithm also includes a deconfliction process with other robots' goals through an internal market-based auction \cite{RileyFrew}. The frontier-based exploration algorithm described in Section \ref{sec:contfront} produces an array of goal points $Ga$ ordered by $cost$. The number of goal points is based on the total number of robots in the system, and each goal point is separated by a minimum distance. These goals are used in Algorithm \ref{algorithm:GoalSelection} to select the best goal $G$. Initially, each agent chooses its lowest cost goal point, where cost is a function of path distance, and broadcasts it to other agents. Each agent compares any received goals $NeighborGoals$ to each potential goal. If the received goal is within the minimum distance $DeconflictRadius$, and the received cost is lower than its own cost, the agent moves down the list until it finds a goal that does not conflict with another agent. If all goals conflict, it continues to its own lowest cost goal. If a received goal is within the minimum distance, but its own cost is lower, it similarly broadcasts its intention. The other agent who broadcast that conflicting goal either receives that goal and performs an identical calculation, causing it to move down its own list; or if communication is broken, it continues to its best goal as if there were no coordination, and in this case both agents navigate to the same area until new goals are found. This selection method does not require a central coordinator or require the agents to wait for responses from others; however, it does not guarantee an optimal solution. 

\begin{figure}[b!]
		\subfloat[]{\includegraphics[width=.48\textwidth]{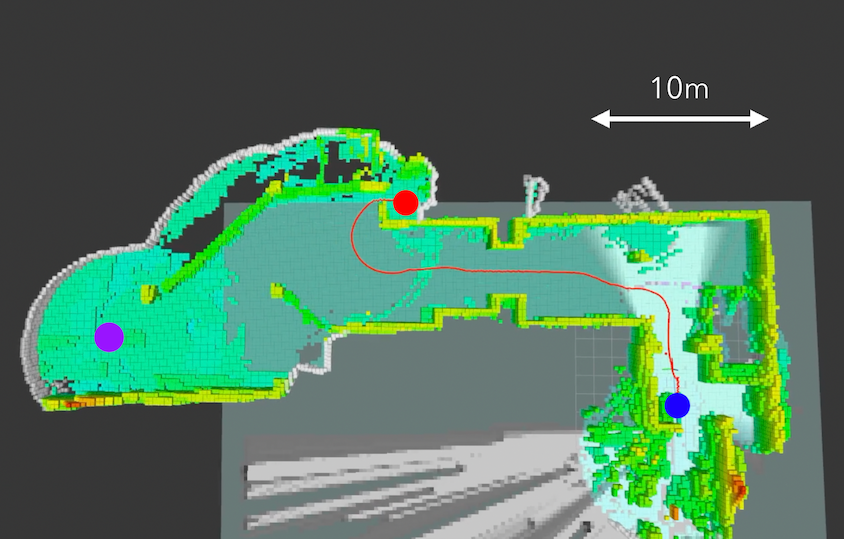}}
		\centering
		\hspace{10pt}
		\subfloat[]{\includegraphics[width=.48\textwidth]{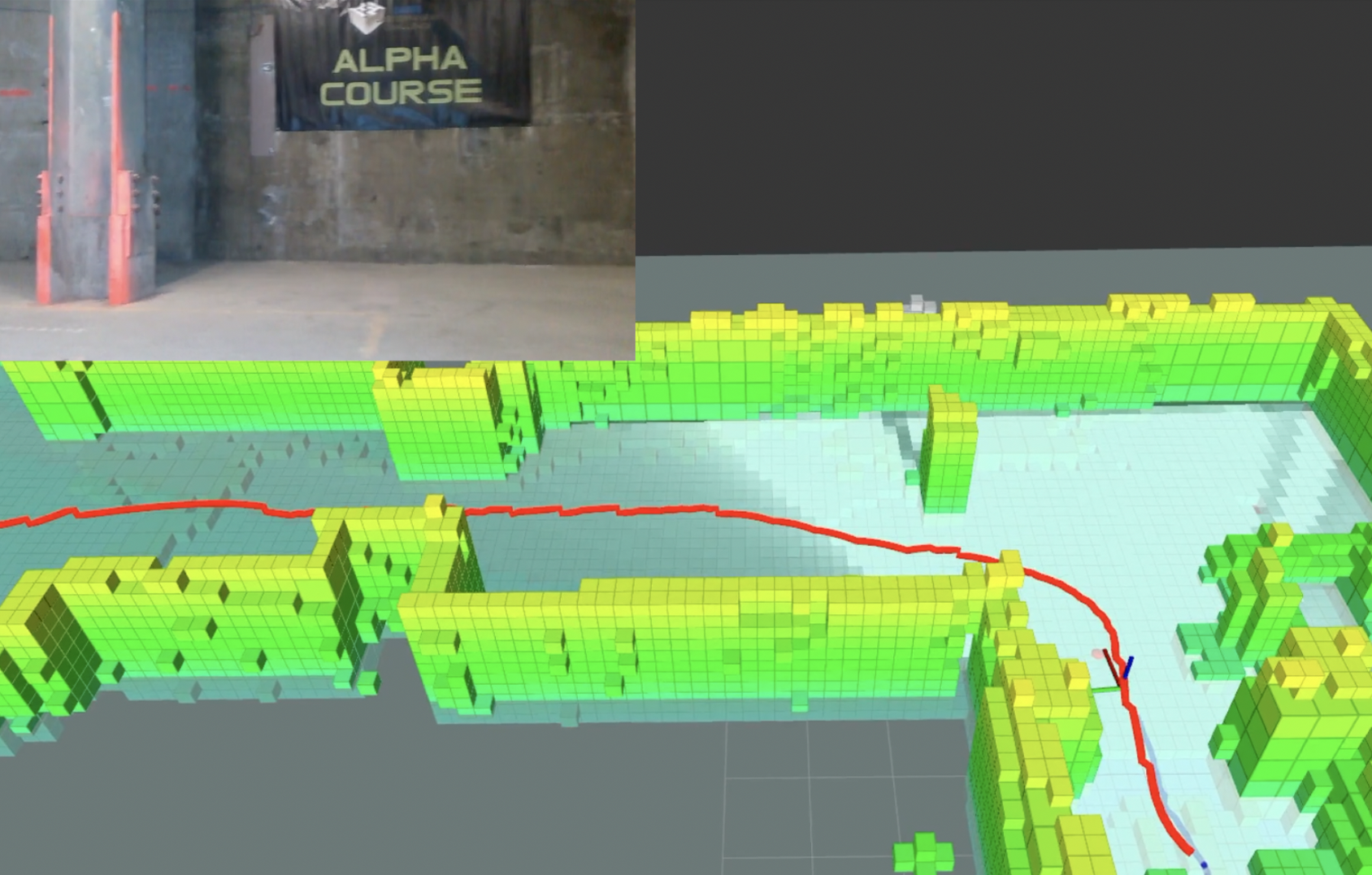}}
		\caption{Example of multi-agent coordination in the SATSOP nuclear facility. a) One agent has already been deployed and has reached the purple dot. It has shared its map, frontier, and candidate goal points to the second agent, who is able to plan a path from the start point (blue) to an unexplored room (red). b) Close up of the generated path and a first person view of the environment in the top-left corner.}
		\label{fig:SATSOP_multi_agent}
\end{figure}

Our algorithm allows coordination without relying on back-and-forth communication. If messages are lost due to communications errors or no communications, then the agents just act as if there were no coordination and proceed to their lowest cost goals. During simulation this method was highly effective in preventing agents from duplicating exploration. However, during competition this was not well-observed due to limited deployment of robots. The largest multi-agent gains were seen due to implicit coordination through map merging, discussed in Section \ref{sec:mapping}, which allowed the robots to plan over maps received from other agents. This was most evident at the starting gate, where the first robot would pass its map back to the next robot in line, and instead of planning a goal point a few meters in front of it, the next robot could create a plan along the entire length of the hallway right from the start. This scenario is shown in Figure \ref{fig:SATSOP_multi_agent}.

\section{Lessons Learned}
\label{sec:LessionsLearned}

\subsection{Artifact Detection}
\label{sec:AD}

\textbf{Initial Evaluation} - Object detection is a well researched problem in computer vision and we evaluated state of the art methods for both 2D and 3D detectors. Common 2D detectors such as region proposal based networks like Fast R-CNN \cite{girshick2015fast} require multiple passes over an image to classify an object and then detect where the object is in the image. In contrast, YOLO \cite{redmon2016you} performs both classification and detection in a single regression making it a significantly faster detection (0.5 FPS for Fast R-CNN and 45 FPS for YOLO). Various methods exist for classifying objects from 3D point clouds \cite{maturana2015voxnet} \cite{qi2017pointnet}. While methods such as Pointnet have been shown to run efficiently on powerful GPUs, they are still limited to classification and not full detection. Extensions such as Voxelnet \cite{zhou2018voxelnet} and PointRCNN \cite{shi2019pointrcnn} have been made to perform object detection however, they require powerful GPUs which are impractical for mobile robots. As a result, we chose to perform 2D detection using the latest YOLO model \cite{farhadi2018yolov3}. Since the challenge requires a singular point estimate within a 5m radius of ground truth, for each artifact we are able to perform depth estimation using external depth data after the initial detection in a 2D image. We shifted our classification hardware from a NVIDIA Jetson TX2 to an NVIDIA GTX 1650 with TensorRT acceleration \cite{vanholder2016efficient}, allowing for classification at 60FPS using the YOLO V3 model \cite{farhadi2018yolov3} at a resolution of $608 \times 608$. As a result of increased inference rate and resolution, artifact filtering/fusing can conclude faster before the robot moves out of view, and do so with a lower false-positive rate. 

Over four runs at the Tunnel Circuit Event, in which our team took 4th place, we successfully scored 17 artifacts and falsely reported 7 artifacts which did not correspond to a true artifact. Of the 17 successful reports, 8 were originally reported at an incorrect location as a result of orientation error in the initial robot to DARPA artifact frame transform computed at the gate, and required the Human Supervisor to adjust the location in order to score the points. Our ability to score points at the Urban circuit was limited due to mobility issues, which prevented us from reaching many artifacts.

\textbf{Artifact Fusion} - Despite training the network in the presence of various conditions (motion blur, dim/intermittent lighting, camera angle, robot platform, etc), the classifier network cannot completely capture these effects, and so the confidence of a single classification can vary. Therefore, during this filtering/fusing step, we require a sequence of consistent positive classifications in order to file a single artifact report. This approach enables us to train the network using typical metrics and then independently tune the filtering/fusing such that the false-positive rate is within an acceptable range for a given robot in a given scenario (rather than requiring us to re-train the network several times). For example, during the Tunnel and Urban circuit events, our Human Supervisor filters incoming artifact reports before submission, and so a higher false-positive rate reduces the chances of missing an artifact; however, during virtual circuit events, the limited number of artifact reports are autonomously submitted, and so a low false-positive rate is necessary. Nevertheless, in the spirit of the competition, we tune our algorithm such that only highly confident artifact detections are forwarded to the base station for reporting.

\textbf{Camera Configurations and Algorithm Modularity} - As we progressed through the challenges, we learned that each new subterranean environment presented new challenges for detecting artifacts. Originally, our artifact detection solution was completely coupled with the use of a Realsense D435, which the team found to be unreliable. The team decided to pursue an artifact detection and localization solution that was agnostic to camera and depth sensor selections. We obtain an artifact depth estimate by projecting the located artifact in 2D image space into the generated 3D OctoMap. Depth computation is done by casting a ray towards the direction of the center of a bounding box produced by the YOLO V3 model and returning the first intersecting voxel. By using an OctoMap, the framework is agnostic to the type of depth sensor used. However, our depth accuracy is then limited by the resolution of the OctoMap. Based on the competition accuracy requirement of 5m, the resolution limitation was accepted.  By using our projection framework, the system operates with any calibrated RGB camera that has a known transformation to the robot's map frame. This flexibility enables us to test our artifact detection pipeline on any RGB camera setup without any significant software changes and more readily adapt to changing subterranean environments.

\subsection{Localization}
\label{sec:LL_loc}
\textbf{Robustness in Austere Environments} - One of the major challenges in the DARPA Subterranean Challenge is achieving reliable localization performance in austere environments. Vision-based solutions have traditionally dominated this space \cite{leutenegger_keyframe-based_2015,10.1007/978-3-319-50115-4_66,qin_vins-mono:_2017} due to the maturity of computer vision feature extraction \cite{Cheung2009,Bay2008,Zhan_2018_CVPR}. It was expected that the maturity of visual-inertial SLAM packages \cite{10.1007/978-3-319-50115-4_66,qin_vins-mono:_2017} would translate to greater algorithm stability. However, we learned that performance in the lab did not transfer to the field. At the SubT Integration Exercise (STIX), visual-inertial SLAM was unstable at times, losing tracking and entering unrecoverable states. Specular highlights caused by reflective surfaces and inadequate lighting, as well as feature-poor scenes, were chiefly responsible for tracking failures. Because the robot fleet needs to be able to localize in a wide variety of previously-unknown austere environments, reliance on visual sensing modality was deemed too large a mission risk. While research in overcoming these issues is ongoing \cite{ovio}, agents were transitioned to LiDAR-inertial localization strategies. Field testing has shown that relying on visual or LiDAR sensing modalities alone may not be sufficient to localize a robot through fog, dust, or smoke. Recent work focusing on multi-modal sensor fusion \cite{khattak_complementary_2020} and robust state estimation \cite{santamaria-navarro_towards_2020} are inspiring for future research directions. 

\textbf{Computational Efficiency} -  During the Tunnel and Urban circuit events, the team leveraged a 2D version of Cartographer, constraining the vehicle to a plane, and making it computationally faster. However, during hour-long missions in more complex environments, Cartographer 3D faced computational limitations with its loop-closure scan-to-map global optimization routine. Minute-scale latency during field tests was not compatible with the rapid mission tempo, and sometimes a lack of convergence led to large uncorrected heading errors. More recently, LIO-SAM \cite{shan_liosam_2020} pushes toward higher real-time performance by reducing raw scans to planar and edge features similar to LOAM \cite{zhang_low-drift_2017}, and simplifying loop closure optimization via scan-to-scan matching. Since real-time localization is the backbone of our autonomy stack, agents may be transitioned to LIO-SAM if proven reliable in a diverse array of underground environments.

\textbf{Multi-Robot Alignment and Localization} - 
When deploying multiple robots, there are several advantages to having them coordinate and share information, as discussed in Section \ref{sec:multiagent}. Sharing information is often more beneficial if the information shares a common reference frame.  DARPA provides several options for localizing against a global reference frame including Apriltags \cite{Malyuta2019,Brommer2018,Wang2016}, generic retro-reflectors, and survey reflectors. Transforms from the targets to the origin are provided by DARPA. The individual teams are responsible for sensing the various targets and using them to align their robots to the global reference frame. Of the provided options, the survey reflectors gave the most accurate and consistent transforms for our systems. Separate analyses concluded that the position variance of Leica prisms is on the order of millimeters \cite{prism_accuracy}, consistent with the centering accuracy specified by Leica \cite{leica_survey_prisms}, whereas Apriltag variance is on the order of centimeters \cite{apriltag_accuracy}. Each system is equipped with a set of three survey prisms, and both the robot and gate transforms are estimated in the survey station's frame using Horn's absolute orientation method \cite{horn1987closed}. This information is used to create an estimate of the robot's pose in the global reference frame.

Multi-robot alignment requires highly accurate orientations; as the distance increases from the origin, minor orientation errors in initial map to world transformation can lead to increasing position errors relative to the world frame across robots. Across ten tests surveying a transform estimate from the same three prisms our chosen method showed a standard deviation between the transforms of $0.06\deg$ of roll, $0.05\deg$ of pitch, and $0.09\deg$ of yaw and an average roll and pitch estimate of  $-0.07\deg$ and $0.05\deg$ respectively relative to a gravity aligned global frame. Horn's method outperformed numerical methods used during the Urban circuit event, which varied significantly despite relatively small differences in the robot's initial orientations. This was due to local minimum inherent to the non-convexity of rotation estimates. The survey prism-based alignment also outperformed estimates of alignment using Apriltags locations derived from the robot's forward-facing camera. Pose estimates generate from Apriltags had high variance since the robots’ front facing camera could only localize all three Apriltags far from any tag and the camera viewed the left and right Apriltags with yaw from the tags center. Both a sensor’s distance \cite{Wang2016} and yaw angle \cite{apriltag_accuracy} are known to decrease estimated pose accuracy, and even these relatively minor effects led to significant pose estimate errors. For the first Tunnel Circuit deployment ICP alignment conducted after the fact estimated the difference between the first robot's map and the third robot's map in the world frame was $0.0527\deg$ of roll, $3.9215\deg$ of pitch, and $-0.4105\deg$ of yaw in the worst case.

\subsection{Mapping}
\label{sec:mapping_lessons}
\textbf{Map Merging Efficiency} - As discussed in Section \ref{sec:mapping}, the ability to merge maps is critical to our multi-agent strategy. Originally, we used a custom merging node that accepted OctoMaps from the multi-agent node as well as the standard OctoMap Server running locally on the robot. A full merge of each map was performed every second. This proved to be computationally intensive as the map size increased, and so a method of comparing the growth of the maps was added to eliminate unnecessary map merges. If the map of a neighboring robot stopped growing, because the robot stopped moving or it lost connection with the communications network, that map was no longer merged, as it was already part of the local merged map. This drastically reduced the computational requirements. We later developed the diff map solution because this solution still had high computational requirements and more importantly, high communications requirements. Our original design transmitted the entire map for each robot every second, while our diff map aware node only transmitted small differences maps every 10 seconds. Compared to transmitting the full map at 10 second intervals, diff maps used an average of 1/100 the bandwidth of our full-map solution.

\begin{figure}[t!]
		\centering
		\subfloat[]{{\includegraphics[width=.47\textwidth]{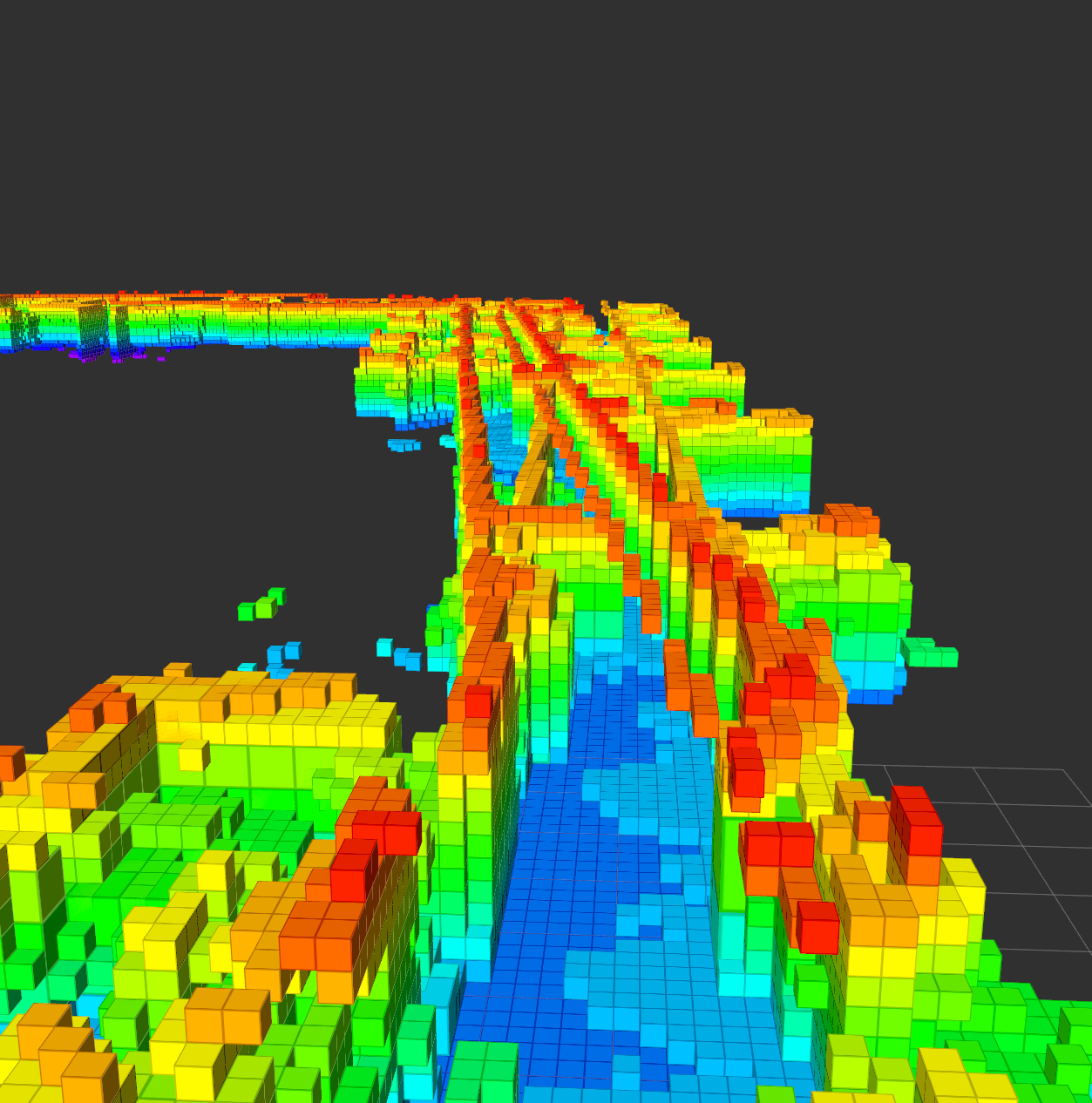}\label{fig:urban_full_merge_3d} }}
		\hspace{10pt}
		\subfloat[]{{\includegraphics[width=.47\textwidth]{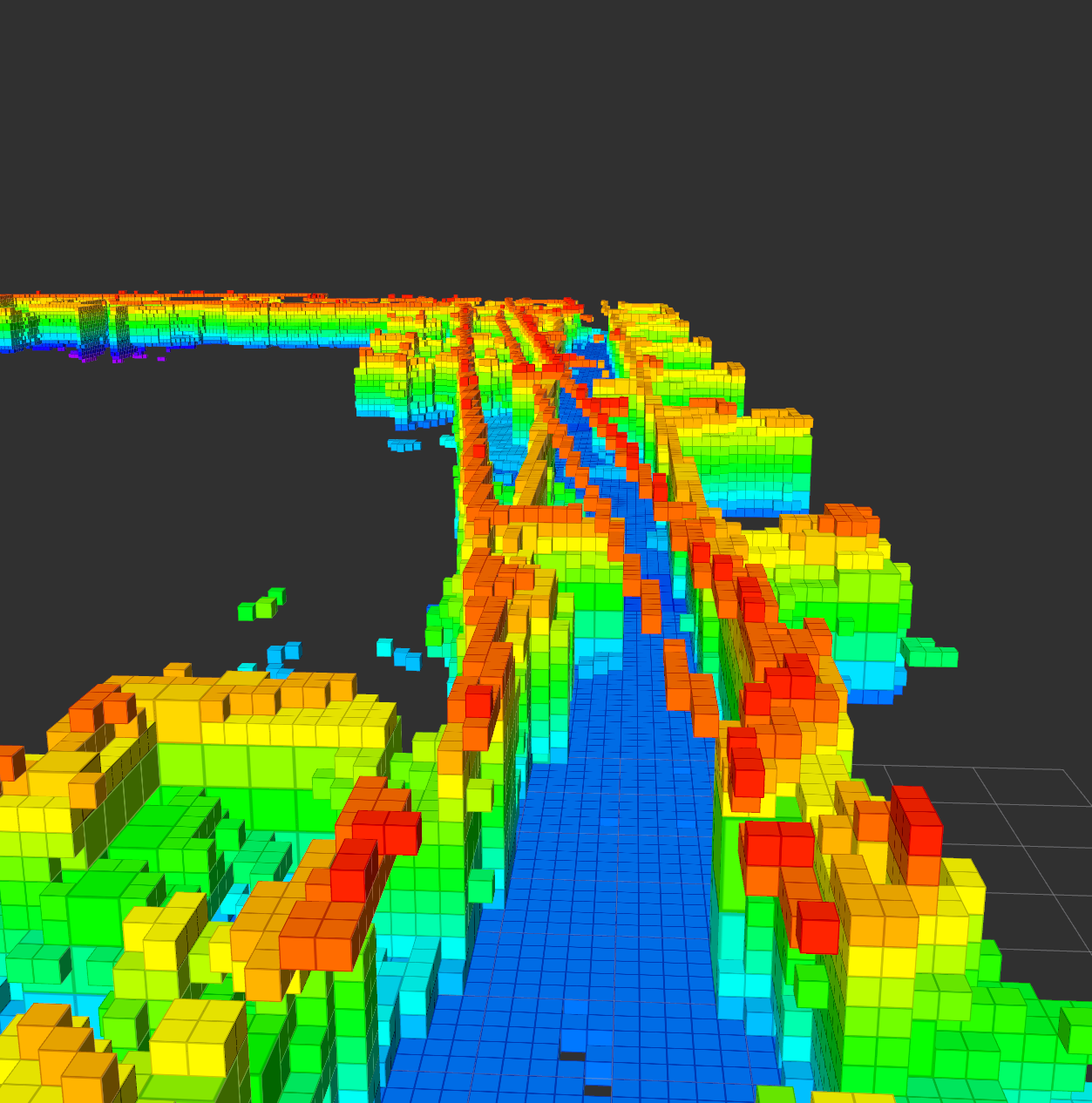}\label{fig:urban_priority_merge_3d} }}
		\caption{Example of merged maps: (a) without prioritization (b) with self-prioritization.}
		\label{fig:urban_merge}
\end{figure}

\textbf{Self-Prioritization} - Maps were not re-aligned prior to merging, as it was assumed the robots were already in a common reference frame with small error. We were using a voxel size of $.15m$ for the maps, and so it was necessary for each robot to have an initial world to map transform with under $.075m$ of translation error to ensure map data is merged properly. However, during testing prior to competition, it was found that small alignment errors could cause obstacles from a neighboring map to block open space for the robot that performed the merge, and thus preventing the robot from planning any further. A self-prioritization feature was added to combat this failure mode. Instead of a full merge, only an append of new information would be performed. In essence, the robot would generate two maps, one using its own sensor data as a truth source, and another merged map where it only appended cells from other maps in “unknown” areas, and never overwrote its own “seen” data. This would prevent misalignment, most frequently caused by poor gate localization, from creating obstacles over free space.

During the Urban circuit event, an error in gate detection resulted in almost exactly the exact scenario used in simulation during development. In Figure \ref{fig:urban_full_merge_3d}, the two maps are fully merged, and it can be seen that the first robot (the left corridor) creates a wall in the middle of the second robot’s map. This would have prevented the robot from being able to plan a path through the hallway. Figure \ref{fig:urban_priority_merge_3d} shows the self-prioritized map where the robot is trusting its own representation of free space more than merged map data. The first map still can be seen outside of the walls, but it does not hinder the robot from navigating down the hallway. A future opportunity is to leverage iterative closest point (ICP) techniques to align the map of a neighboring agent prior to merging.

\subsection{Navigation}
\label{sec:LL_nav}
\textbf{Graph-Based Navigation} - The most common failure mode leading to a stuck robot was when the graph-based planner generated an edge that the local centering controller could not navigate over. Often, this was a result of rough terrain or an inconsistency in the map. If the vehicle attempted to traverse the edge and failed, but the edge is left as unexplored, the system could enter a loop of attempting to explore an edge that is not traversable. This exact situation occurred a few times during the Tunnel Circuit event, and resulted in the agents wasting valuable exploration time. Although the fleet moved on from this approach in the Urban circuit event, this fault could be remedied by maintaining a list of recently visited vertices and the edges left and arrived on. Then, if an edge could not be traversed by the centering controller, the planner would ignore it as a possible destination. Overall, this approach is very successful in the NIOSH competition environment and in the Edgar Experimental Mine due to their graph-like structures.

\textbf{Planner and Controller Disagreements} - We have learned that one of the most challenging aspects of path planning for ground vehicles in cluttered, subterranean environments is generating safe paths. We noticed during initial testing of the global planner describer in Section \ref{sec:contfront} that the generated global paths to the goal points were sensible and avoided large obstacles in the map. However, they did not take into account different aspects of the terrain like small to medium size rocks or raised sections of concrete. As a result, the robot could wind up physically stuck if it followed a path over terrain it actually could not navigate. To solve this problem, our team developed a reactive controller based around high density depth sensor data of the ground immediately in front of the robot. This helped the robot avoid obstacles not accounted for by the global planner, or stop the vehicle if it was too close to a hazardous obstacle. Similarly, our aerial platforms required the development of additional obstacle avoidance capabilities in order to effectively follow planned paths safely through cluttered environments. The vision-based local controller presented in Section \ref{sec:AVnav} has proved to be an adequate solution for detecting and avoiding small obstacles along planned paths. The team is working on improving the current planning algorithm to produce more terrain aware paths for ground vehicle navigation. The main lesson here is that path planning algorithms developed for cluttered subterranean environments needs to consider the inherent limitations in mobility and sensing of the platform in order to ensure safe and traversable paths. 

\textbf{Path Stitching}
When planning over a large horizon, the robot may travel a significant distance in the time it takes our fast-marching based planning algorithm to generate a feasible path. We learned that this can cause undesirable motion when the system transitions from following the old path to the new path, depending on how far the robot has deviated from the new path in the time it has taken to plan it. As a result, it's necessary to plan from where the robot is expected to be as opposed to where it is when the planning process begins.  To this end, a receding horizon path stitching algorithm has been added to plan from where the robot will be if it continues to follow the previous path over a short time horizon, $\delta t_\text{horizon}$.  This process is detailed in Algorithm \ref{algorithm:RecedingHorizon}.  Here the stitching algorithm takes as input the current map, $\mathcal{M}$, robot pose, $\bm{x}$, robot speed, $u$, and stitching horizon and outputs a path at regular planning intervals that contains a segment of the previous path, path$_{RH,-2\Delta t_\text{horizon}:\bm{x}_{RH}}$, connected to the new path segment planned from a stitch point, $\bm{x}_\text{RH}$ close to where to robot will be in $\delta t_\text{horizon}$.

\begin{algorithm}
\begin{algorithmic}[1]
  \STATE path$_{RH} \gets \{\}$ \\
  \STATE $[\text{path}, \bm{p}_\text{goal}] \gets $FrontierViewPlanner$(\mathcal{M}, \bm{x}, \text{...})$\\
  \STATE path$_{RH} \gets $path \\
  \WHILE{True}
    \STATE $\bm{x}_\text{RH} \gets $FindLookahead$(\text{path}, \bm{p}_\text{goal}, \bm{x}, u, \Delta t_\text{horizon})$\\
    \STATE $[\text{path}, \bm{p}_{goal}] \gets $FrontierViewPlanner$(\mathcal{M}, \bm{x}_\text{RH}, \text{...})$\\
    \STATE path$_{RH} \gets [$path$_{RH,-2\Delta t_\text{horizon}:\bm{x}_\text{RH}}; $path$; \bm{p}_\text{goal}]$ \\
  \ENDWHILE
\end{algorithmic}
\caption{RecedingHorizonStitching$(\mathcal{M},\bm{x}, u, \Delta t_\text{horizon})$}
\label{algorithm:RecedingHorizon}
\end{algorithm}

\subsection{Communications}

An initial review of the difficulties in underground communication \cite{zhang2001radio,boutin2008radio} led us to a strategy of ``backbone", where the goal was to create a basic communication network stemming out from the base station. Agents needed to be completely self-sufficient in a lost-comm condition. In general, each mobile vehicle would explore until certain criteria are met (artifact discovered, etc), returning to communications range when needed. This strategy allowed us to accept an incomplete coverage map and provides a natural progression of capability through the use of deployable relay beacons to extend the communications backbone back to the human operator. Our communications solution has undergone three versions to date, corresponding to Circuit Events.

\begin{figure}[t!]
    \centering
    \includegraphics[width=0.9\textwidth]{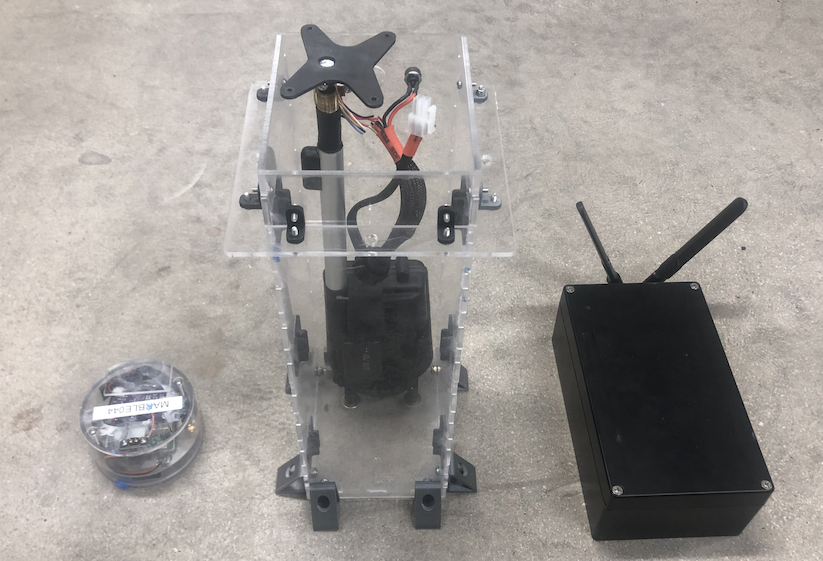}
    \caption{Custom communications beacons designed and built for the (left) Tunnel, (middle) Urban, and (right) Cave circuit events.}
    \label{fig:beacon_evolution}
\end{figure}

\textbf{Version 1: Tunnel} - The initial beacon design for the Tunnel circuit was designed to be small and disposable. Small housings with minimal elevation (Figure \ref{fig:beacon_evolution}, left) were used to house a low-power radio. Meshing functions were provided by the open-source \textit{B.A.T.M.A.N} stack in a custom OpenWRT-based firmware. While other radios with superior properties from Silvus, Rajant, and others were considered, cost considerations necessitated a commodity solution. To function effectively within the backbone strategy, reconnect times were required to be as quick as possible in response to a topology change, which occurs when a robot returns to communication range. \textit{B.A.T.M.A.N} is not optimized for this use case, necessitating a replacement mesh solution. Further, the low-power radios paired with quarter-wave antennas resulted in poor network performance, leading to observed behavior of robots returning to the Base 
Station frequently to report artifacts. 

\textbf{Version 2: Urban} - Version 2 beacons shown in the center of Figure \ref{fig:beacon_evolution} consisted of an extensible power mast with half-wave antennas, a battery, a high-power rugged radio, a single-board computer and support electronics all enclosed in an acrylic case. Meshing was provided by a custom mesh networking system called \textit{meshmerize} that prioritizes reconnection times instead of optimal routing. Beyond online coordination with the base station, an offline command and data relay system drawn from UAV studies \cite{frew2009networking} was implemented onboard the beacons' single-board computer. Version 2 beacons are more capable than Version 1, however several operational problems became apparent in testing and competition. The extensible antenna necessitated a minimum vertical size of the beacon, which had several negative consequences downstream; in particular, the size of the beacon had a negative impact on maximum traversable pitch of the carrier robot. The new meshing solution performed extremely well, transferring well over a gigabyte of data over the course of one particular one-hour run and allowing ground vehicles to act as relays for one another with reconnection times on the order of milliseconds. 

\textbf{Version 3: Cave / Final} - The Version 3 beacons were designed to be more compact and still retain high-level functions, resulting in the beacon in the far-right of Figure \ref{fig:beacon_evolution}. Instead of acrylic, a NEMA-rated aluminum enclosure was used to provide durability and `puddle-proof' water intrusion protection. The radio is cooled through a thermally conductive bonding to the metal lid of the case. The extensible mast and support electronics were removed and replaced with a PCB to facilitate faster assembly of new beacons; existing radios, battery, and single-board computer components were re-used. 

Initial testing with the \textit{udp\_mesh} transport layer, described in Section \ref{sec:comms}, and \textit{meshmerize} mesh layer have yielded performance improvements compared to our original solutions. In one lab test to determine maximum performance, we observed that our ROS message throughput is limited by the underlying \textit{meshmerize} mesh layer at approximately 20 Mbit/s. Operational testing with high and low priority data streams reveals that \textit{udp\_mesh} has eliminated the previously observed latency issues, an improvement over our previous implementation. As a robot explores an environment, the telemetry and artifact reports are nearly instantaneous, while map portions fill in as bandwidth allows; these tests validate both the efficacy and usefulness of our prioritization method. Moreover, when robots returned to within communications range, we observed reconnection times on the order of one second, in stark contrast to the tens of seconds that \textit{B.A.T.M.A.N.} required.  Mechanically, by reducing the beacon size, previous issues with vehicle pitch handling have been eliminated.

\subsection{Multi-Agent Coordination}

\textbf{Multi-agent Communications Hopping} - The ability to store messages and relay communications through other robots greatly improved information sharing across the network, and demonstrated one of the benefits of multiple robots even without coordination. On multiple occasions, robots were able to transmit artifact reports back to the base station through a ``hop", and then continue exploring deeper into the environment.

\textbf{Data Prioritization} - Field deployments provided meaningful feedback regarding communication bottlenecks within the fleet. The original multi-agent system consisted of low-bandwidth messages similar to those described in Section \ref{sec:multiagent} as well as high-bandwidth messages, which included merged OctoMaps and artifact image data. This technique works well in short-range, high-bandwidth environments, but when bandwidth is highly constricted, the larger data messages prevented priority data such as artifact reports from being delivered. As a consequence, the multi-agent framework was streamlined to use the map diffs described in Section \ref{sec:mapping} and use a combination of broadcast, for low-bandwidth data, and point-to-point messages, for map diffs and artifact images. These changes led to reduced traffic and more reliable transmission, with success rate for artifact image transmission increasing from 30\% to 100\%. In most situations this hybrid approach is a good compromise and still allows the robots to tolerate intermittent communications connectivity.

\subsection{Human - Robot Interface}

\textbf{Human Interactions} - 

As stated in the Introduction, teams are allowed a single human supervisor that is allowed to interact with the robots after they have been deployed into the environment. As the number of robots and functionality of each robot increases, the potential load on the human supervisor increases as well. Reducing the reliance of the fleet on the human operator is challenging. A solution to this challenge is presented in \cite{Olson2012}, where a system for controlling a fleet of 14 robots is presented. Interactions between a human supervisor and differing fleets of robots is explored in \cite{Wang2007}. Providing an interface that is user friendly is also key to improving the management of a fleet of robots \cite{Nevatia2008}.

For our solution, early design decisions focused on very limited human interaction with the robots, as the fleet relied on a fully autonomous solution. However, during field deployments, it became clear that certain scenarios can benefit swift Human Supervisor intervention. The information transfer from the robot fleet to the DARPA command post was originally designed as a rigid process. However, during the first field deployment at the Tunnel circuit event, the Human Supervisor detected that robots were sometimes reporting erroneous artifact locations and types. Adding flexibility into the reporting system provided the Human Supervisor the opportunity to discern reliable information from unreliable information, and ultimately modify that information before sending it up the chain of command. Additionally, functionality was added to allow the Human Supervisor control over the robots via a joystick as well as  manual goal point in the GUI. While manual control is not desired for normal operation, there have been unexpected events in which it was the only method of recovering an agent and resuming exploration. Manual goal points can be useful when the Human Supervisor identifies an area of interest for the robot to inspect further.

\textbf{Remote Oversight} - As the fleet grew in size and complexity, the workload of the Human Supervisor increased substantially. The GUI and RViz were modified to provide dedicated information on each agent, as well as a main window with all agents in the merged map. With the introduction of the survey operator during robot startup, it became valuable for the Human Supervisor to have a visualization, in order to verify accurate gate localization. These seemingly small improvements worked to drastically reduce the Human Supervisor workload while providing greater oversight of the robot fleet and the overall mission.

\section{Conclusion}

This paper has presented our continued work to address many open areas of robotics research including platforms design, object detection and classification, path planning and navigation, multi-agent coordination, communications, and human-robot interaction. Specifically, we have presented several novel extensions of state-of-the-art research and a complete field robotics solution for multi-agent autonomous subterranean exploration and rapid situational awareness. 

The graph-based exploration strategy detailed in Section \ref{sec:metrictopo} performed well in the narrow, tunnel-like environments found in Edgar Experimental Mine and NIOSH Safety Mine. The frontier-based exploration strategy from Section \ref{sec:contfront} also performed well in the Edgar Experimental Mine, and allowed agents to explore larger environments like the GeoTech warehouse and the SATSOP nuclear power facility. The team has learned that we need to improve our terrain awareness when planning paths through cluttered environments. The circuit events have presented challenging communication environments and our team has learned to overcome network saturation and reliability problems through the implementation of a UDP-Mesh communication solution presented in Section \ref{sec:comms}. Sharing information across agents is critical to improving exploration efficiency and providing situational awareness, and generating and transmitting map differences versus entire maps decreases the risk of network saturation as demonstrated in Section \ref{sec:mapping}. Artifacts in the competition are generally spaced far apart, making it difficult for a single agent to find them all in the time limit. By sharing map differences and candidate goal points through our multi-agent coordination scheme from Section \ref{sec:multiagent}, agents are now able to coordinate their exploration, ultimately increasing the amount of space seen and the potential to find artifacts. 

One of the most significant lessons from this research effort is that developing robust autonomous platforms for rapid situational awareness in subterranean environments requires significant and frequent field testing. In isolated lab testing, it is nearly impossible to recreate all conditions inherent to underground environments that make multi-agent autonomy so challenging. An integrative approach involving frequent submodule and full-system evaluations in subterranean environments of representative topology, complexity, and scale is invaluable to the development of robust multi-agent autonomy.

\subsection*{Acknowledgments}
This work was supported through the DARPA Subterranean Challenge, cooperative agreement number HR0011-18-2-0043. A special thanks to Daniel Torres, Cesar Galant, Zoe Turin, for assisting with the design, testing, and deployments of our platforms and algorithms. Thank you to the Colorado School of Mines and the Edgar Experimental Mine for allowing us to conduct mock deployments in the mine. Special thanks also to Simon Wunderlich and his team at Meshmerize GmbH for their mesh networking support. Finally, thank you to all of the DARPA staff who have planned and executed absolutely incredible Subterranean Challenge system track circuit events.

\newpage
\bibliographystyle{apalike}
\bibliography{Refs}

\begin{thebibliography}{}

\bibitem[Abbas et~al., 2019]{apriltag_accuracy}
Abbas, S.~M., Aslam, S., Berns, K., and Muhammad, A. (2019).
\newblock Analysis and improvements in apriltag based state estimation.
\newblock {\em Sensors}, 19(24).

\bibitem[Ahmad and Fierro, 2019]{ahmad2019real}
Ahmad, S. and Fierro, R. (2019).
\newblock Real-time quadrotor navigation through planning in depth space in
  unstructured environments.
\newblock In {\em 2019 International Conference on Unmanned Aircraft Systems
  (ICUAS)}, pages 1467--1476. IEEE.

\bibitem[Ahmad et~al., 2021a]{ahmad20213d}
Ahmad, S., Mills, A.~B., Rush, E.~R., Frew, E.~W., and Humbert, J.~S. (2021a).
\newblock 3d reactive control and frontier-based exploration for unstructured
  environments.
\newblock In {\em 2021 IEEE/RSJ International Conference on Intelligent Robots
  and Systems (IROS)}. IEEE.

\bibitem[Ahmad et~al., 2021b]{ahmad2021apf}
Ahmad, S., Sunberg, Z.~N., and Humbert, J.~S. (2021b).
\newblock Apf-pf: Probabilistic depth perception for 3d reactive obstacle
  avoidance.
\newblock In {\em 2021 American Control Conference (ACC)}, pages 32--39. IEEE.

\bibitem[Ahmad et~al., 2021c]{ahmad2021depth}
Ahmad, S., Sunberg, Z.~N., and Humbert, J.~S. (2021c).
\newblock End-to-end probabilistic depth perception and 3d obstacle avoidance
  using pomdp.
\newblock {\em Journal of Intelligent \& Robotic Systems}, 103(2):1--18.

\bibitem[Alvarez et~al., 2019]{Escobar2019}
Alvarez, H.~E., Ohradzansky, M., Keshavan, J., Ranganathan, B., and Humbert,
  J.~S. (2019).
\newblock Bio-inspired approaches for small-object detection and avoidance.
\newblock {\em IEEE Transactions on Robotics}.

\bibitem[Bay et~al., 2008]{Bay2008}
Bay, H., Ess, A., Tuytelaars, T., and Van~Gool, L. (2008).
\newblock Speeded-up robust features ({SURF}).
\newblock {\em Comput. Vis. Image Underst.}, 110(3):346–359.

\bibitem[Bircher et~al., 2016]{bircher2016receding}
Bircher, A., Kamel, M., Alexis, K., Oleynikova, H., and Siegwart, R. (2016).
\newblock Receding horizon" next-best-view" planner for 3d exploration.
\newblock In {\em 2016 IEEE international conference on robotics and automation
  (ICRA)}, pages 1462--1468. IEEE.

\bibitem[Bormann et~al., 2016]{Bormann2016}
Bormann, R., Jordan, F., Li, W., Hampp, J., and H{\"{a}}gele, M. (2016).
\newblock {Room segmentation: Survey, implementation, and analysis}.
\newblock In {\em Proceedings of the IEEE International Conference on Robotics
  and Automation}, pages 1019--1026, Stockholm, Sweden.

\bibitem[Bouman et~al., 2020]{bouman2020autonomous}
Bouman, A., Ginting, M.~F., Alatur, N., Palieri, M., Fan, D.~D., Touma, T.,
  Pailevanian, T., Kim, S.-K., Otsu, K., Burdick, J., and akbar Agha-mohammadi,
  A. (2020).
\newblock Autonomous spot: Long-range autonomous exploration of extreme
  environments with legged locomotion.

\bibitem[Boutin et~al., 2008]{boutin2008radio}
Boutin, M., Benzakour, A., Despins, C.~L., and Affes, S. (2008).
\newblock Radio wave characterization and modeling in underground mine tunnels.
\newblock {\em IEEE Transactions on Antennas and Propagation}, 56(2):540--549.

\bibitem[Brass et~al., 2011]{Brass2011}
Brass, P., Cabrera-Mora, F., Gasparri, A., and Xiao, J. (2011).
\newblock {Multirobot tree and graph exploration}.
\newblock {\em IEEE Transactions on Robotics}, 27(4):707--717.

\bibitem[Brommer et~al., 2018]{Brommer2018}
Brommer, C., Malyuta, D., Hentzen, D., and Brockers, R. (2018).
\newblock Long-duration autonomy for small rotorcraft {UAS} including
  recharging.
\newblock In {\em {IEEE}/{RSJ} International Conference on Intelligent Robots
  and Systems}, page arXiv:1810.05683. {IEEE}.

\bibitem[Burgard et~al., 2002]{Burgard2002}
Burgard, W., Moors, M., and Scheider, F. (2002).
\newblock {\em {Collaborative Exploration of Unknown Environments with Teams of
  Mobile Robots}}, volume 2466 of {\em Lecture Notes in Computer Science}.
\newblock Springer Berlin Heidelberg, Berlin, Heidelberg.

\bibitem[Charrow et~al., 2015]{charrow2015information}
Charrow, B., Liu, S., Kumar, V., and Michael, N. (2015).
\newblock Information-theoretic mapping using cauchy-schwarz quadratic mutual
  information.
\newblock In {\em 2015 IEEE International Conference on Robotics and Automation
  (ICRA)}, pages 4791--4798. IEEE.

\bibitem[{Cheung} and {Hamarneh}, 2009]{Cheung2009}
{Cheung}, W. and {Hamarneh}, G. (2009).
\newblock n0-sift: n -dimensional scale invariant feature transform.
\newblock {\em IEEE Transactions on Image Processing}, 18(9):2012--2021.

\bibitem[Dang et~al., 2019]{dang2019explore}
Dang, T., Khattak, S., Mascarich, F., and Alexis, K. (2019).
\newblock Explore locally, plan globally: A path planning framework for
  autonomous robotic exploration in subterranean environments.
\newblock In {\em 2019 19th International Conference on Advanced Robotics
  (ICAR)}, pages 9--16. IEEE.

\bibitem[Dang et~al., 2020]{Dang2020}
Dang, T., Tranzatto, M., Khattak, S., Mascarich, F., Alexis, K., and Hutter, M.
  (2020).
\newblock Graph-based subterranean exploration path planning using aerial and
  legged robots.
\newblock {\em Journal of Field Robotics}, 37(8):1363--1388.

\bibitem[DARPA, 2021]{DARPA2021}
DARPA (2021).
\newblock {DARPA Subterranean Challenge}.
\newblock https://subtchallenge.com.

\bibitem[Debord et~al., 2018]{debord2018trajectory}
Debord, M., H{\"o}nig, W., and Ayanian, N. (2018).
\newblock Trajectory planning for heterogeneous robot teams.
\newblock In {\em 2018 IEEE/RSJ International Conference on Intelligent Robots
  and Systems (IROS)}, pages 7924--7931. IEEE.

\bibitem[Dharmadhikari et~al., 2020]{dharmadhikari2020motion}
Dharmadhikari, M., Dang, T., Solanka, L., Loje, J., Nguyen, H., Khedekar, N.,
  and Alexis, K. (2020).
\newblock Motion primitives-based path planning for fast and agile exploration
  using aerial robots.
\newblock In {\em 2020 IEEE International Conference on Robotics and Automation
  (ICRA)}, pages 179--185. IEEE.

\bibitem[Doucet et~al., 2000]{doucet2000sequential}
Doucet, A., Godsill, S., and Andrieu, C. (2000).
\newblock On sequential monte carlo sampling methods for bayesian filtering.
\newblock {\em Statistics and computing}, 10(3):197--208.

\bibitem[Dubey et~al., 2017]{dubey2017droan}
Dubey, G., Arora, S., and Scherer, S. (2017).
\newblock Droan—disparity-space representation for obstacle avoidance.
\newblock In {\em 2017 IEEE/RSJ International Conference on Intelligent Robots
  and Systems (IROS)}, pages 1324--1330. IEEE.

\bibitem[Dudek et~al., 1991]{Dudek1991}
Dudek, G., Jenkin, M., Milios, E., and Wilkes, D. (1991).
\newblock {Robotic Exploration as Graph Construction}.
\newblock {\em IEEE Transactions on Robotics and Automation}, 7(6):859--865.

\bibitem[Ebadi et~al., 2020]{Ebadi2020}
Ebadi, K., Chang, Y., Palieri, M., Stephens, A., Hatteland, A., Heiden, E.,
  Thakur, A., Funabiki, N., Morrell, B., Wood, S., Carlone, L., and
  Agha-Mohammadi, A.~A. (2020).
\newblock {LAMP: Large-Scale Autonomous Mapping and Positioning for Exploration
  of Perceptually-Degraded Subterranean Environments}.
\newblock {\em IEEE International Conference on Robotics and Automation}, pages
  80--86.

\bibitem[Englot and Hover, 2013]{englot2013three}
Englot, B. and Hover, F.~S. (2013).
\newblock Three-dimensional coverage planning for an underwater inspection
  robot.
\newblock {\em The International Journal of Robotics Research},
  32(9-10):1048--1073.

\bibitem[Farhadi and Redmon, 2018]{farhadi2018yolov3}
Farhadi, A. and Redmon, J. (2018).
\newblock Yolov3: An incremental improvement.
\newblock {\em Computer Vision and Pattern Recognition, cite as}.

\bibitem[Frew and Brown, 2009]{frew2009networking}
Frew, E.~W. and Brown, T.~X. (2009).
\newblock Networking issues for small unmanned aircraft systems.
\newblock {\em Journal of Intelligent and Robotic Systems}, 54(1-3):21--37.

\bibitem[Gao et~al., 2018]{gao2018online}
Gao, F., Wu, W., Lin, Y., and Shen, S. (2018).
\newblock Online safe trajectory generation for quadrotors using fast marching
  method and bernstein basis polynomial.
\newblock In {\em 2018 IEEE International Conference on Robotics and Automation
  (ICRA)}, pages 344--351. IEEE.

\bibitem[Gerkey and Mataric, 2002]{gerkey2002}
Gerkey, B. and Mataric, M. (2002).
\newblock {Sold!: auction methods for multirobot coordination}.
\newblock {\em IEEE Transactions on Robotics and Automation}, 18(5):758--768.

\bibitem[Girshick, 2015]{girshick2015fast}
Girshick, R. (2015).
\newblock Fast r-cnn.
\newblock In {\em Proceedings of the IEEE international conference on computer
  vision}, pages 1440--1448.

\bibitem[Griffiths et~al., 2006]{Griffiths2006MaximizingMA}
Griffiths, S., Saunders, J.~D., Curtis, A., Barber, D.~B., McLain, T.~W., and
  Beard, R. (2006).
\newblock Maximizing miniature aerial vehicles obstacle and terrain avoidance
  for mavs.
\newblock {\em IEEE Robotics and Automation Letters}.

\bibitem[Harms et~al., 2017]{harms2017development}
Harms, H., Schmiemann, J., Schattenberg, J., and Frerichs, L. (2017).
\newblock Development of an adaptable communication layer with {QoS}
  capabilities for a multi-robot system.
\newblock In {\em Iberian Robotics conference}, pages 782--793. Springer.

\bibitem[Hess et~al., 2016]{hess2016}
Hess, W., Kohler, D., Rapp, H., and Andor, D. (2016).
\newblock Real-time loop closure in 2d lidar slam.
\newblock In {\em 2016 IEEE International Conference on Robotics and Automation
  (ICRA)}, pages 1271--1278.

\bibitem[Horn, 1987]{horn1987closed}
Horn, B.~K. (1987).
\newblock Closed-form solution of absolute orientation using unit quaternions.
\newblock {\em Josa a}, 4(4):629--642.

\bibitem[Hornung et~al., 2013]{hornung13octomap}
Hornung, A., Wurm, K.~M., Bennewitz, M., Stachniss, C., and Burgard, W. (2013).
\newblock {OctoMap}: An efficient probabilistic {3D} mapping framework based on
  octrees.
\newblock {\em Autonomous Robots}.
\newblock Software available at \url{http://octomap.github.com}.

\bibitem[Howard, 2006]{Howard2006}
Howard, A. (2006).
\newblock {Multi-robot Simultaneous Localization and Mapping using Particle
  Filters}.
\newblock {\em The International Journal of Robotics Research},
  25(12):1243--1256.

\bibitem[Huang et~al., 2019]{Huang2019}
Huang, Y.-W., Lu, C.-L., Chen, K.-L., Ser, P.-S., Huang, J.-T., Shen, Y.-C.,
  Chen, P.-W., Chang, P.-K., Lee, S.-C., and Wang, H.-C. (2019).
\newblock {Duckiefloat: a Collision-Tolerant Resource-Constrained Blimp for
  Long-Term Autonomy in Subterranean Environments}.

\bibitem[{Kasper} et~al., 2019]{ovio}
{Kasper}, M., {McGuire}, S., and {Heckman}, C. (2019).
\newblock A benchmark for visual-inertial odometry systems employing onboard
  illumination.
\newblock In {\em 2019 IEEE/RSJ International Conference on Intelligent Robots
  and Systems (IROS)}, pages 5256--5263.

\bibitem[Khattak et~al., 2020]{khattak_complementary_2020}
Khattak, S., Nguyen, H., Mascarich, F., Dang, T., and Alexis, K. (2020).
\newblock Complementary {Multi}–{Modal} {Sensor} {Fusion} for {Resilient}
  {Robot} {Pose} {Estimation} in {Subterranean} {Environments}.
\newblock In {\em 2020 {International} {Conference} on {Unmanned} {Aircraft}
  {Systems} ({ICUAS})}, pages 1024--1029, Athens, Greece. IEEE.

\bibitem[Ko et~al., 2003]{Ko2003}
Ko, J., Stewart, B., Fox, D., Konolige, K., and Limketkai, B. (2003).
\newblock {A Practical, Decision-theoretic Approach to Multi-robot Mapping and
  Exploration}.
\newblock In {\em 2003 IEEE/RSJ International Conference on Intelligent Robots
  and Systems (IROS 2003)}, volume~4, pages 3232--3238, Las Vegas, NV.

\bibitem[Kroon, 2020]{kroon2020MSFM}
Kroon, D.-J. (2020).
\newblock Accurate fast marching.

\bibitem[Lackner and Lienhart, 2016]{prism_accuracy}
Lackner, S. and Lienhart, W. (2016).
\newblock Impact of prism type and prism orientation on the accuracy of
  automated total station measurements.
\newblock In {\em Joint International Symposium on Deformation MonitoringAt:
  Vienna}.

\bibitem[Lajoie et~al., 2020]{Lajoie2020}
Lajoie, P.~Y., Ramtoula, B., Chang, Y., Carlone, L., and Beltrame, G. (2020).
\newblock {DOOR-SLAM: Distributed, Online, and Outlier Resilient SLAM for
  Robotic Teams}.
\newblock {\em IEEE Robotics and Automation Letters}, 5(2):1656--1663.

\bibitem[Lee, 1982]{lee1982medial}
Lee, D.-T. (1982).
\newblock Medial axis transformation of a planar shape.
\newblock {\em IEEE Transactions on pattern analysis and machine intelligence},
  4(4):363--369.

\bibitem[{Leica Geosystems}, 2020]{leica_survey_prisms}
{Leica Geosystems} (2020).
\newblock Leica reflectors datasheet.
\newblock Rev. 712323-3.0.0en.

\bibitem[Leutenegger et~al., 2015]{leutenegger_keyframe-based_2015}
Leutenegger, S., Lynen, S., Bosse, M., Siegwart, R., and Furgale, P. (2015).
\newblock Keyframe-based visual–inertial odometry using nonlinear
  optimization.
\newblock {\em The International Journal of Robotics Research}, 34(3):314--334.

\bibitem[Littman, 1994]{littman1994witness}
Littman, M.~L. (1994).
\newblock The witness algorithm: Solving partially observable markov decision
  processes.
\newblock {\em Brown University, Providence, RI}.

\bibitem[Littman et~al., 1995]{littman1995learning}
Littman, M.~L., Cassandra, A.~R., and Kaelbling, L.~P. (1995).
\newblock Learning policies for partially observable environments: Scaling up.
\newblock In {\em Machine Learning Proceedings 1995}, pages 362--370. Elsevier.

\bibitem[Liu et~al., 2018]{liu2018search}
Liu, S., Mohta, K., Atanasov, N., and Kumar, V. (2018).
\newblock Search-based motion planning for aggressive flight in se (3).
\newblock {\em IEEE Robotics and Automation Letters}, 3(3):2439--2446.

\bibitem[Malyuta et~al., 2019]{Malyuta2019}
Malyuta, D., Brommer, C., Hentzen, D., Stastny, T., Siegwart, R., and Brockers,
  R. (2019).
\newblock Long-duration fully autonomous operation of rotorcraft unmanned
  aerial systems for remote-sensing data acquisition.
\newblock {\em Journal of Field Robotics}, page arXiv:1908.06381.

\bibitem[{Mascarich} et~al., 2020]{9172496}
{Mascarich}, F., {Nguyen}, H., {Dang}, T., {Khattak}, S., {Papachristos}, C.,
  and {Alexis}, K. (2020).
\newblock A self-deployed multi-channel wireless communications system for
  subterranean robots.
\newblock In {\em 2020 IEEE Aerospace Conference}, pages 1--8.

\bibitem[Matthies et~al., 2014]{matthies2014stereo}
Matthies, L., Brockers, R., Kuwata, Y., and Weiss, S. (2014).
\newblock Stereo vision-based obstacle avoidance for micro air vehicles using
  disparity space.
\newblock In {\em 2014 IEEE International Conference on Robotics and Automation
  (ICRA)}, pages 3242--3249. IEEE.

\bibitem[Maturana and Scherer, 2015]{maturana2015voxnet}
Maturana, D. and Scherer, S. (2015).
\newblock Voxnet: A 3d convolutional neural network for real-time object
  recognition.
\newblock In {\em 2015 IEEE/RSJ International Conference on Intelligent Robots
  and Systems (IROS)}, pages 922--928. IEEE.

\bibitem[Miller et~al., 2020]{Miller2020}
Miller, I.~D., Cohen, A., Kulkarni, A., Laney, J., Taylor, C.~J., Kumar, V.,
  Cladera, F., Cowley, A., Shivakumar, S.~S., Lee, E.~S., Jarin-Lipschitz, L.,
  Bhat, A., Rodrigues, N., and Zhou, A. (2020).
\newblock {Mine Tunnel Exploration Using Multiple Quadrupedal Robots}.
\newblock {\em IEEE Robotics and Automation Letters}, 5(2):2840--2847.

\bibitem[Nevatia et~al., 2008]{Nevatia2008}
Nevatia, Y., Stoyanov, T., Rathnam, R., Pfingsthorn, M., Markov, S., Ambrus,
  R., and Birk, A. (2008).
\newblock Augmented autonomy: Improving human-robot team performance in urban
  search and rescue.
\newblock In {\em 2008 IEEE/RSJ International Conference on Intelligent Robots
  and Systems}, pages 2103--2108.

\bibitem[Nobre et~al., 2017]{10.1007/978-3-319-50115-4_66}
Nobre, F., Heckman, C.~R., and Sibley, G.~T. (2017).
\newblock Multi-sensor slam with online self-calibration and change detection.
\newblock In Kuli{\'{c}}, D., Nakamura, Y., Khatib, O., and Venture, G.,
  editors, {\em 2016 International Symposium on Experimental Robotics}, pages
  764--774, Cham. Springer International Publishing.

\bibitem[Ohradzansky et~al., 2018]{Ohradzansky2018}
Ohradzansky, M., Alvarez, H.~E., Keshavan, J., Ranganathan, B., and Humbert,
  J.~S. (2018).
\newblock Autonomous bio-inspired small-object detection and avoidance.
\newblock In {\em 2018 IEEE International Conference on Robotics and Automation
  (ICRA)}, pages 1--9.

\bibitem[Ohradzansky et~al., 2020]{OhradzanskyMills}
Ohradzansky, M.~T., Mills, A.~B., Rush, E.~R., Riley, D.~G., Frew, E.~W., and
  Humbert, J.~S. (2020).
\newblock Reactive control and metric-topological planning for exploration.
\newblock In {\em 2020 IEEE International Conference on Robotics and Automation
  (ICRA)}, pages 4073--4079. IEEE.

\bibitem[Oleynikova et~al., 2017]{oleynikova2017voxblox}
Oleynikova, H., Taylor, Z., Fehr, M., Siegwart, R., and Nieto, J. (2017).
\newblock Voxblox: Incremental 3d euclidean signed distance fields for on-board
  mav planning.
\newblock In {\em 2017 Ieee/rsj International Conference on Intelligent Robots
  and Systems (iros)}, pages 1366--1373. IEEE.

\bibitem[Oleynikova et~al., 2018]{Oleynikova2018}
Oleynikova, H., Taylor, Z., Siegwart, R., and Nieto, J. (2018).
\newblock {Sparse 3D Topological Graphs for Micro-Aerial Vehicle Planning}.
\newblock In {\em Proceedings of the IEEE International Conference on
  Intelligent Robots and Systems}, pages 8478--8485, Madrid, Spain.

\bibitem[Olson et~al., 2012]{Olson2012}
Olson, E., Strom, J., Morton, R., Richardson, A., Ranganathan, P., Goeddel, R.,
  Bulic, M., Crossman, J., and Marinier~III, R. (2012).
\newblock Progress toward multi-robot reconnaissance and the magic 2010
  competition.
\newblock {\em Journal of Field Robotics}, 29:762--792.

\bibitem[Otsu et~al., 2020]{Otsu2020}
Otsu, K., Tepsuporn, S., Thakker, R., Vaquero, T.~S., Edlund, J.~A., Walsh, W.,
  Miles, G., Heywood, T., Wolf, M.~T., and Agha-Mohammadi, A.~A. (2020).
\newblock {Supervised Autonomy for Communication-degraded Subterranean
  Exploration by a Robot Team}.
\newblock {\em IEEE Aerospace Conference Proceedings}.

\bibitem[Palazzolo and Stachniss, 2017]{palazzolo2017information}
Palazzolo, E. and Stachniss, C. (2017).
\newblock Information-driven autonomous exploration for a vision-based mav.
\newblock {\em ISPRS Annals of the Photogrammetry, Remote Sensing and Spatial
  Information Sciences}, 4:59.

\bibitem[Qi et~al., 2017]{qi2017pointnet}
Qi, C.~R., Su, H., Mo, K., and Guibas, L.~J. (2017).
\newblock Pointnet: Deep learning on point sets for 3d classification and
  segmentation.
\newblock In {\em Proceedings of the IEEE conference on computer vision and
  pattern recognition}, pages 652--660.

\bibitem[Qin et~al., 2018]{qin_vins-mono:_2017}
Qin, T., Li, P., and Shen, S. (2018).
\newblock {VINS}-{Mono}: {A} {Robust} and {Versatile} {Monocular}
  {Visual}-{Inertial} {State} {Estimator}.
\newblock {\em IEEE Transactions on Robotics}, 34(4):1004--1020.

\bibitem[Redmon et~al., 2016]{redmon2016you}
Redmon, J., Divvala, S., Girshick, R., and Farhadi, A. (2016).
\newblock You only look once: Unified, real-time object detection.
\newblock In {\em Proceedings of the IEEE conference on computer vision and
  pattern recognition}, pages 779--788.

\bibitem[Riley and Frew, 2021]{RileyFrew}
Riley, D.~G. and Frew, E.~W. (2021).
\newblock Assessment of coordinated heterogeneous exploration of complex
  environments.
\newblock In {\em 2021 IEEE Conference on Control Technology and Applications
  (CCTA)}.

\bibitem[Rou{\v{c}}ek et~al., 2020]{Roucek2020}
Rou{\v{c}}ek, T., Pecka, M., {\v{C}}{\'{i}}{\v{z}}ek, P.,
  Petř{\'{i}}{\v{c}}ek, T., Bayer, J., {\v{S}}alansk{\'{y}}, V., Heřt, D.,
  Petrl{\'{i}}k, M., B{\'{a}}{\v{c}}a, T., Spurn{\'{y}}, V., Pomerleau, F.,
  Kubelka, V., Faigl, J., Zimmermann, K., Saska, M., Svoboda, T., and
  Krajn{\'{i}}k, T. (2020).
\newblock {DARPA Subterranean Challenge: Multi-robotic Exploration of
  Underground Environments}.
\newblock {\em Lecture Notes in Computer Science (including subseries Lecture
  Notes in Artificial Intelligence and Lecture Notes in Bioinformatics)}, 11995
  LNCS(April):274--290.

\bibitem[Santamaria-Navarro et~al., 2020]{santamaria-navarro_towards_2020}
Santamaria-Navarro, A., Thakker, R., Fan, D.~D., Morrell, B., and
  Agha-mohammadi, A.-a. (2020).
\newblock Towards {Resilient} {Autonomous} {Navigation} of {Drones}.
\newblock {\em arXiv:2008.09679 [cs]}.
\newblock arXiv: 2008.09679.

\bibitem[Santos-Victor and Sandini, 1997]{SantosVictor1997}
Santos-Victor, J. and Sandini, G. (1997).
\newblock Embedded visual behaviors for navigation.
\newblock {\em Robotics and Autonomous Systems}, 19:299--313.

\bibitem[Schmid et~al., 2020]{schmid2020efficient}
Schmid, L., Pantic, M., Khanna, R., Ott, L., Siegwart, R., and Nieto, J.
  (2020).
\newblock An efficient sampling-based method for online informative path
  planning in unknown environments.
\newblock {\em IEEE Robotics and Automation Letters}, 5(2):1500--1507.

\bibitem[Sethian, 1999]{sethian1999level}
Sethian, J.~A. (1999).
\newblock {\em Level set methods and fast marching methods: evolving interfaces
  in computational geometry, fluid mechanics, computer vision, and materials
  science}, volume~3.
\newblock Cambridge university press.

\bibitem[Shan et~al., 2020]{shan_liosam_2020}
Shan, T., Englot, B., Meyers, D., Wang, W., Ratti, C., and Rus, D. (2020).
\newblock {LIO}-{SAM}: {Tightly}-coupled {Lidar} {Inertial} {Odometry} via
  {Smoothing} and {Mapping}.
\newblock {\em arXiv:2007.00258 [cs]}.
\newblock arXiv: 2007.00258.

\bibitem[Sheng et~al., 2004]{Sheng2004}
Sheng, W., Yang, Q., Ci, S., and Xi, N. (2004).
\newblock {Multi-robot area exploration with limited-range communications}.
\newblock In {\em 2004 IEEE/RSJ International Conference on Intelligent Robots
  and Systems (IROS)}, volume~2, pages 1414--1419, Sendai, Japan.

\bibitem[Shi et~al., 2019]{shi2019pointrcnn}
Shi, S., Wang, X., and Li, H. (2019).
\newblock Pointrcnn: 3d object proposal generation and detection from point
  cloud.
\newblock In {\em Proceedings of the IEEE Conference on Computer Vision and
  Pattern Recognition}, pages 770--779.

\bibitem[Simmons et~al., 2000]{Simmons2000}
Simmons, R., Apfelbaum, D., Burgard, W., and Fox, D. (2000).
\newblock {Coordination for multi-robot exploration and mapping}.
\newblock In {\em Proceedings of the AAAI National Conference on Artificial
  Intelligence}, pages 852--858, Austin, TX.

\bibitem[Smallwood and Sondik, 1973]{smallwood1973optimal}
Smallwood, R.~D. and Sondik, E.~J. (1973).
\newblock The optimal control of partially observable markov processes over a
  finite horizon.
\newblock {\em Operations research}, 21(5):1071--1088.

\bibitem[Smith, 2013]{smith2013sequential}
Smith, A. (2013).
\newblock {\em Sequential Monte Carlo methods in practice}.
\newblock Springer Science \& Business Media.

\bibitem[Smith and Hollinger, 2018]{Smith2018}
Smith, A.~J. and Hollinger, G.~A. (2018).
\newblock {Distributed inference-based multi-robot exploration}.
\newblock {\em Autonomous Robots}, 42(8):1651--1668.

\bibitem[Srinivasan et~al., 1998]{Srinivasan1998RobotNI}
Srinivasan, M.~V., Chahl, J.~S., Weber, K., Venkatesh, S., Nagle, M.~G., and
  Zhang, S.-W. (1998).
\newblock Robot navigation inspired by principles of insect vision.
\newblock {\em Robotics and Autonomous Systems}, 26:203--216.

\bibitem[Thrun et~al., 2004]{Thrun2004}
Thrun, S., Thayer, S., Whittaker, W., Baker, C., Burgard, W., Ferguson, D.,
  Hahnel, D., Montemerlo, M., Morris, A., Omohundro, Z., Reverte, C., and
  Whittaker, W. (2004).
\newblock {Autonomous exploration and mapping of abandoned mines: Software
  architecture of an autonomous robotic system}.
\newblock {\em IEEE Robotics and Automation Magazine}, 11(4):79--91.

\bibitem[Vanholder, 2016]{vanholder2016efficient}
Vanholder, H. (2016).
\newblock Efficient inference with tensorrt.

\bibitem[Wang and Lewis, 2007]{Wang2007}
Wang, J. and Lewis, M. (2007).
\newblock Human control for cooperating robot teams.
\newblock pages 9--16.

\bibitem[Wang and Olson, 2016]{Wang2016}
Wang, J. and Olson, E. (2016).
\newblock {AprilTag 2: Efficient and robust fiducial detection}.
\newblock In {\em 2016 IEEE/RSJ International Conference on Intelligent Robots
  and Systems (IROS)}, pages 4193--4198. IEEE.

\bibitem[Werger and Mataric, 2000]{Werger2000}
Werger, B.~B. and Mataric, M.~J. (2000).
\newblock {Broadcast of local eligibility}.
\newblock In {\em Proceedings of the fourth international conference on
  Autonomous agents - AGENTS '00}, pages 21--22, New York, New York, USA. ACM
  Press.

\bibitem[Yamauchi, 1999]{Yamauchi1999}
Yamauchi, B. (1999).
\newblock {Decentralized coordination for multirobot exploration}.
\newblock {\em Robotics and Autonomous Systems}, 29(2-3):111--118.

\bibitem[Zhan et~al., 2018]{Zhan_2018_CVPR}
Zhan, H., Garg, R., Saroj~Weerasekera, C., Li, K., Agarwal, H., and Reid, I.
  (2018).
\newblock Unsupervised learning of monocular depth estimation and visual
  odometry with deep feature reconstruction.
\newblock In {\em The IEEE Conference on Computer Vision and Pattern
  Recognition (CVPR)}.

\bibitem[Zhang and Singh, 2017]{zhang_low-drift_2017}
Zhang, J. and Singh, S. (2017).
\newblock Low-drift and real-time lidar odometry and mapping.
\newblock {\em Autonomous Robots}, 41(2):401--416.

\bibitem[Zhang and Suen, 1984]{zhang1984fast}
Zhang, T. and Suen, C.~Y. (1984).
\newblock A fast parallel algorithm for thinning digital patterns.
\newblock {\em Communications of the ACM}, 27(3):236--239.

\bibitem[Zhang et~al., 2001]{zhang2001radio}
Zhang, Y.~P., Zheng, G.~X., and Sheng, J. (2001).
\newblock Radio propagation at 900 {MHz} in underground coal mines.
\newblock {\em IEEE Transactions on Antennas and Propagation}, 49(5):757--762.

\bibitem[Zhao, 2007]{zhao2007parallel}
Zhao, H. (2007).
\newblock Parallel implementations of the fast sweeping method.
\newblock {\em Journal of Computational Mathematics}, pages 421--429.

\bibitem[Zhou and Tuzel, 2018]{zhou2018voxelnet}
Zhou, Y. and Tuzel, O. (2018).
\newblock Voxelnet: End-to-end learning for point cloud based 3d object
  detection.
\newblock In {\em Proceedings of the IEEE Conference on Computer Vision and
  Pattern Recognition}, pages 4490--4499.

\bibitem[Zlot et~al., 2002]{Zlot2002}
Zlot, R., Stentz, A.~T., Dias, M.~B., and Thayer, S. (2002).
\newblock {Multi-robot exploration controlled by a market economy}.
\newblock In {\em Proceedings of the IEEE International Conference on Robotics
  and Automation}, volume~3, pages 3016--3023, Washington, DC.

\end{thebibliography}

\end{document}